\documentclass{article}
\usepackage{arxiv}

\usepackage{pdflscape,rotating,placeins}

\usepackage{mathptmx}
\DeclareMathAlphabet{\mathcal}{OMS}{cmsy}{m}{n}

\RequirePackage{fix-cm}
\usepackage{siunitx}
\usepackage{amsmath,bm}

\usepackage[utf8]{inputenc}
\usepackage[T1]{fontenc}
\usepackage{graphicx}
\usepackage[figurename=Fig.,labelfont=bf,labelsep=period]{caption}
\usepackage{subcaption}
\usepackage{amsfonts}
\usepackage{amssymb}
\usepackage{amsbsy} 

\usepackage[colorlinks=true,citecolor=red,linkcolor=black]{hyperref}

\usepackage{url}            
\usepackage{booktabs}       
\usepackage{nicefrac}       
\usepackage{microtype}      
\usepackage{lipsum}
\usepackage{algorithm,tikz}
\usepackage[]{algpseudocode}
\usepackage{pgfplots}
\usepackage{subcaption}
\usepackage[]{threeparttable}
\usepackage{multirow,bm}

\newcommand{\transpose}{^{\mathrm{T}}}
\newcommand{\superscript}[1]{\ensuremath{^{\textrm{#1}}}}
\newcommand{\subscript}[1]{\ensuremath{_{\textrm{#1}}}}

\newcommand{\ubar}[1]{\underline{#1}}

\newcommand{\Am}{{\mathbf A}}

\newcommand{\Bm}{{\mathbf B}}

\newcommand{\Cm}{{\mathbf C}}

\newcommand{\Dm}{\mathbf D}

\newcommand{\Dtr}{\mathcal{D}_\mathrm{tr}}
\newcommand{\Dval}{\mathcal{D}_\mathrm{val}}

\newcommand{\Em}{\mathbf{E}}

\newcommand{\g}{{\mathbf g }}

\newcommand{\kpre}{k\subscript{pre}}

\newcommand{\kpost}{k\subscript{post}}
\newcommand{\Km}{{\mathbf K}}

\newcommand{\Lm}{{\mathbf L}}

\newcommand{\Mm}{{\mathbf M}}

\newcommand{\ny}{n\subscript{y}}
\newcommand{\npow}{n\subscript{pow}}
 
\newcommand{\nw}{n_\mathrm{w}}
 
\newcommand{\ngm}{n_\mathrm{g}} 
\newcommand{\Ntr}{N_\mathrm{tr}}
\newcommand{\Nval}{N_\mathrm{val}}

\newcommand{\Qy}{Q\subscript{y}}

\newcommand{\rr}{{\mathbf r}}

\newcommand{\rk}{r\subscript{k}}

\newcommand{\ub}{u\subscript{b}}

\newcommand{\uu}{{\mathbf u }}

\newcommand{\w}{{\mathbf w }}

\newcommand{\x}{{\mathbf x }}
\newcommand{\xo}{{\overline{\mathbf x }}}

\newcommand{\X}{{\mathbf X}}

\newcommand{\y}{{\mathbf y}}
\newcommand{\Y}{{\mathbf Y}}

\newcommand{\ksi}{{\zeta}}

\newcommand{\xii}{\boldsymbol{\xi}}

\newcommand{\thetaa}{\boldsymbol{\theta}}
 
\newcommand{\chii}{\boldsymbol{\chi}}
\newcommand{\epsv}{{\varepsilon_\mathrm{val}}}

\newcommand{\sgn}{\operatorname{sgn}}

\makeatletter
\newcommand*\dashline{~\rotatebox[origin=c]{90}{\scalebox{0.85}{$\dabar@\dabar@\dabar@$}}~}
\makeatother

\newcommand\Tstrut{\rule{0pt}{2.6ex}}         
\newcommand\Bstrut{\rule[-0.9ex]{0pt}{0pt}}   

\usepackage{tikz}
\usetikzlibrary{matrix,chains,positioning,decorations.pathreplacing,arrows}
\usetikzlibrary{positioning,calc}
\usetikzlibrary{patterns,decorations.pathmorphing,decorations.markings}
\usetikzlibrary{shapes}
\usetikzlibrary{fit}
\tikzstyle{block} = [draw,rectangle,thick,minimum height=2em,minimum width=2em]
\tikzstyle{sum} = [draw,circle,inner sep=0mm,minimum size=2mm]
\tikzstyle{connector} = [->,thick]
\tikzstyle{line} = [thick]
\tikzstyle{branch} = [circle,inner sep=0pt,minimum size=1mm,fill=black,draw=black]
\tikzstyle{guide} = []
\tikzset{%
	every neuron/.style={
		circle,
		draw,
		fill = green!20,
		thick,
		minimum size=1.1cm
	},
	neuron missing/.style={
		draw=none, 
		scale=4,
		fill = none,
		text height=0.333cm,
		execute at begin node=\color{black}$\vdots$
	},
	input neuron/.style={
		circle,
		draw,
		thick,
		fill = yellow!20,
		minimum size=1cm
	},
	output neuron/.style={
		circle,
		draw,
		thick,
		fill = red!20,
		minimum size=1cm
	},
}
\tikzset{%
wind turbine/.pic={
  \tikzset{path/.style={fill, draw=white, ultra thick, line join=round}}
  \path [path] 
    (-.25,0) arc (180:360:.25 and .0625) -- (.0625,3) -- (-.0625,3) -- cycle;
  \foreach \i in {90, 210, 330}{
    \ifcase#1
    \or
      \path [path, shift=(90:3), rotate=\i] 
        (.5,-.1875) arc (270:90:.5 and .1875) arc (90:-90:1.5 and .1875);
    \or
      \path [path, shift=(90:3), rotate=\i] 
        (0,0.125) -- (2,0.125) -- (2,0) -- (0.5,-0.375) -- cycle;
    \or
      \path [path, shift=(90:3), rotate=\i]
        (0,-0.125) arc (180:0:1 and 0.125) -- ++(0,0.125) arc (0:180:1 and 0.25) -- cycle;
    \fi
  }
  \path [path] (0,3) circle [radius=.25];
}} 

\usetikzlibrary{3d,decorations.text,shapes.arrows,positioning,fit,backgrounds}
\tikzset{pics/fake box/.style args={
		#1 with dimensions #2 and #3 and #4}{
		code={
			\draw[gray,ultra thin,fill=#1]  (0,0,0) coordinate(-front-bottom-left) to
			++ (0,#3,0) coordinate(-front-top-right) --++
			(#2,0,0) coordinate(-front-top-right) --++ (0,-#3,0) 
			coordinate(-front-bottom-right) -- cycle;
			\draw[gray,ultra thin,fill=#1] (0,#3,0)  --++ 
			(0,0,#4) coordinate(-back-top-left) --++ (#2,0,0) 
			coordinate(-back-top-right) --++ (0,0,-#4)  -- cycle;
			\draw[gray,ultra thin,fill=#1!80!black] (#2,0,0) --++ (0,0,#4) coordinate(-back-bottom-right)
			--++ (0,#3,0) --++ (0,0,-#4) -- cycle;
			\path[gray,decorate,decoration={text effects along path,text={}}] (#2/2,{2+(#3-2)/2},0) -- (#2/2,0,0);
		}
}}
\tikzset{circle dotted/.style={dash pattern=on .05mm off 2mm,
		line cap=round}}
 
\title{A Bi-fidelity DeepONet Approach for Modeling Uncertain and Degrading Hysteretic Systems}

\author{
  Subhayan De\\
Mechanical Engineering\\
Northern Arizona University\\
Flagstaff, AZ 86011\\
  \texttt{Subhayan.De@nau.edu} \\
  \And 
  Patrick T. Brewick\\
  Civil and Environmental Engineering\\
  and Earth Sciences\\
  University of Notre Dame\\
  Notre Dame, IN 46556\\
  \texttt{pbrewick@nd.edu}\\
} 

\pgfplotsset{compat=1.16}

\begin{document}

\maketitle

\begin{abstract}

Nonlinear systems, such as with degrading hysteretic behavior, are often encountered in engineering applications. In addition, due to the ubiquitous presence of uncertainty and the modeling of such systems becomes increasingly difficult. 
On the other hand, datasets from pristine models developed without knowing the nature of the degrading effects can be easily obtained. 
In this paper, we use datasets from pristine models without considering the degrading effects of hysteretic systems as low-fidelity representations that capture many of the important characteristics of the true system's behavior to train a deep operator network (DeepONet). Three numerical examples are used to show that the proposed use of the DeepONets to model the discrepancies between the low-fidelity model and the true system's response leads to significant improvements in the prediction error in the presence of uncertainty in the model parameters for degrading hysteretic systems. 

\end{abstract}

\section{Introduction} 

Uncertainty is a universal concern when building or constructing models of physical systems.  For instance, parametric uncertainty can manifest in the material properties, geometry and/or loading conditions of a chosen model of engineering system, differing between realizations of the system.  However, structural uncertainty also exists \cite{kennedy2001bayesian} because some physical phenomena are often ignored, idealized, or simplified during model construction.  Both parametric and structural uncertainties can be found in models of real-world engineering system, and, as such, these uncertainty sources must be properly addressed if the model is intended to serve as a direct representation, or digital twin, of the real-world system. 
Fortunately, numerous techniques exist for incorporating parametric uncertainty into a model environment, including polynomial chaos expansion \cite{ghanem2003stochastic,xiu2002wiener},
Gaussian process regression \cite{williams2006gaussian,forrester2007multi}, and other response surface-based methods \cite{isukapalli1998stochastic,giunta2006promise,chi2012response}. However, the cost associated with developing these models rises sharply as the
dimension of the uncertain variables increases \cite{doostan2011non}.

Within the scientific machine learning (SciML) community, neural networks have been increasingly utilized for modeling physical systems in the presence of uncertainty \cite{tripathy2018deep,de2021uncertainty}.  Raissi et al.~\cite{raissi2019physics} augmented the standard data discrepancy loss in their network with the residual of the corresponding governing equations, creating a so-called ``physics-informed neural network'' or PINN.  This gave rise to significant interest in employing PINNs to solve several different problems in computational mechanics, with some notable examples including Karniadakis et al. \cite{karniadakis2021physics}, Cai et al. \cite{cai2021physics}, and Viana and Subramaniyan \cite{viana2021survey}. Among the numerous applications of PINNs, most typically focused on quantifying parametric uncertainties in models \cite{yang2020physics,yang2021b,zhu2019physics,yang2019adversarial,geneva2020modeling,winovich2019convpde}, while SciML studies that address structural uncertainty in modeling remain scarce. Notably, Blasketh et al. \cite{blakseth2021deep} combined neural networks with governing equations to capture modeling and discretization errors. Also, Zhang et al. \cite{zhang2019quantifying} adopted a physics-informed approach to quantify both the parametric and approximation uncertainty of the neural network. 

More recent developments in SciML have drawn inspiration from the universal approximation theorem for operators in Chen and Chen \cite{chen1995universal}.  This has led to the proposal of several novel network architectures that are capable of approximating mesh-independent solution operators for the governing equations of physical systems \cite{lu2019deeponet,li2020neural,li2020multipole,li2020fourier,li2021markov,lu2021learning,li2021physics}. These architectures manage to overcome two particular shortcomings related to how PINNs learn a system’s behavior.  First, these new neural operators are not dependent on the mesh used to generate the training dataset. Secondly, these neural operators are truly data-driven in that they do not require any prior knowledge of the governing equations. The deep operator network (DeepONet), proposed in Lu et al. \cite{lu2019deeponet,lu2021learning}, is one example of these network operator architectures.  DeepONets utilize two separate networks -- a branch network and a trunk network.  For the trunk network, the inputs are the coordinates where the system response is being queried, whereas the branch network has the external source term serve as its input. The system response is then produced by combining the outputs of these two networks in a linear regression model.

Numerous examples of the emerging power of DeepONets abound.  For instance, Lu et al. \cite{lu2021one} demonstrated how the computational domain could be decomposed into several small subdomains to facilitate learning the governing equation with a single solution of the equation. Cai et al. \cite{cai2021deepm} used DeepONets for a multi-physics problem that involved modeling field variables across multiple scales. Wang et al. \cite{wang2021learning,wang2021long} trained DeepONets based on the error resulting from satisfying the governing equations.  Goswami et al. \cite{goswami2021physics} applied a variational formulation of the governing equations to model brittle fracture.  The DeepONet architecture is compared against the iterative neural operator architecture originally proposed in Li et al. \cite{li2020fourier} by both Kovachki et al. \cite{kovachki2021neural} and Lu et al. \cite{lu2021comprehensive}. Approximation errors, error bounds, and convergence rates for these neural operator architectures are the primary focus of several subsequently works, notably Kovachki et al. \cite{kovachki2021universal}, Lanthaler et al. \cite{lanthaler2021error}, Deng et al. \cite{deng2021convergence}, and Marcati and Schwab \cite{marcati2021exponential}.  

One of the limiting factors routinely encountered in SciML is the absence of sufficient data for training and DeepONets are no exception.  For complex physical systems, the amount of training data capable of being generated is often limited by the computational burdens associated with simulating those systems.  It is similarly infeasible to repeat experimental or field tests enough times to obtain a sufficient volume of data.  Thus, accruing enough data to properly train a DeepONet can quickly become an intractable problem.  Fortunately, alternative methods have recently been suggested that leverage models at multiple levels of fidelity to overcome this training data challenge \cite{de2020transfer,de2022bi}.  In the presence of a small or limited data sets from the true (high-fidelity) system, a similar but computationally inexpensive model, known as low-fidelity, can often be deployed to capture many critical aspects of the true system’s behavior. In their work, De et al. \cite{de2022bi} trained a DeepONet that uses the low-fidelity response as the input of the branch network and then predicted the discrepancy between the low- and high-fidelity data with the understanding that it is often more feasible to predict the discrepancy than the true (high-fidelity) system response.  Importantly, modeling the discrepancy requires only a small training data set of the true system’s response, meaning a bi-fidelity approach can be highly advantageous when training data from high-fidelity simulations is scarce. On a similar approach, Lu et al. \cite{lu2022multifidelity} trained a separate DeepONet with the low-fidelity data along with a DeepONet for the discrepancy. 
Howard et al. \cite{howard2022multifidelity} implemented DeepONets to learn the linear and nonlinear correlation between the high- and low-fidelity data and also applied to the closure problem of multiscale systems \cite{ahmed2023multifidelity}. 
Thakur et al. \cite{thakur2022multi} combined wavelet transformation with kernel integral operator \cite{li2020fourier,tripura2022wavelet} and applied to model the discrepancy between the high- and low-fidelity data. 

Nonlinearity is one type of system complexity that is regularly encountered in engineering systems.  Properly capturing the full extent of nonlinear behavior often requires highly detailed and parameterized models.  For instance, Bouc and Wen \cite{wen1976method} proposed a model for hysteretic systems that has since been regularly applied to describe the behavior of seismic isolation devices \cite{ramallo2002smart}  While the initial model had only a few parameters, it was later expanded upon to include the effects of degradation \cite{baber1981random} to account for strength deterioration and stiffness reduction, as well as pinching \cite{noori1985random} to account for the sudden loss of stiffness encountered with the opening and closing of cracks, such as those in concrete and masonry structures.  While these enhancements to the original Bouc-Wen model for hysteresis help align model performance with real-world observations, they also introduce additional layers of parametric and computational complexity.  These challenges only compound when a hysteretic element becomes merely a component in a much larger modeling framework, such as when modeling a base isolation system for a full building model.  Further, as previously identified, it is also paramount to consider the parametric and structural uncertainties certain to be encountered when attempting to model the response of a degrading hysteretic system.

This paper proposes using a bi-fidelity DeepONet to generate response predictions for degrading hysteretic systems under uncertainty.  Three numerical examples are presented that demonstrate the performance of the proposed bi-fidelity DeepONets for systems covering a range of complexity.  The first two examples consider the case of a base-isolation system modeled using 4 and 100 degrees-of-freedom, respectively.  For these examples, the low-fidelity models use only the classical Bouc-Wen formulation, ignoring the degrading and pinching effects included in the true, high-fidelity system.  The third example considers a half-car model with non-linear suspension system modeled with degrading hysteresis.  In this case, the low-fidelity system only considers a quarter-car model, which removes two degrees of freedom and several parameters when compared to the true half-car model.

The remainder of the paper is organized as follows.  A brief background on the bi-fidelity DeepONet approach and descriptions of the considered hysteretic models are both provided in Section 2.  Section 3 presents the proposed approach for modeling the discrepancy between a low-fidelity and true response for uncertain and partially unknown nonlinear dynamical systems.  The three aforementioned numerical examples are presented in Section 4 along with a discussion of their results, which is followed by concluding remarks and suggestions for future directions.

\section{Background}  
In this section, we discuss the neural operator architecture, known as the DeepONet \cite{lu2019deeponet}, and its training process. Thereafter, we briefly discuss the degrading hysteretic model that is used in this paper.

\subsection{Deep Operator Network (DeepONet) \cite{lu2019deeponet}} \label{sec:DeepONet_back}

\begin{figure}[!htb]

	\centering
	\begin{tikzpicture} 
	\draw[draw=cyan,fill=cyan!10,thick,dashed,rounded corners=2ex] (-5,-2.4) rectangle ++(8.6,4); 
\node[rotate=90,color=cyan] at (-5.3,-0.4) {Branch Network}; 
\draw[draw=magenta,fill=magenta!10,thick,dashed,rounded corners=2ex] (-5,-7) rectangle ++(8.6,4); 
\node[rotate=90,color=magenta] at (-5.3,-5) {Trunk Network}; 
	\node[inner sep=0pt] (russell) at (-0.68,0)
    {\includegraphics[scale=0.108]{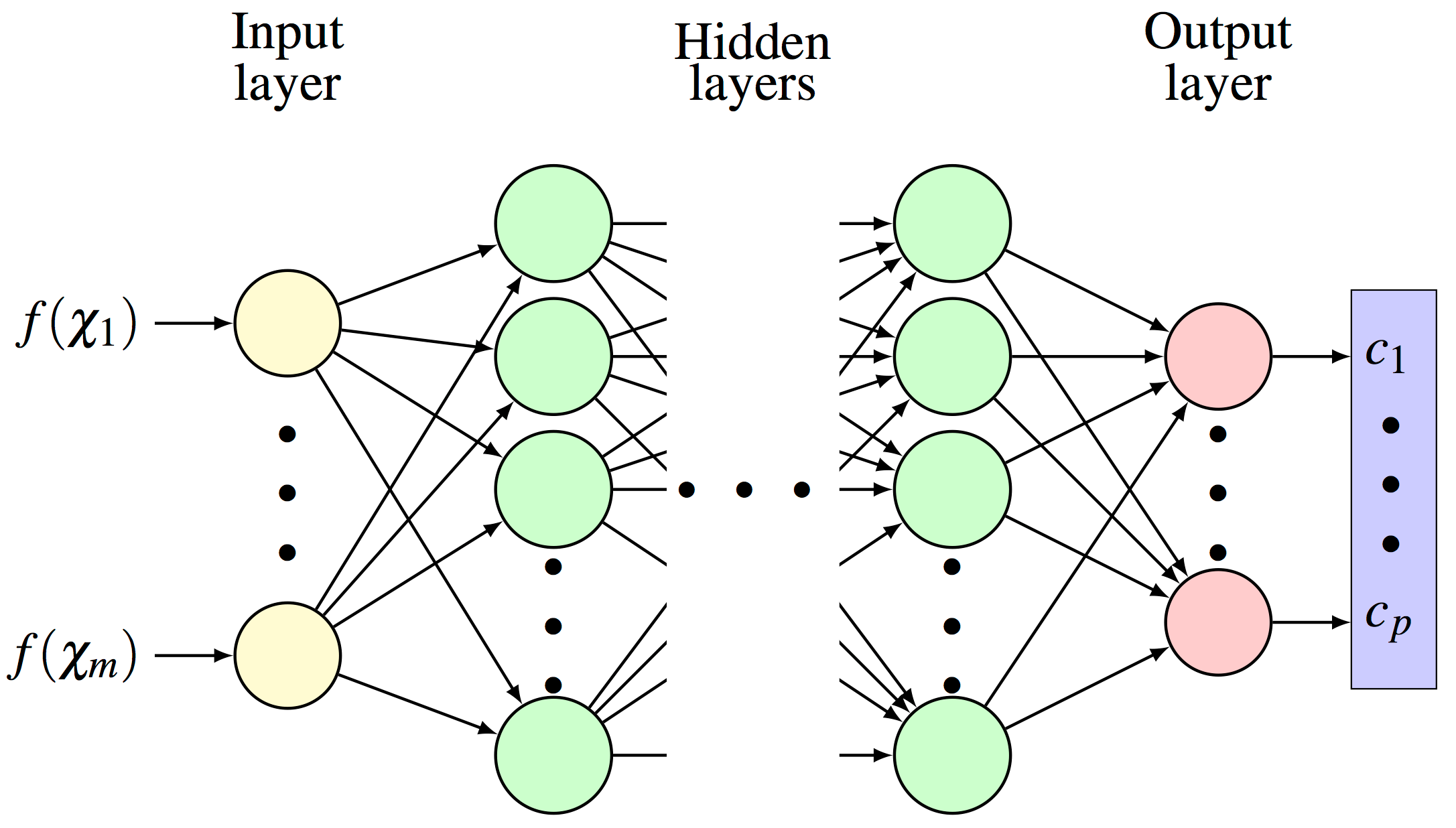}};
    \node[inner sep=0pt] (russell) at (-0.4,-5)
    {\includegraphics[scale=0.108]{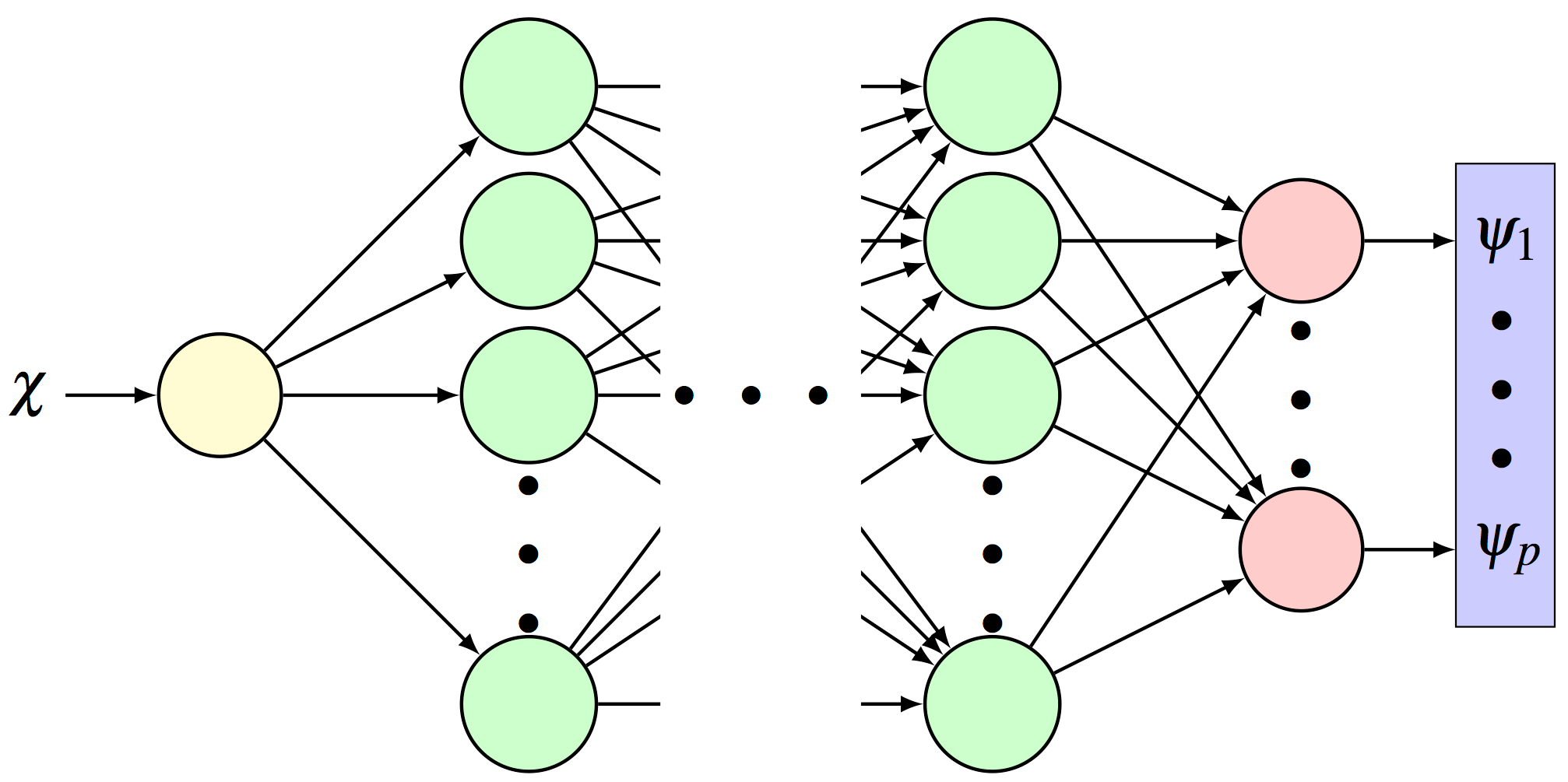}};
    \draw [thick] (3.375,-0.5) -- (4,-0.5); 
    \draw [-latex,thick] (4,-0.5) -- (4,-2.25); 
    \draw [thick] (3.36,-5) -- (4,-5); 
    \draw [-latex,thick] (4,-5) -- (4,-2.75); 
    \draw[-latex,thick] (4,-2.5) -- (5.25,-2.5); 
    \draw[-latex,thick] (5.5,-1.5) -- (5.5,-2.225); 
    \node[draw=black,thick,circle,fill=blue!10,inner sep=0pt,minimum size=6pt] () at (4,-2.5) {\Large$\times$}; 
    \draw [-latex,thick] (5.5,-2.5) -- (6.75,-2.5) node [right] {$G_{\thetaa}(f)(\chii)$}; 
    \node[draw=black,thick,circle,fill=blue!10,inner sep=0pt,minimum size=6pt] () at (5.5,-2.5) {\Large$+$}; 
    \node[] at (5.5,-1.25) {$c_0$}; 
    \end{tikzpicture}
	\caption{DeepONet architecture (stacked) \cite{lu2019deeponet} used in this paper. }\label{fig:deeponet} 
\end{figure}
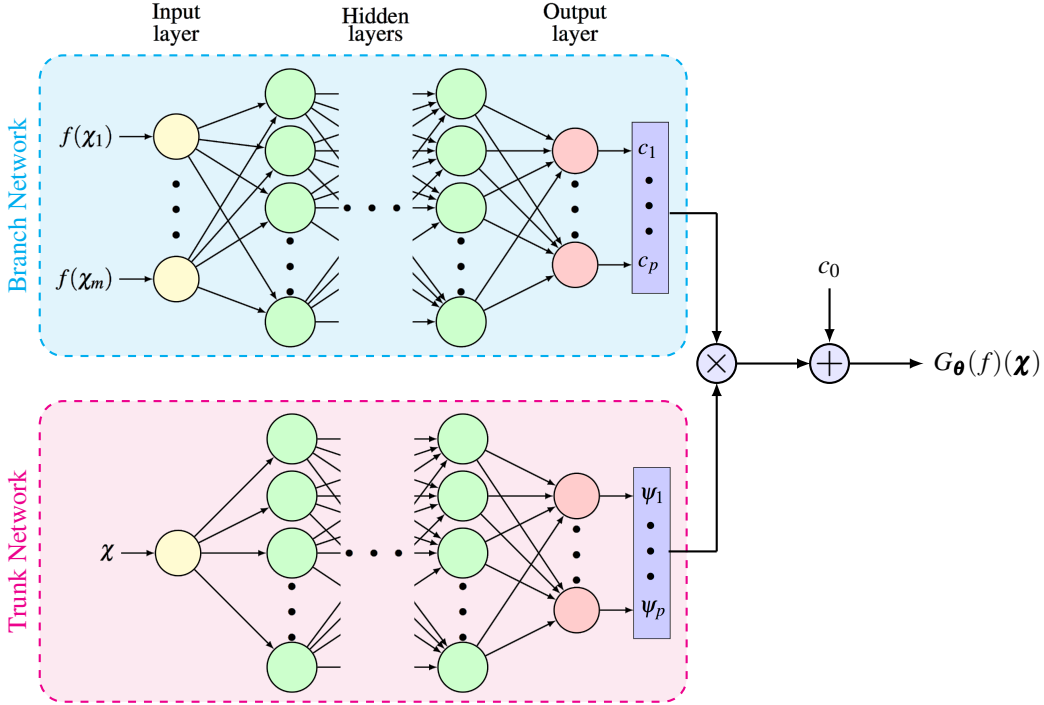 

Consider a physical system governed by the following equation
\begin{equation} \label{eq:gov}
    \mathcal{L}(y)(\chii) = f(\chii), \qquad \chii \in \Omega,
\end{equation} 
where $\mathcal{L}$ denotes a differential operator; $y$ is the response of the system; $\chii$ is a location inside the domain $\Omega$; and $f$ is the external force. 
For an initial value problem, as considered in this paper, an initial condition $y(\chii_0) = y_0$ is also used. Let us assume the solution operator of \eqref{eq:gov} is given by $G(f)(\chii)$. The DeepONet architecture proposed in Lu et al. \cite{lu2019deeponet} based on the universal approximation theorem in Chen and Chen \cite{chen1995universal} approximates this solution as 
\begin{equation} \label{eq:deep} 
    y(\chii)\approx G_{\thetaa}(f)(\chii) = c_0 + \sum_{i=1}^p c_i(f(\chii_1),\dots,f(\chii_m))\psi_i(\chii), 
\end{equation}
where $c_0$ is a constant bias parameter. The coefficients, $c_i(\cdot),~i=1,\dots,p,$ are the output from a neural network, known as the \textit{branch network}, with external force $f(\cdot)$ measured at $\{\chii_i\}_{i=1}^m$ locations as input. The bases, $\psi_i(\cdot),~i=1,\dots,p,$ are the output from a neural network, known as the \textit{trunk network}, with the location $\chii$ as the input. Hence, the trainable parameters $\thetaa$ of the DeepONet contain weights and biases from the \textit{branch} and \textit{trunk} networks as well as the constant bias $c_0$. A schematic of the \textit{branch} and \textit{trunk} networks in a DeepONet is shown in Figure \ref{fig:deeponet}. 
Due to its construction in \eqref{eq:deep}, the DeepONet can learn the solution operator that is independent of the discretizaion used in generating the training data. 

The parameters $\thetaa$ of the DeepONet can be estimated  using a training dataset $\Dtr$ that consists of $N_d$ measurements of the response at locations $\{\chii_i\}_{i=1}^{N_d}$ for each of $N_f$ realizations of the external force $\{f_j(\chii)\}_{j=1}^{N_f}$ by solving the following optimization problem 
\begin{equation} \label{eq:opt}
\begin{split}
 \min\limits_{\thetaa} J(\thetaa) :&= \frac{1}{N_fN_d} \sum_{j=1}^{N_f}\sum_{i=1}^{N_d} \lvert G_{\thetaa}(f_j)(\chii_i) - G(f_j)(\chii_i) \rvert^2\\
 &= \frac{1}{N_fN_d} \sum_{j=1}^{N_f}\sum_{i=1}^{N_d} \left\lvert c_0 + \sum_{k=1}^p c_k(f_j(\chii_1),\dots,f_j(\chii_m))\psi_k(\chii_i) - G(f_j)(\chii_i) \right\rvert^2.\\
\end{split}
\end{equation} 
We use the Adam algorithm \cite{kingma2014adam}, a popular variant of the stochastic gradient descent \cite{bottou2010large,bottou2012stochastic,de2020topology}, to solve the optimization problem \eqref{eq:opt} in this paper. 
Note that the locations $\{\chii_i\}_{i=1}^{N_d}$ can be different than the location $\chii$ where we would seek the response from the trained DeepONet. While different realizations of the external force $f(\chii)$ can be sampled from function spaces, such as Gaussian random field and Chebyshev polynomials \cite{lu2019deeponet}, we utilize samples from the probability distributions of the uncertain variables of the problem to instantiate multiple realizations of $f(\chii)$.

\subsection{Hysteretic System Model} \label{sec:hyst}
\begin{figure}[!htb]
    \centering
    \includegraphics[scale=1.2]{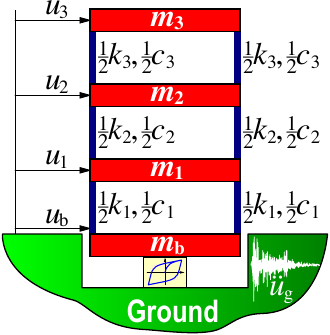}\\
    \includegraphics[scale=1.2]{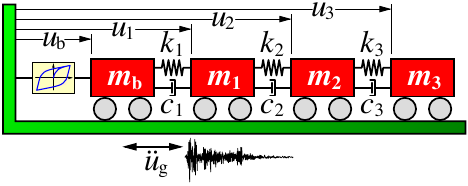}
    \caption{Four degree-of-freedom system with hysteretic base layer.}
    \label{fig:4dof}
\end{figure}

In this paper, we use the Bouc-Wen formulation \cite{wen1976method} to model hysteretic systems, which is smoother and more realistic \cite{Nagarajaiah2000} than other models of hysteresis, such as the bilinear model. 
To illustrate the model and the effects of degradation, let us consider a four degree-of-freedom system adapted from De et al. \cite{de2019hybrid} as shown in Figure \ref{fig:4dof}, which is also used as the first example in Section \ref{sec:examples}. 
In this model, the non-elastic force from the hysteretic component of the structure is given by $q_\mathrm{y}z$, where $q_{\mathrm{y}}=\Qy\left(1-1/\rk\right)$, $\kpre$ is the pre-yield stiffness, $\kpost$ is the post-yield stiffness, $\Qy$ is the yield force, $\rk=\kpost/\kpre$ is the hardening ratio, and $z$ is an evolutionary variable. In particular, the evolution of $z$ can be given by \cite{ma2004parameter} 
\begin{equation} \label{eq:bouc}
\dot z = g_\mathrm{BW}:= A \dot u\subscript{b} - \beta \dot u\subscript{b}|z|^{\npow}-\gamma z|\dot u\subscript{b}||z|^{\npow-1}, 
\end{equation}
where $u\subscript{b}$ is the displacement of the base layer of the structure and the choice of $A = 2\beta = 2\gamma = \kpre/\Qy$ makes $z$ stay in $[-1,1]$ with consistent loading and unloading stiffnesses \cite{spencer1}.  The variable $\npow$ controls the sharpness of the hysteresis loop corners and is chosen as one in this paper.  

\begin{figure}[!htb]
     \centering
     \begin{subfigure}[b]{0.45\textwidth}
         \centering
         \includegraphics[width=\textwidth]{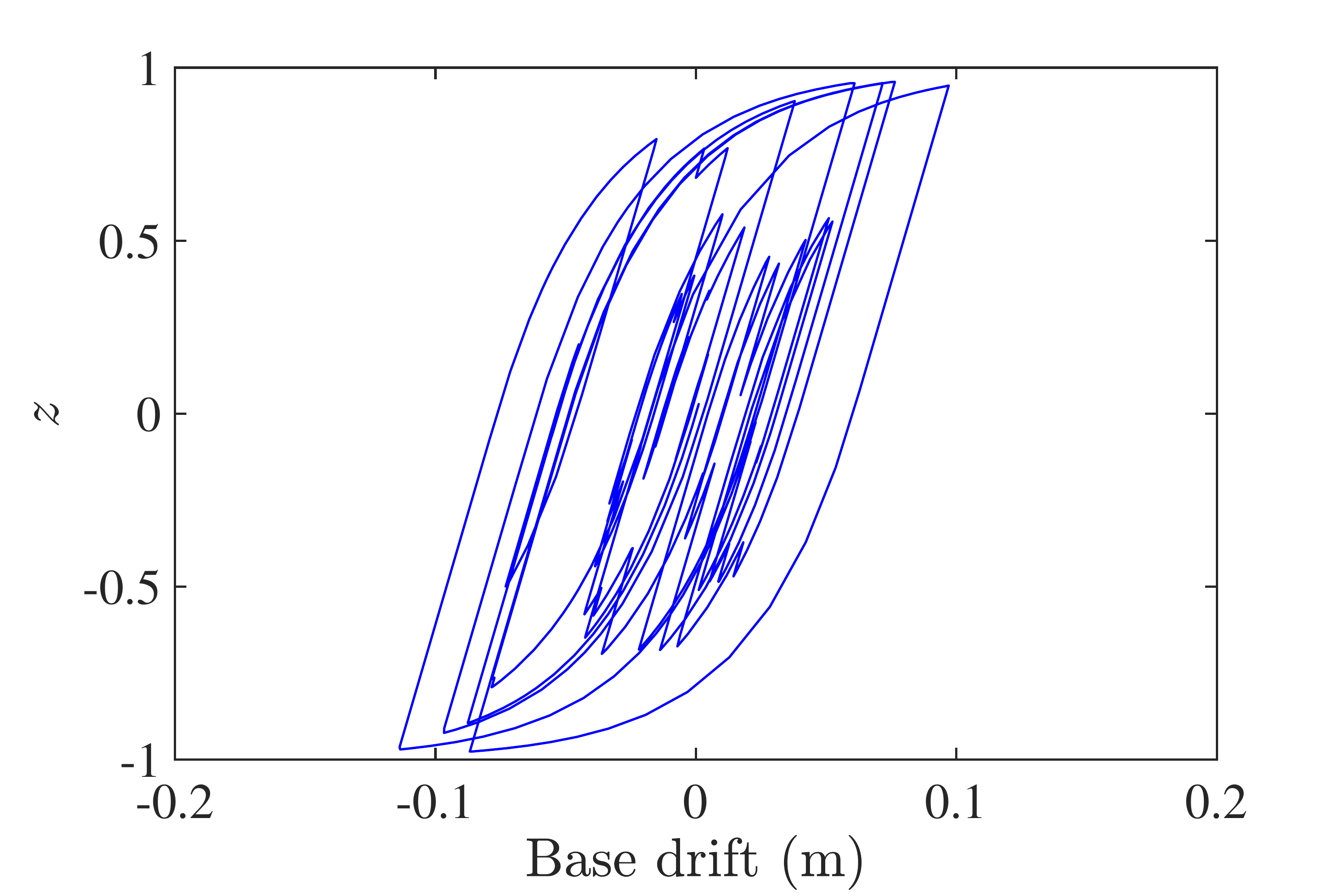}
         \caption{Bouc-Wen hysteresis model (see \eqref{eq:bouc})}
         \label{fig:bouc}
     \end{subfigure}
     \hfill
     \begin{subfigure}[b]{0.45\textwidth}
         \centering
         \includegraphics[width=\textwidth]{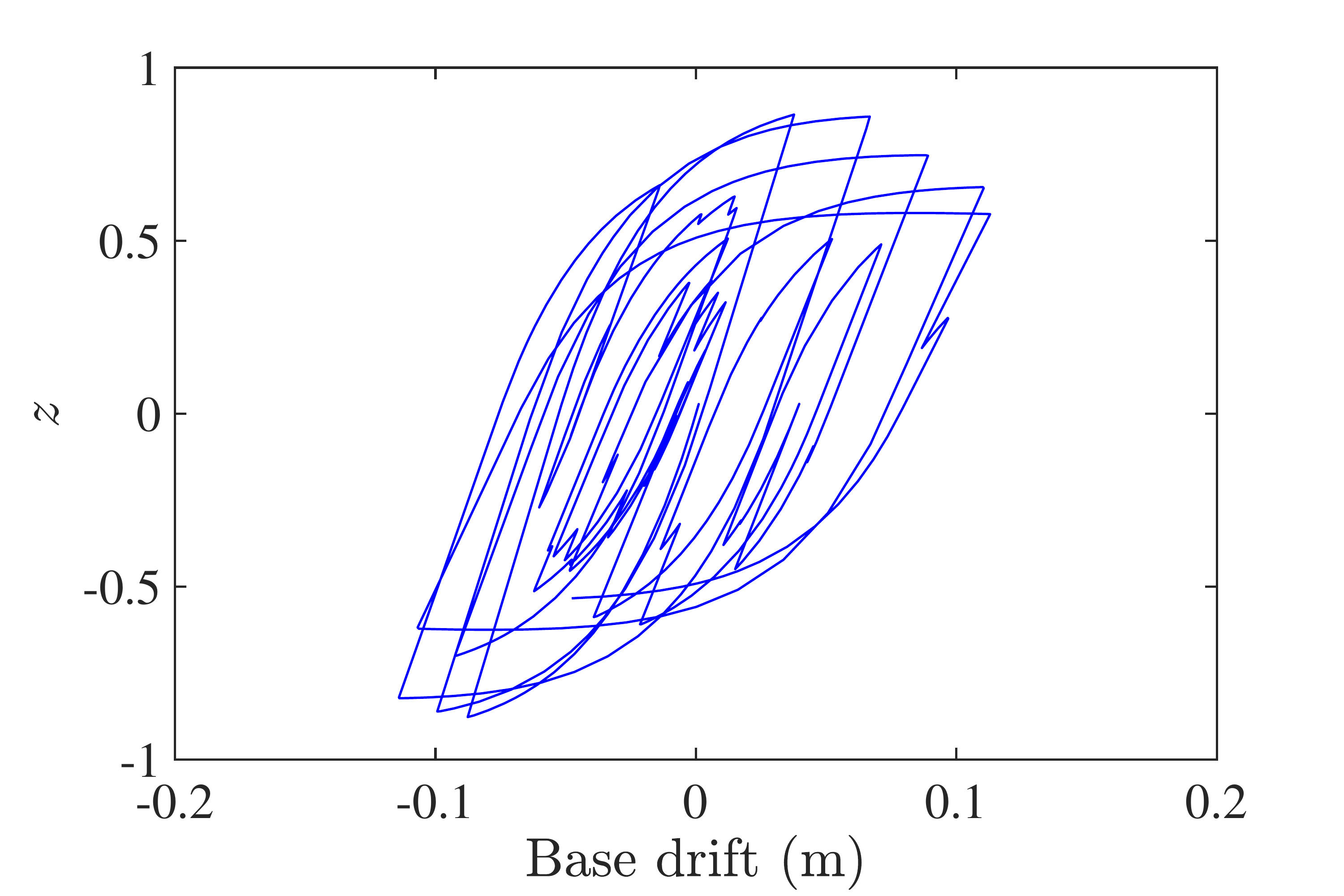}
         \caption{With degradation (see \eqref{eq:baber_wen})}
         \label{fig:degrade}
     \end{subfigure}
     \\
     \begin{subfigure}[b]{0.45\textwidth}
         \centering
         \includegraphics[width=\textwidth]{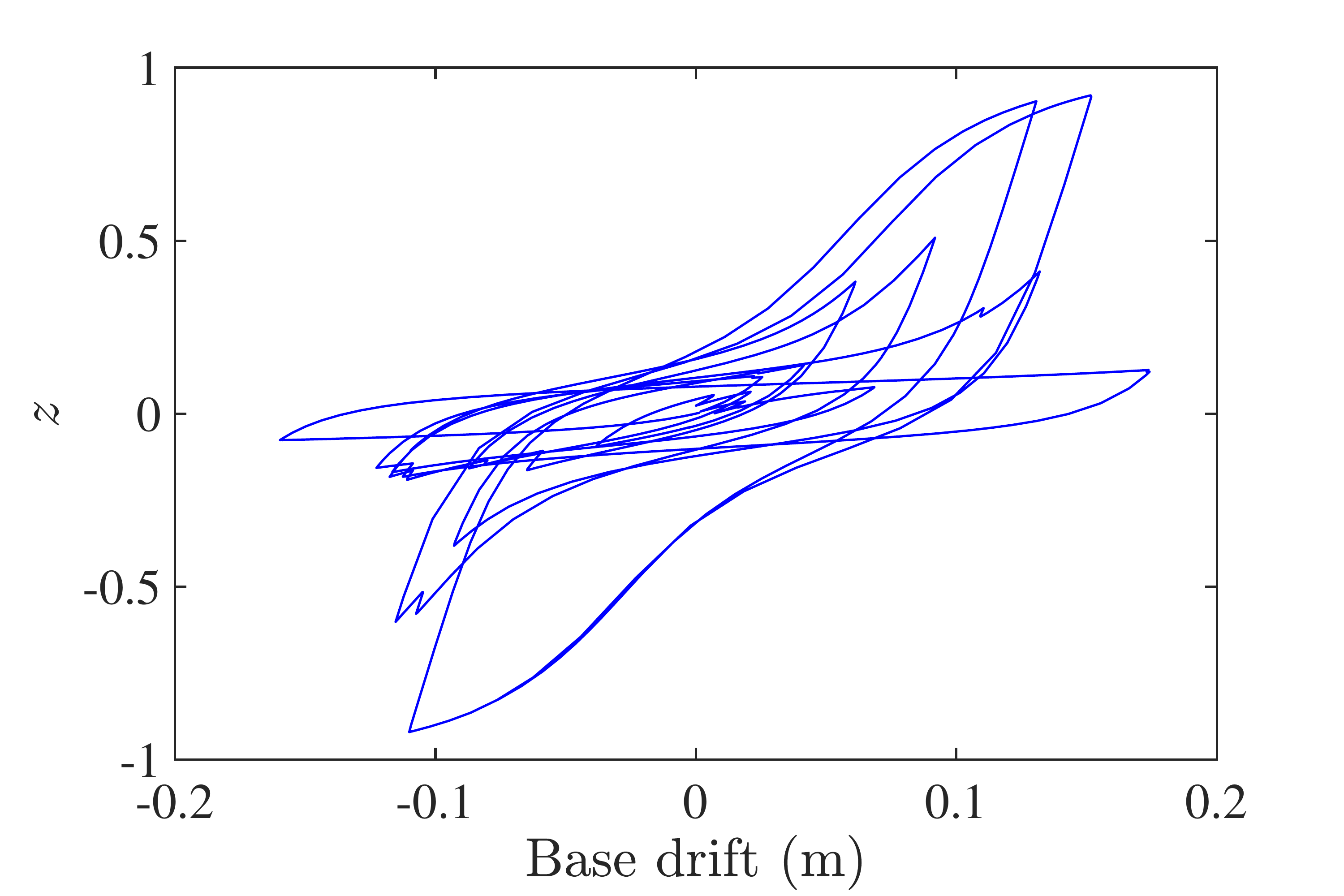}
         \caption{With pinching (see \eqref{eq:pinching})}
         \label{fig:pinching}
     \end{subfigure}
        \caption{Bouc-Wen hysteresis model and the effects of degradation and pinching.}
        \label{fig:hysteresis plots}
\end{figure}

We incorporate the effect of degradation by implementing the Baber-Wen model \cite{baber1981random,wu2008real} given by 
\begin{equation} \label{eq:baber_wen} 
\dot z = g_\mathrm{degrade}:= \frac{1}{\eta}\left[\bar{A} \dot u\subscript{b} - \nu\left(\beta \dot u\subscript{b}|z|^{\npow}+\gamma z|\dot u\subscript{b}||z|^{\npow-1}\right)\right],
\end{equation}
where the degradation shape functions are $\bar{A}$, $\nu$, and $\eta$ that depend on the degradation parameters $\delta_{A}$, $\delta_{\nu}$, $\delta_{\eta}$, and a measure of response duration and severity $e$ as follows 
\begin{equation}
\begin{split}
    e&=\int_0^tz(\tau)\dot u\subscript{b}(\tau)\mathrm{d} \tau; \\
    \bar{A}&=A(1-\delta_Ae);\\
    \nu&=1+\delta_{\nu}e; \\ 
    \eta&=1+\delta_{\eta}e. \\ 
\end{split} 
\end{equation} 
To further incorporate the effect of pinching in \eqref{eq:baber_wen} to explain a sharp reduction in the stiffness that may occur due to the opening and closing of cracks, the following equation can be used to model the evolution of $z$ \cite{ma2004parameter,ma2006system,wu2008real} 
\begin{equation} \label{eq:pinching}
    \dot{z}= g_\mathrm{pinch} := \frac{h(z)}{\eta}\left[{A} \dot u\subscript{b} - \nu\left(\beta \dot u\subscript{b}|z|^{\npow}+\gamma z|\dot u\subscript{b}||z|^{\npow-1}\right)\right],
\end{equation} 
where $h(z)$ is a pinching shape function that depends on the parameters measure of total slip $\zeta_s$, pinching slope $p$, pinching initiation $q$, pinching magnitude $\psi$, pinching rate $\delta_\psi$, and pinching severity $\lambda$ as follows 
\begin{equation}
    \begin{split}
        h(z)&=1-\zeta_1\exp\left[-\frac{(z\sgn(\dot u\subscript{b})-qz_u)^2}{\zeta_2^2}\right];\\
        z_u&=\left[{\nu(\beta+\gamma)}\right]^{-1/\npow};\\ 
        \zeta_1&=\zeta_s(1-\exp(-pe));\\
        \zeta_2&=(\psi+\delta_\psi e)(\lambda+\zeta_1).\\
    \end{split}
\end{equation} 
Figure \ref{fig:hysteresis plots} shows the effects of degradation and pinching on the Bouc-Wen model of hysteresis for the four degree-of-freedom system in Figure \ref{fig:4dof} with $\delta_A=0.6$, $\delta_\nu=0.02$, $\delta_\eta=0.02$, $\zeta_s=0.5$, $p=1.0$, $q=0.5$, $\psi=0.25$, $\delta_\psi=0.15$, and $\lambda=0.5$.
More details about the system are provided in Section \ref{sec:examples}. 

\section{Methodology} \label{sec:method} 
As discussed in the previous section, to model the degradation effect of hysteretic systems, we need to estimate many parameters. In the presence of noise, this exercise can become difficult for a full-scale structure. In this paper, instead, we try to take advantage of the recent developments in SciML to model the degradation effects. In particular, 
we develop a bi-fidelity approach for degrading hysteretic systems.  This approach treats the response of the degrading structure or system as the ``true'' or high-fidelity representation.  However, unlike typical SciML methods that attempt to model the true system directly, the proposed approach creates a model of a pristine, or non-degrading, system as a lower fidelity representation and then employs a DeepONet to describe the discrepancy between the true, high-fidelity system and this low-fidelity approximation.

To illustrate the approach, let us consider a generic nonlinear multi-degree-of-freedom mass-spring-damper system, \textit{i.e.} a system without a hysteretic base layer, whose governing equation is given by 
\begin{equation} \label{eq:mdof} 
    \Mm \ddot{\uu}(t;\xii) + \Cm \dot{\uu}(t;\xii) + \Km \uu(t;\xii) + \widetilde{\Lm} \g\left( \ddot{\uu},\dot{\uu},\uu;t,\xii \right) = \w(t), 
\end{equation}
where $\Mm\in\mathbb{R}^{n\times n}$ is the mass matrix; $\Cm\in\mathbb{R}^{n\times n}$ is the damping matrix; $\Km\in\mathbb{R}^{n\times n}$ is the stiffness matrix; $\w(t)\in\mathbb{R}^{n}$ is the external force; $\ddot{\uu}(t;\xii)\in\mathbb{R}^{n}$, $\dot{\uu}(t;\xii)\in\mathbb{R}^{n}$, and $\uu(t;\xii)\in\mathbb{R}^{n}$ are acceleration, velocity, and displacement vectors of the system, respectively; and $\xii$ is the uncertain parameter vector. The nonlinearity in the system is expressed in the term $\g\left( \ddot{\uu},\dot{\uu},\uu;t,\xii \right)\in\mathbb{R}^{n_g}$ with an influence matrix $\widetilde\Lm\in\mathbb{R}^{n\times n_g}$. 
With the hysteretic layer, the governing equation of the system can be expressed in state-space form as follows 
\begin{equation}\label{eq:state-space}
\begin{split}
\dot{\X}(t;\xii) &= \Am \X(t;\xii) + \Bm \w(t) + \Lm \g\left(\X,\dot{\X};t.\xii\right) + \Lm_\mathrm{h}g_\mathrm{hyst}(t;\xii), \quad \X(0) = \x_0;\\
\Y(t;\xii) &= \Cm \X(t;\xii) + \Dm \w(t) + \Em\g\left(\X,\dot{\X};t,\xii\right) +\Em_\mathrm{h}g_\mathrm{hyst}(t;\xii) , \\
\end{split}
\end{equation}
where $\X(t;\xii):=\left\{\begin{array}{cc}
     \dot \uu(t;\xii)  \\
     \uu(t;\xii) \\
     z(t;\xii)
\end{array}\right\}\in\mathbb{R}^{(2n+1)}$ is the state vector; $\Am:=\in \mathbb{R}^{(2n+1)\times (2n+1)}$ is the state matrix; $\Bm\in\mathbb{R}^{(2n+1)\times n}$ is the influence matrix for $\w(t)$;  
$g_{\mathrm{hyst}}(t;\xii)$ is the nonlinear hysteretic function; $\Lm\in\mathbb{R}^{(2n+1)\times (\ngm+1)}$ is the influence matrix for $\g(\cdot,\cdot;t,\xii)$; $\x_0$ is the initial state vector; $\Y(t;\xii)\in\mathbb{R}^{\ny}$ denotes the output; and  $\Cm\in\mathbb{R}^{\ny\times n}$, $\Dm\in\mathbb{R}^{\ny\times \nw}$, and $\Em \in \mathbb{R}^{\ny\times \ngm}$ denote the influence matrices for the state vector $\X(t;\xii)$, external force $\w(t)$, and the uncertain and nonlinear function $\g(\cdot,\cdot;t,\xii)$, respectively. Further, note that $g_{\mathrm{hyst}}(t;\xii)$ is represented by the right hand side term in either $g_{\!_\mathrm{BW}}(t;\xii)$, $g_\mathrm{degrade}(t)$, or $g_\mathrm{pinch}(t;\xii)$ in the hysteretic model equations \eqref{eq:bouc}, \eqref{eq:baber_wen}, or \eqref{eq:pinching}, respectively. Additional nonlinearities beyond those captured by the hysteretic functions could be incorporated into the general function $g(t;\xii)$.

A low-fidelity representation of a fully degrading hysteretic system can be constructed by using the original Bouc-Wen hysteresis model in \eqref{eq:bouc} that ignores the effects of degradation. Hence, in this case, 
the corresponding low-fidelity governing equations can be given by 
\begin{equation}\label{eq:LF_state-space}
\begin{split}
\dot{\x}(t;\xii) &= \Am \x(t;\xii) + \Bm \w(t) + \Lm \g\left(\x,\dot{\x};t,\xii\right) + \Lm_\mathrm{h}g_{\!_\mathrm{BW}}(t;\xii), \quad \x(0) = \x_0;\\
\y(t;\xii) &= \Cm \x(t;\xii) + \Dm \w(t) + \Em\g\left(\x,\dot{\x};t,\xii\right) + \Em_\mathrm{h}g_{\!_\mathrm{BW}}(t,\xii), \\
\end{split} 
\end{equation} 
where $\x$ and $\y$ are low-fidelity state and output, respectively. 
Therefore, a discrepancy or correction term $\y_\mathrm{corr}(t;\xii)$, \textit{i.e.}, the difference between the low-fidelity and true output, can be described by 
\begin{equation}\label{eq:y_c}
    \y_\mathrm{corr}(t;\xii) = \Y(t;\xii)-\y(t;\xii). 
\end{equation} 
In this paper, it is assumed that the modeling the discrepancy term $\y_\mathrm{corr}(t;\xii)$ is computationally easier than modeling the response of the original, fully degrading hysteretic system $\Y(t;\xii)$. A DeepONet is employed for this task and within this DeepONet the low-fidelity response is treated as the input such that 
\begin{equation}
\begin{split}
\y_\mathrm{corr}(t;\xii)&\approx G_{\thetaa}(\y)(t,\xii)\\
&=    c_0 + \sum_{i=1}^p c_i(\y(t_1,\xii_1),\dots,\y(t_m,\xii_m))\psi_i(t,\xii).\\ 
\end{split} 
\end{equation}



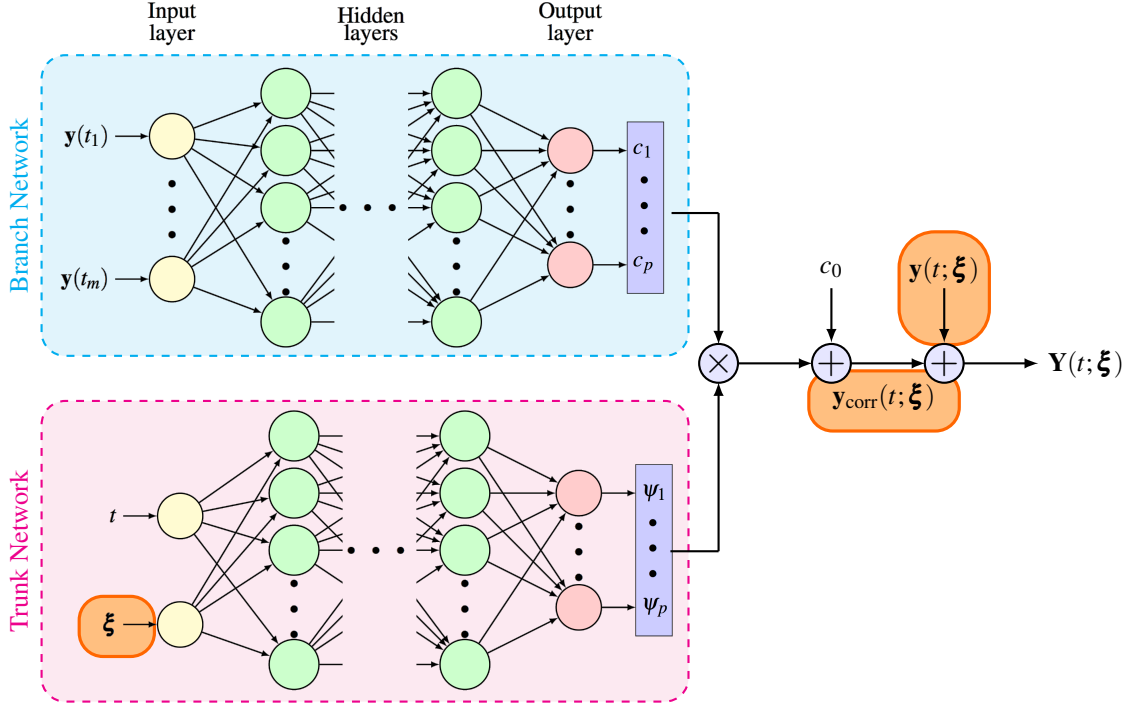
\begin{figure}[!htb]
        \centering
		\begin{tikzpicture}
\draw[draw=cyan,fill=cyan!10,thick,dashed,rounded corners=2ex] (-5,-2.4) rectangle ++(8.6,4); 
\node[rotate=90,color=cyan] at (-5.3,-0.4) {Branch Network}; 
\draw[draw=magenta,fill=magenta!10,thick,dashed,rounded corners=2ex] (-5,-7) rectangle ++(8.6,4); 
\node[rotate=90,color=magenta] at (-5.3,-5) {Trunk Network}; 
\draw[draw=red!20!orange,fill=orange!50,very thick,rounded corners=3ex] (6.4,-2.25) rectangle ++(1.2,1.5); 
\draw[draw=red!20!orange,fill=orange!50,very thick,rounded corners=2ex] (5.2,-3.4) rectangle ++(2,0.8); 
\draw[draw=red!20!orange,fill=orange!50,very thick,rounded corners=2ex] (-4.5,-6.4) rectangle ++(1,0.8); 
	\node[inner sep=0pt] (russell) at (-0.68,0)
    {\includegraphics[scale=0.108]{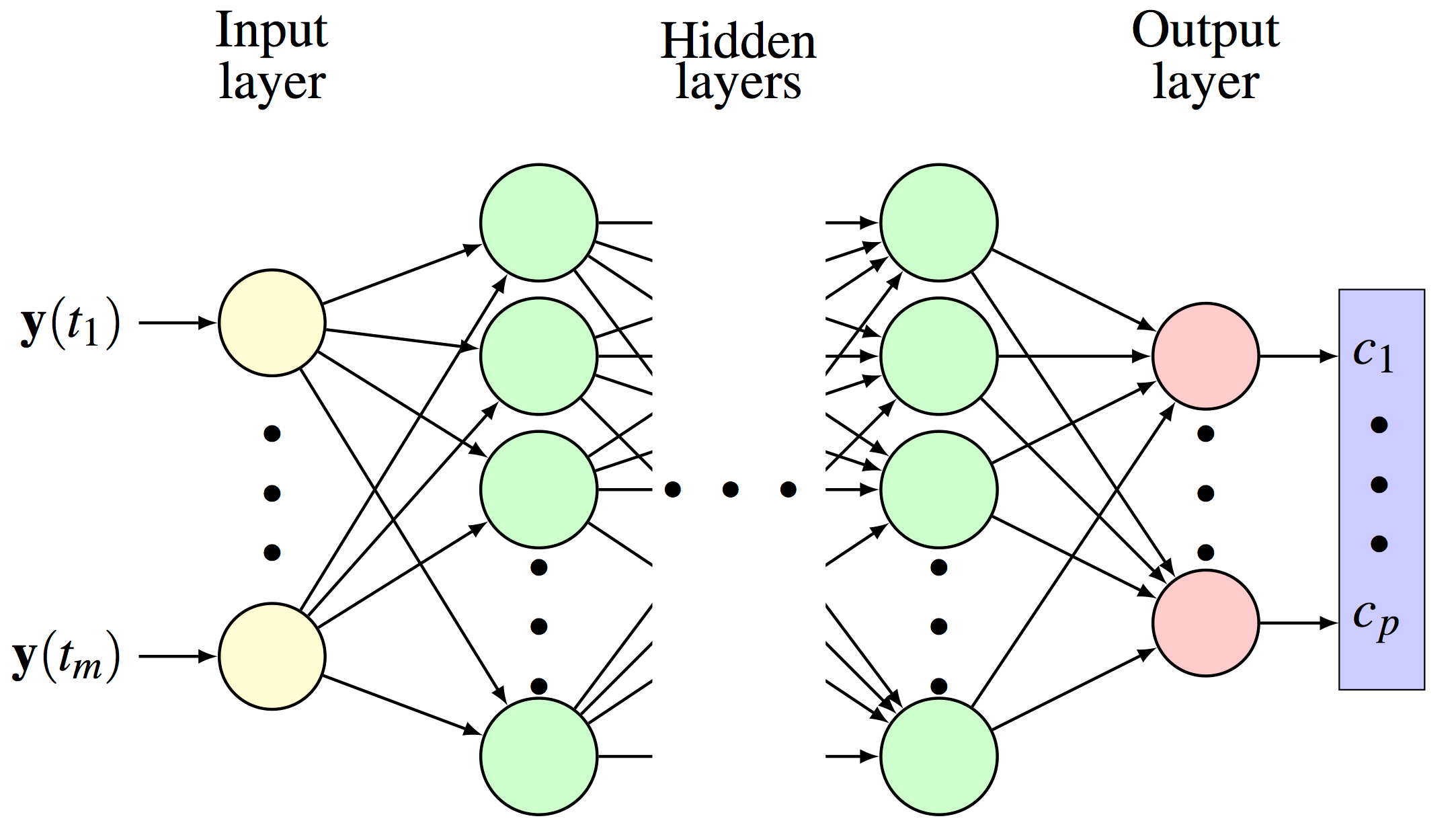}};
    \node[inner sep=0pt] (russell) at (-0.4,-5)
    {\includegraphics[scale=0.108]{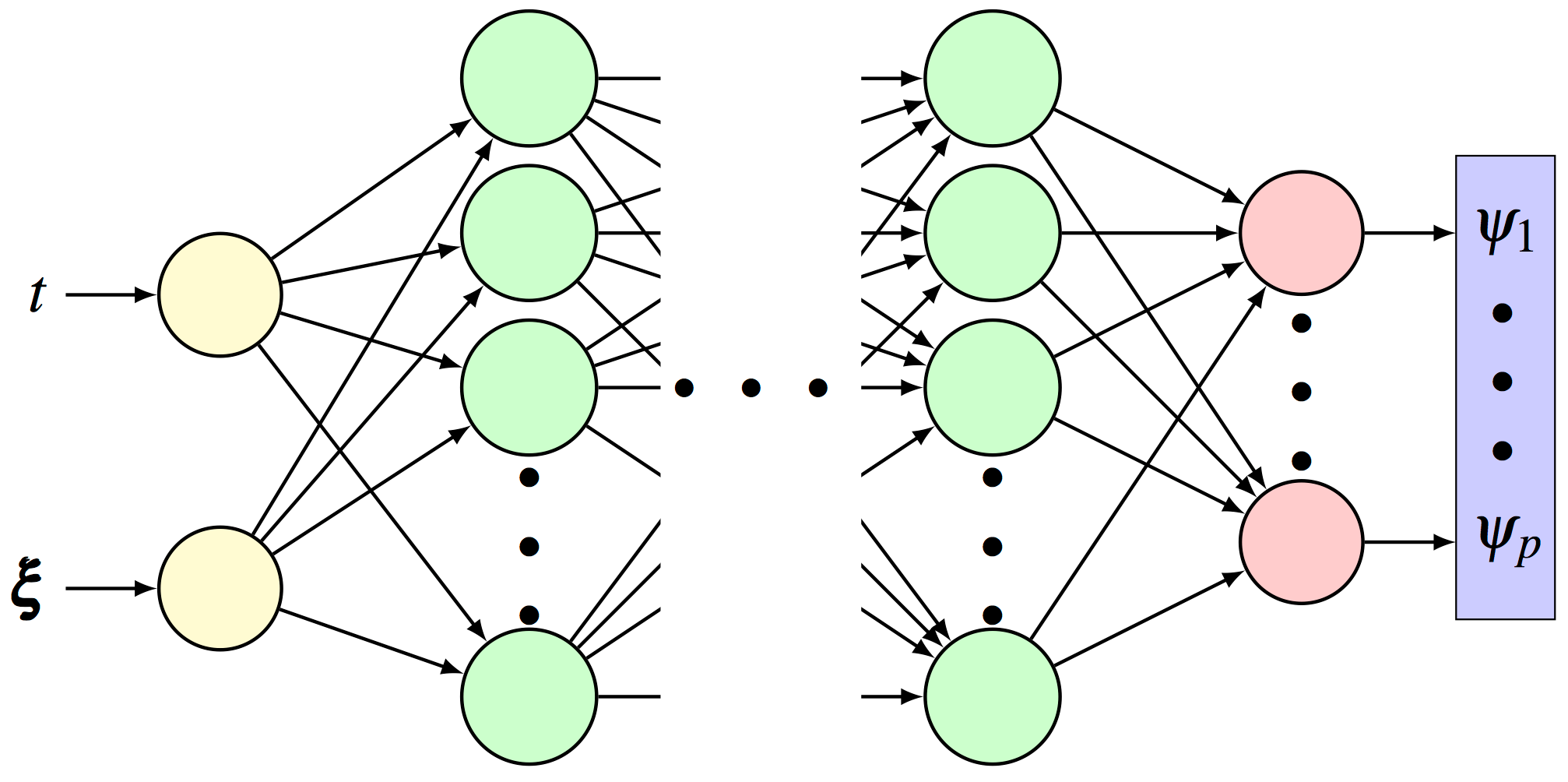}};
    \draw [thick] (3.375,-0.5) -- (4,-0.5); 
    \draw [-latex,thick] (4,-0.5) -- (4,-2.25); 
    \draw [thick] (3.36,-5) -- (4,-5); 
    \draw [-latex,thick] (4,-5) -- (4,-2.75); 
    \draw[-latex,thick] (4,-2.5) -- (5.25,-2.5); 
    \draw[-latex,thick] (5.5,-1.5) -- (5.5,-2.225); 
    \draw[-latex,thick] (7,-1.5) -- (7,-2.225); 
    \node[draw=black,thick,circle,fill=blue!10,inner sep=0pt,minimum size=6pt] () at (4,-2.5) {\Large$\times$}; 
    \draw [-latex,thick] (5.75,-2.5) -- (6.75,-2.5); 
    \node [above] at (6.2,-3.25) {$\y_\mathrm{corr}(t;\xii)$}; 
    \node[draw=black,thick,circle,fill=blue!10,inner sep=0pt,minimum size=6pt] () at (7,-2.5) {\Large$+$}; 
    \node[draw=black,thick,circle,fill=blue!10,inner sep=0pt,minimum size=6pt] () at (5.5,-2.5) {\Large$+$}; 
    \node[] at (5.5,-1.25) {$c_0$}; 
    \node[] at (7,-1.25) {$\y(t;\xii)$}; 
    \draw [-latex,thick] (7.25,-2.5) -- (8.25,-2.5) node [right] {$\Y(t;\xii)$}; 
    
    \end{tikzpicture}
    \caption{DeepONet used for bi-fidelity modeling of degrading hysteresis systems in Section \ref{sec:method} with differences from the standard DeepONet architecture presented in Section \ref{sec:DeepONet_back} shown in solid (orange) boxes. }
    \label{fig:bf_approach}
\end{figure}

\section{Numerical Examples} \label{sec:examples} 
In this section, we illustrate the proposed approach for using DeepONets to approximate the discrepancy between the response of a degrading hysteretic system and its corresponding low-fidelity representation through three numerical examples. These examples utilize two separate datasets, namely a training dataset $\Dtr=\left\{(t_i,\xii_i),\Y(t_i;\xii_i),\{\y^{(i)}(t_j;\xii_j)\}_{j=1}^m\right\}_{i=1}^{\Ntr}$ and a validation dataset $\Dval=\left\{(t_i,\xii_i),\Y(t_i;\xii_i),\{\y^{(i)}(t_j;\xii_j)\}_{j=1}^m\right\}_{i=1}^{\Nval}$. 
The DeepONets are first trained using $\Dtr$ and then their accuracy is measured via the validation dataset $\Dval$.  The relative root mean squared error (RMSE), $\epsv$, is the chosen validation error metric and is defined as follows 
\begin{equation} \label{eq:val_rmse} 
    \epsv = \frac{\lVert \Y_\mathrm{pred} - \Y_\mathrm{val} \rVert_2}{\lVert\Y_\mathrm{val}\rVert_2}, 
\end{equation} 
where $\lVert\cdot\rVert_2$ is the $\ell_2$-norm of its argument, $\Y_\mathrm{pred}$ is the vector of predicted response using DeepONets, and $\Y_\mathrm{val}$ is original response from the validation dataset $\Dval$. 


\subsection{Example 1: Four Degree-of-freedom System} 
In our first example, we use the four degree-of-freedom system shown in Figure \ref{fig:4dof} subjected to base excitation. The governing equation of the system is given by \begin{equation} \label{eq:4dof} 
\begin{split}
    &\Mm \ddot{\uu}(t;\xii) + \Cm \dot{\uu}(t;\xii) + \Km \uu(t;\xii) = -\Mm\mathbf{1} \ddot{u}_g(t)+\Cm\mathbf{1}\dot{u}_b+\Km\mathbf{1}u_b,  \\
    &m_b\ddot{u}_b + \mathbf{1}^T\Cm\mathbf{1}\dot{u}_b +\mathbf{1}^T\Km\mathbf{1}{u}_b + f_b = -m_b\ddot{u}_g +\mathbf{1}^T\Cm\mathbf{1}\dot{u}_b
\end{split}
\end{equation} 
where $\Mm\in\mathbb{R}^{3\times3}$ is the mass matrix with $m_1=m_2=m_3=300$ Mg; $\Km\in\mathbb{R}^{3\times3}$ is the stiffness matrix with $k_1=k_2=k_3=40$ MN/m; $\mathbf{1}$ is a column vector of consisting of three ones; $\uu=[u_1,u_2,u_3]^T$ is the vector of displacement relative to the ground; Rayleigh damping is used with 3\% damping ratio for the first two modes (\textit{i.e.}, $\Cm=\beta_1\Mm+\beta_2\Km$, where $\beta_1$ and $\beta_2$ are estimated from the damping ratios and natural frequencies of the superstructure.); $\ddot{u}_g$ is the base excitation; and the base mass is $m_b=500$ Mg. The parameters $c_b,\kpost,\rk,$ and $\Qy$ are assumed to be uncertain and are distributed according to the distributions described in Table \ref{tab:ExI_param}. 

\begin{table}[!htb]
    \centering 
    \begin{threeparttable}[!htp]
		\caption{Probability distributions of the uncertain parameters in Example I. } \label{tab:ExI_param} 
		\begin{tabular}{c c c c c c c} 
			\hline 
			\Tstrut
			  Parameter & \multirow{2}{*}{Distribution} & \multirow{2}{*}{Mean} & \multirow{2}{*}{Std. Dev.} & Lower  & Upper \\ 
            (Unit) & & & & bound & bound \\[0.5ex] 
			\hline 
			\Tstrut
			 $\kpost$ [MN/m] & Lognormal & 4 & 0.25 & 0 & $\infty$\\ 
			 $c_\mathrm{b}$ [MN$\cdot$s/m] & Lognormal & 20 & 4 & 0 & $\infty$\\
			 $r_k$ & Uniform & 0.16 & 0.00577 & 0.15 & 0.17 \Bstrut\\ 
			 $\Qy$ (\%)$^\dagger$ & Uniform & 5.0 & 0.57735 & 4.0 & 6.0 \\ [1ex] 
			\hline 
		\end{tabular}
\begin{tablenotes}
			\item[$^\dagger$] in \% of the total weight of the structure.
		\end{tablenotes}
	\end{threeparttable}
    \label{tab:ex1_param}
\end{table}

\begin{figure}[!htb]
    \centering
    \includegraphics[scale=0.3]{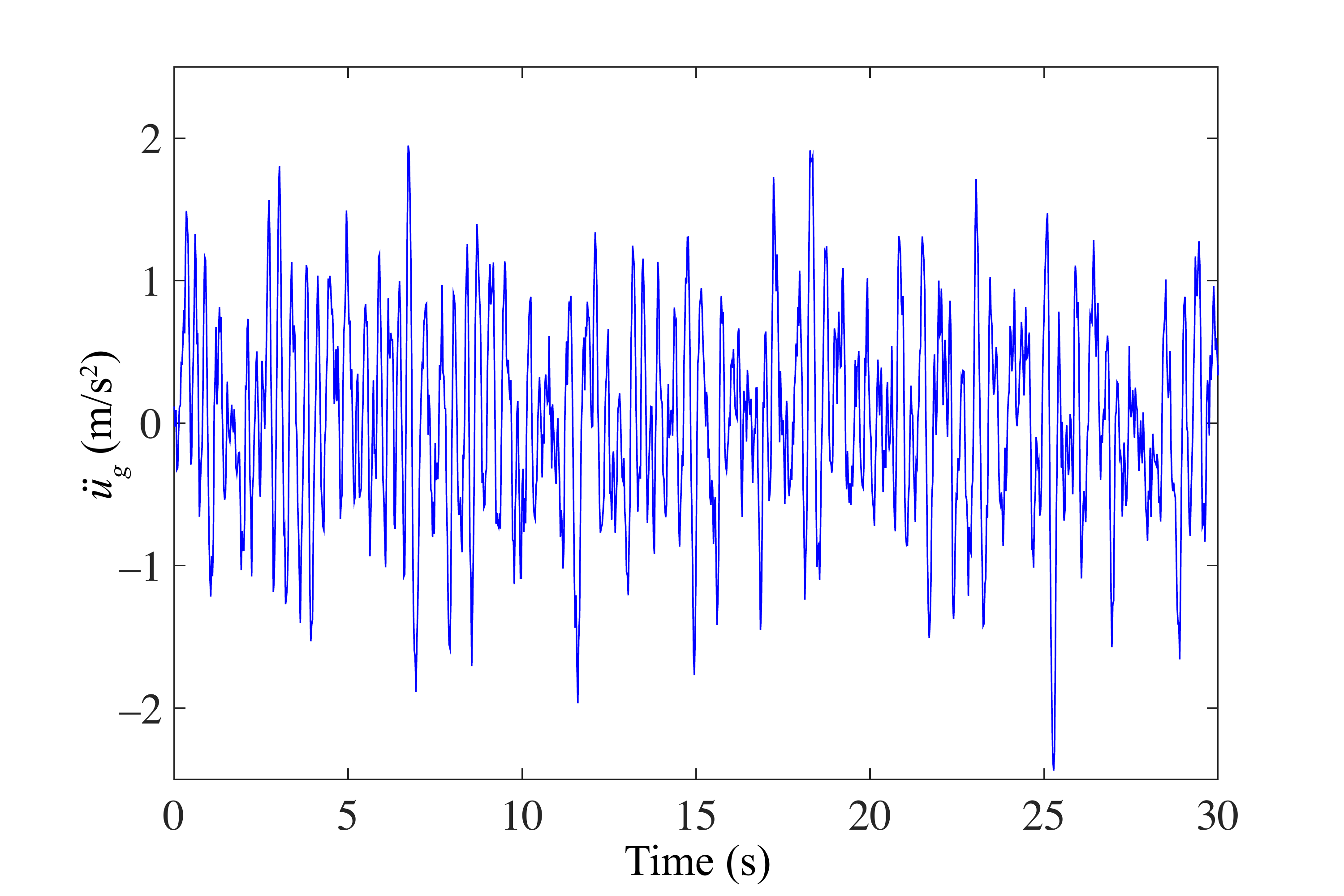}
    \caption{Time history realization of $\ddot{u}_g$ used in Example 1.}
    \label{fig:KT_exc}
\end{figure}

A sample realization of the base excitation applied to the structure is shown in Figure \ref{fig:KT_exc}.  The realizations for the excitation are obtained as a stationary filtered white noise from a Kanai-Tajimi filter \cite{spencer1998benchmark} with spectral density \cite{ramallo2002smart} 
\begin{equation}
    S_{\ddot{u}_g\ddot{u}_g}(\omega)=\frac{S_0\left( 4\zeta^2_g\omega_g^2\omega^2+\omega_g^4 \right)}{\left( \omega^2-\omega^2_g \right)^2 + 4\zeta_g^2\omega_g^2\omega^2}, 
\end{equation}
where $\omega_g=17$ rad/s and $\zeta_g=0.3$. The spectral intensity $S_0$ is assumed as 
\begin{equation}
    S_0=\sigma_w^2\frac{0.03\zeta_g}{\pi\omega_g\left( 4\zeta_g^2+1 \right)}g^2, 
\end{equation}
where $g$ is the gravitational acceleration and $\sigma_w=2$ is selected following De et al. \cite{de2019hybrid}. 

\subsubsection{Results} \label{sec:ex1_results}

We use a DeepONet with a branch network consisting of three hidden layers with 50 neurons in each layer and a trunk network consisting of two hidden layers with 50 neurons each. For the activation function, we use the ELU (exponential linear unit) \cite{clevert2015fast}. In the DeepONet expression \eqref{eq:deep}, we use $p=8$. The number of hidden layers in the trunk and branch networks as well as the number of neurons per hidden layer are selected using an iterative procedure, where the number of neurons per hidden layer is gradually increased to an upper limit (50 in this example).  Hidden layers are then added to the network until no reduction of $\epsv$ is observed. For the number of terms $p$ in \eqref{eq:deep}, we increase $p$ by two to its maximum value, which is the same as the number of neurons in the hidden layers, to observe any reduction of $\epsv$. In this example, we explore two cases, namely the original (high-fidelity) system with only degradation (case I) and with both degradation and pinching (case II), to compare the performance of the bi-fidelity approach against a standard DeepONet. Results from these two cases are described next. 

\begin{figure}
    \centering
     \begin{subfigure}[b]{0.45\textwidth}
         \centering
         \includegraphics[scale=0.275]{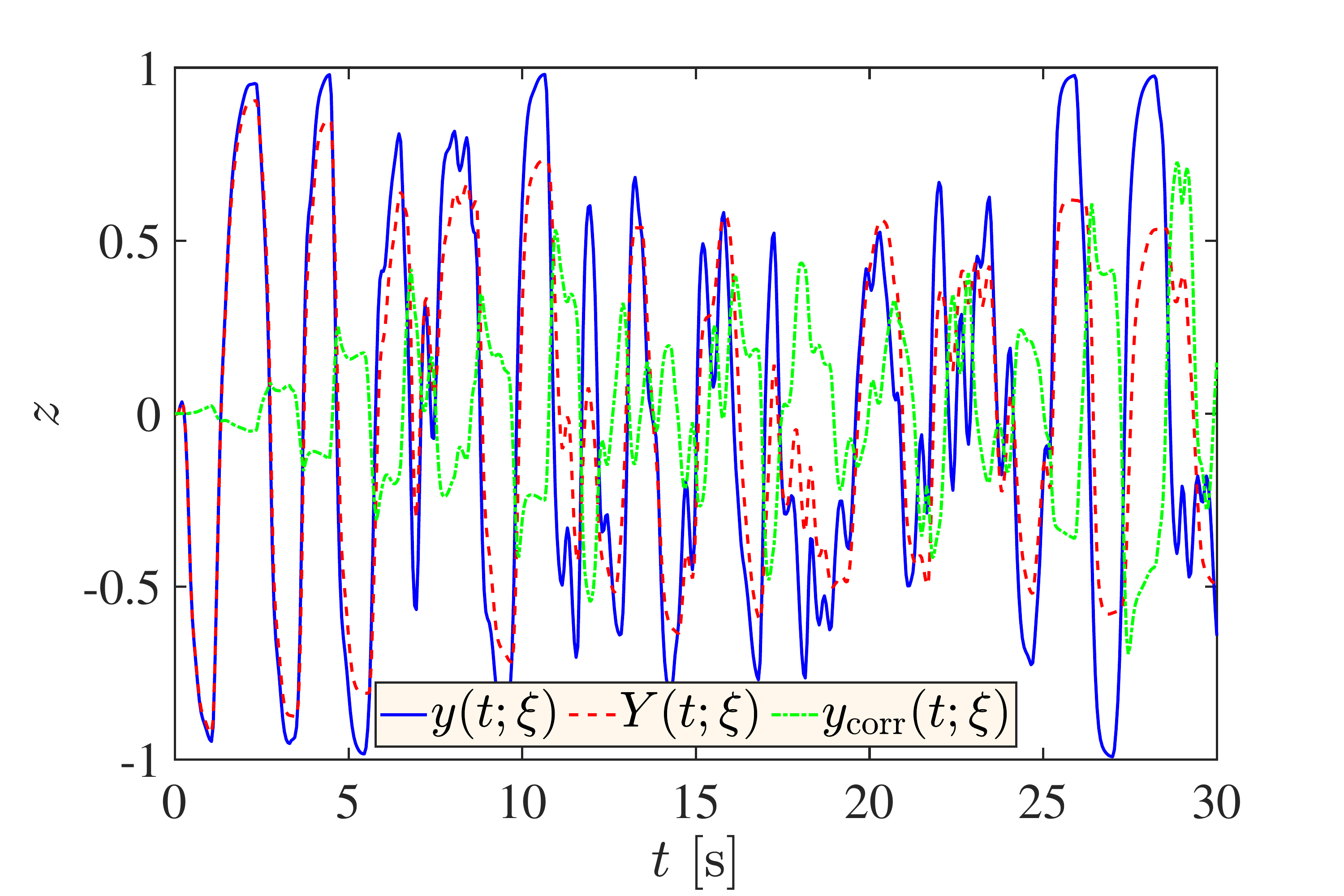}
         \caption{Auxiliary variable $z$} 
         \label{fig:ExI_z_Degrade}
     \end{subfigure}
     \hfill
     \begin{subfigure}[b]{0.45\textwidth}
         \centering
         \includegraphics[scale=0.275]{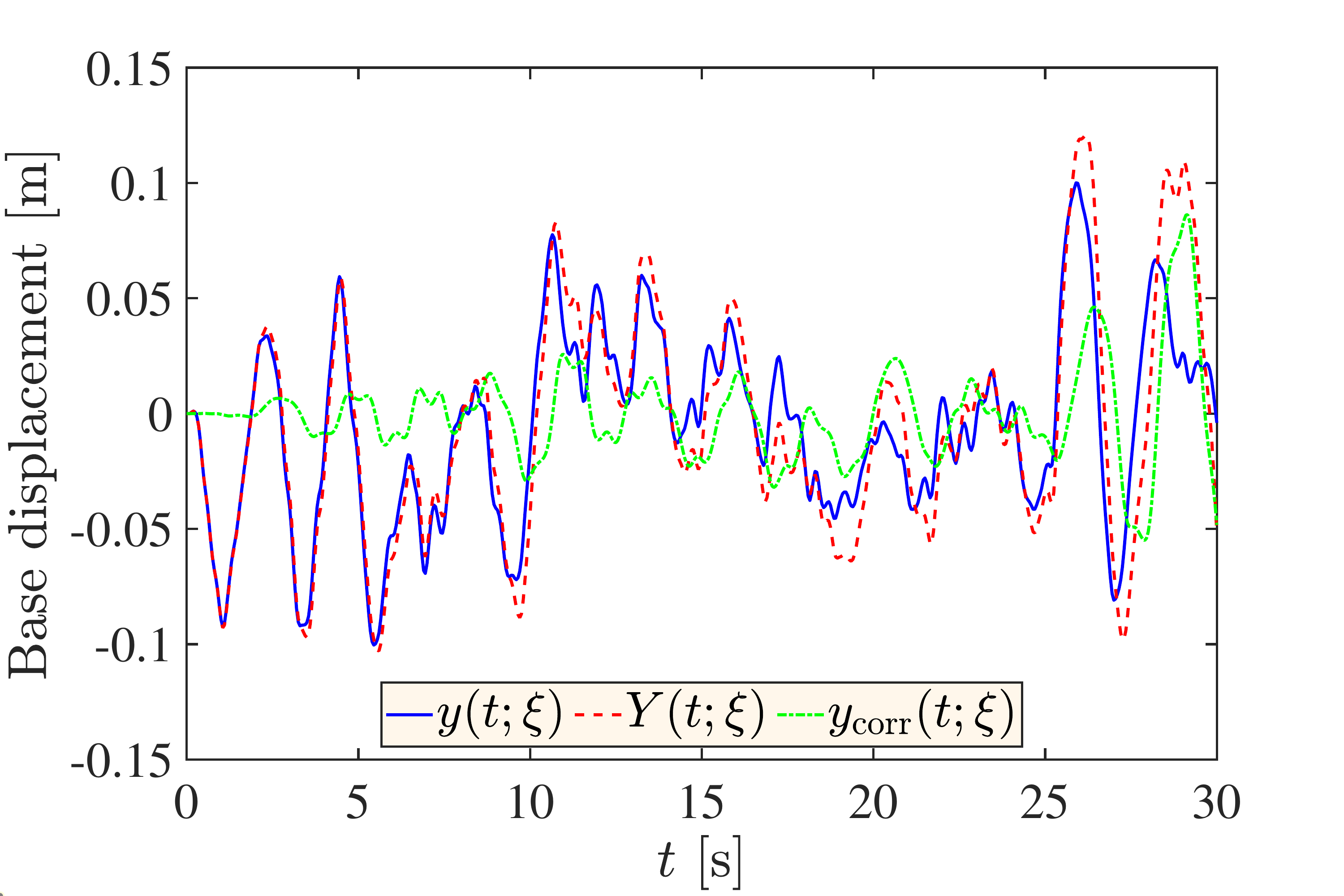}
         \caption{Base displacement}
         \label{fig:ExI_base}
     \end{subfigure}
     \\
     \begin{subfigure}[b]{0.45\textwidth}
         \centering
         \includegraphics[scale=0.275]{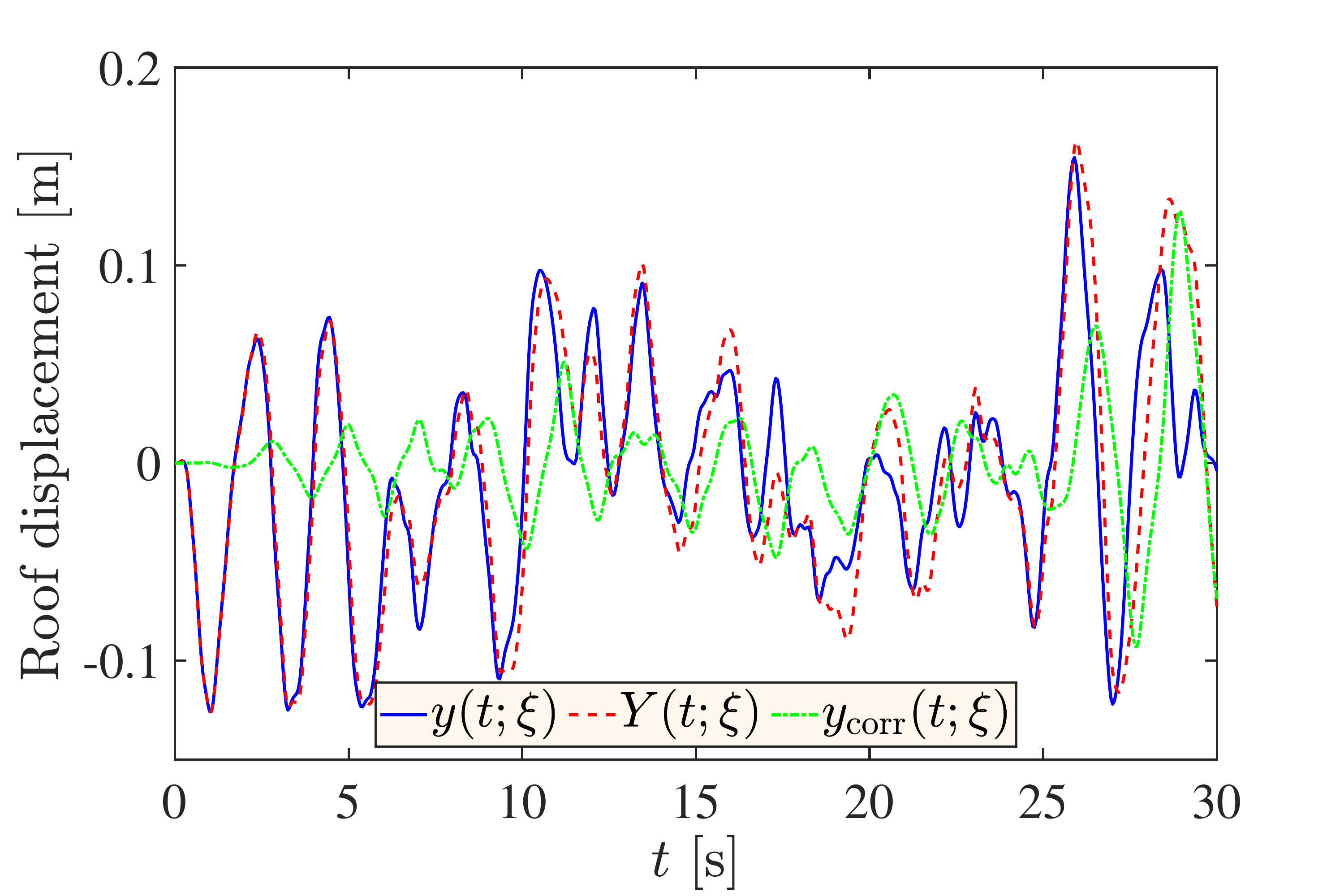}
         \caption{Roof displacement}
         \label{fig:ExI_roof}
     \end{subfigure}
    \caption{Comparison of responses realizations from the low- and high-fidelity model in Example 1. 
    Responses from the standard Bouc-Wen model (blue solid line) are treated as the low-fidelity model, $y(t;\xii)$. The high-fidelity model, $Y(t;\xii)$, is the Baber-Wen model that considers degradation (red dash-dotted line). The discrepancy between the models, or correction term,  $y_\mathrm{corr}(t;\xii)$ (green dashed line) is then modeled using a DeepONet. }
    \label{fig:ExI_Response_Comp}
\end{figure}

\textbf{Case I:} This case adopts the standard Bouc-Wen formulation in \eqref{eq:bouc} as the low-fidelity model and the degrading effect in \eqref{eq:baber_wen} as the high-fidelity model with parameters specified in Section \ref{sec:hyst}. Typical realizations of the auxiliary (evolutionary) variable $z$, base displacement $\ub$, and roof displacement $u_3$ from these two models and the discrepancies between them are shown in Figure \ref{fig:ExI_Response_Comp}.  Notably, the time histories of the base displacement and roof displacement exhibit highly similar behavior and, thus, similar discrepancies, wherein the discrepancy is relatively small at the beginning of the time history but gradually grows to be nearly as large as the model responses by the end. The plot for the auxiliary variable evinces many of these traits, as well.

The network is trained using dataset $\Dtr$ with $\Ntr=200$ response time histories and validated using dataset $\Dval$ with $\Nval=250$ response time histories, where the each one of the quantities of interest (QoI) $z$, $\ub$, and $u_3$ is separately employed as the training and validation response data.  For training, we utilize the Adam algorithm \cite{kingma2014adam} for 10,000 epochs with a learning rate of $10^{-3}$, which is halved every 2500 epochs.  Note that the initial value of the learning rate is selected based on results from some preliminary runs. 

\begin{figure}
    \centering
     \begin{subfigure}[b]{0.45\textwidth}
         \centering
         \includegraphics[scale=0.275]{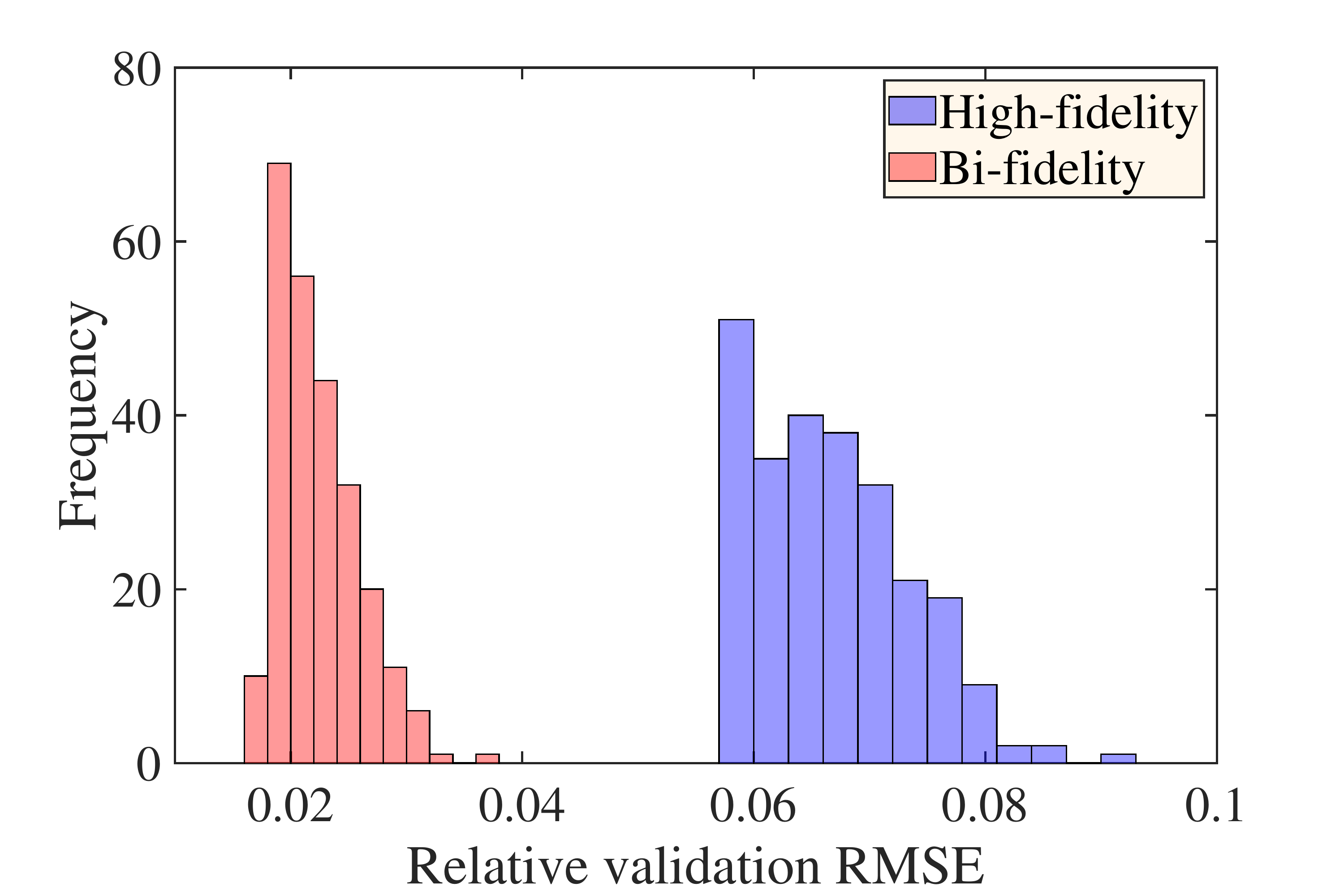}
         \caption{Auxiliary variable $z$} 
         \label{fig:ExI_Hist_Degrade}
     \end{subfigure}
     \hfill
     \begin{subfigure}[b]{0.45\textwidth}
         \centering
         \includegraphics[scale=0.275]{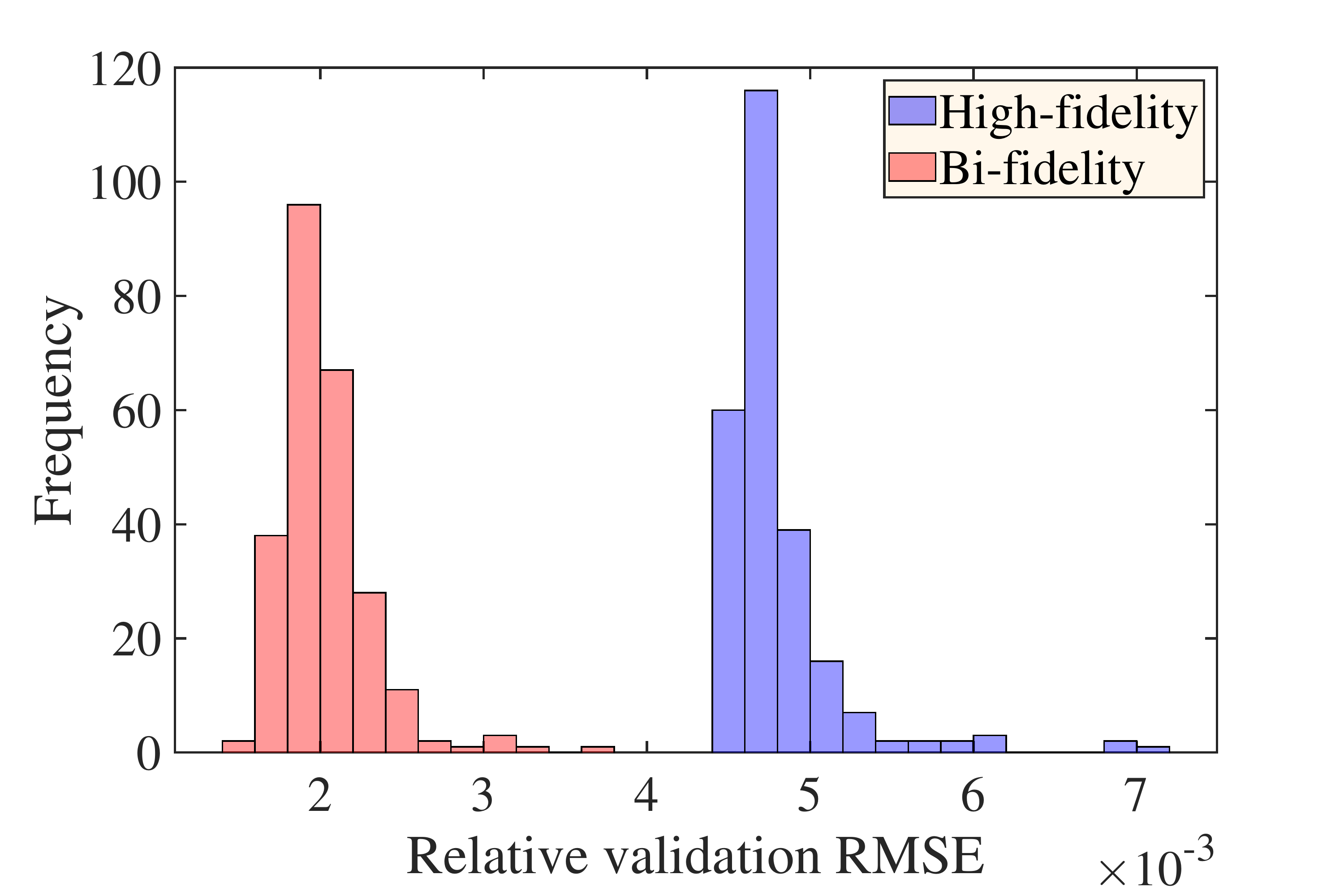}
         \caption{Base displacement $u_b$}
         \label{fig:ExI_Hist_Degrade2}
     \end{subfigure}
     \\
     \begin{subfigure}[b]{0.45\textwidth}
         \centering
         \includegraphics[scale=0.275]{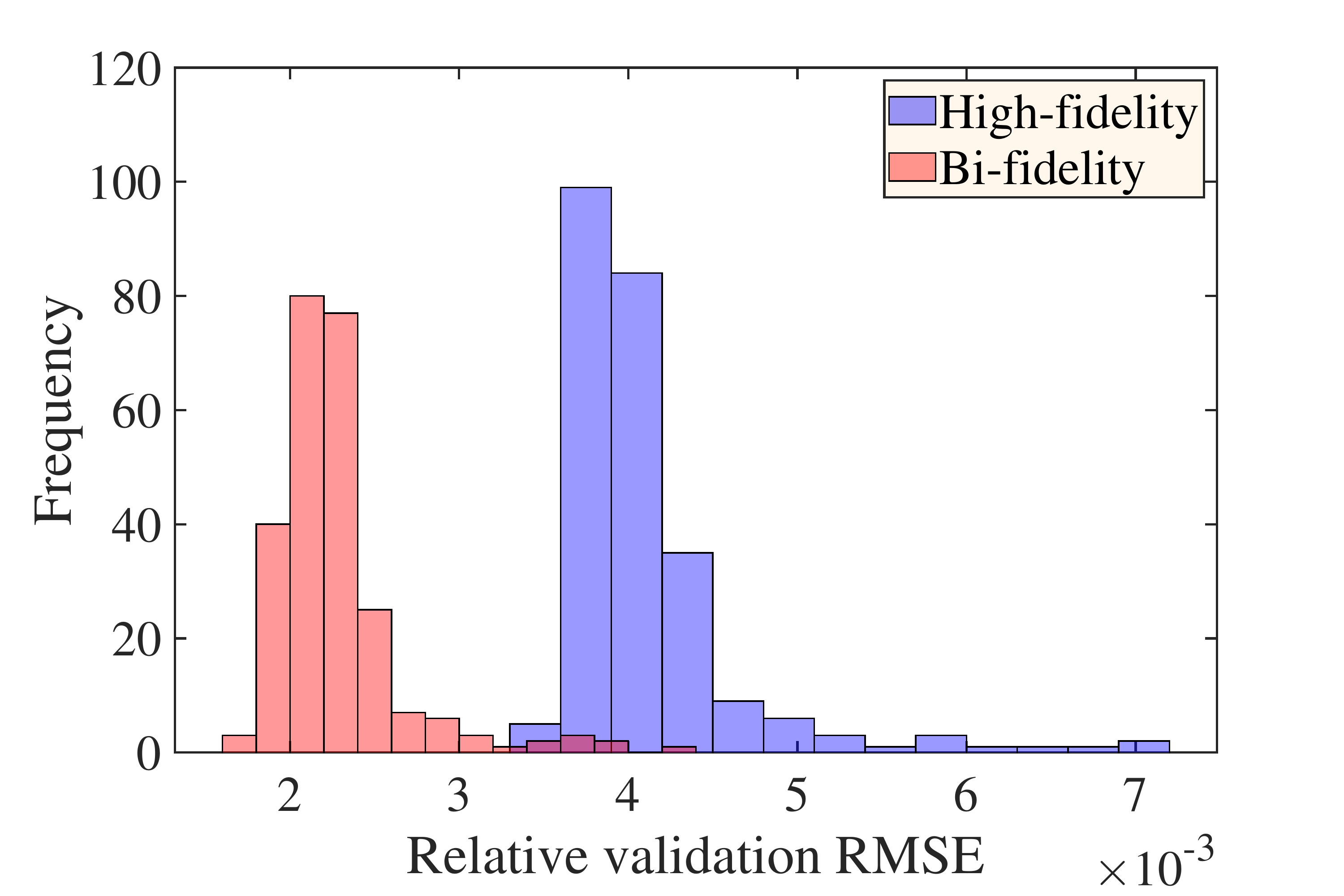}
         \caption{Roof displacement $u_3$}
         \label{fig:ExI_Hist_Degrade3}
     \end{subfigure} 
    \caption{Comparison of histograms of the relative validation error from the DeepONets trained using only the degrading Baber-Wen model (high-fidelity) and using the the bi-fidelity approach, with Bouc-Wen model as the low-fidelity model in Case I of Example 1.}
    \label{fig:Histogram_Comp_Degrade}
\end{figure}

Histograms of the relative validation errors in predicting $z$, $\ub$, and $u_3$ are shown in Figure \ref{fig:Histogram_Comp_Degrade}. These histograms are compared against those of the relative validation errors derived using a standard DeepONet \cite{lu2019deeponet} that is trained using the same number of data points but with only data from the high-fidelity model.  The apparent similarity in the discrepancies for the base and roof displacements partially manifests in Figures~\ref{fig:ExI_Hist_Degrade2} and \ref{fig:ExI_Hist_Degrade3}, as they present comparable magnitudes for their relative validation RMSE.  However, the difference in performance between the proposed bi-fidelity DeepONet and the standard implementation is more pronouced for the base displacement QoI, as the histograms for the roof displacement exhibit a noticeable degree of overlap.  While the difference in performance is largest when using auxiliary variable $z$, the relative validation RMSE errors for that QoI are one order of magnitude larger than those from using the base and roof displacement QoIs for both DeepONet implementations.

\begin{figure}[!htb]
    \centering
    \includegraphics[scale=0.275]{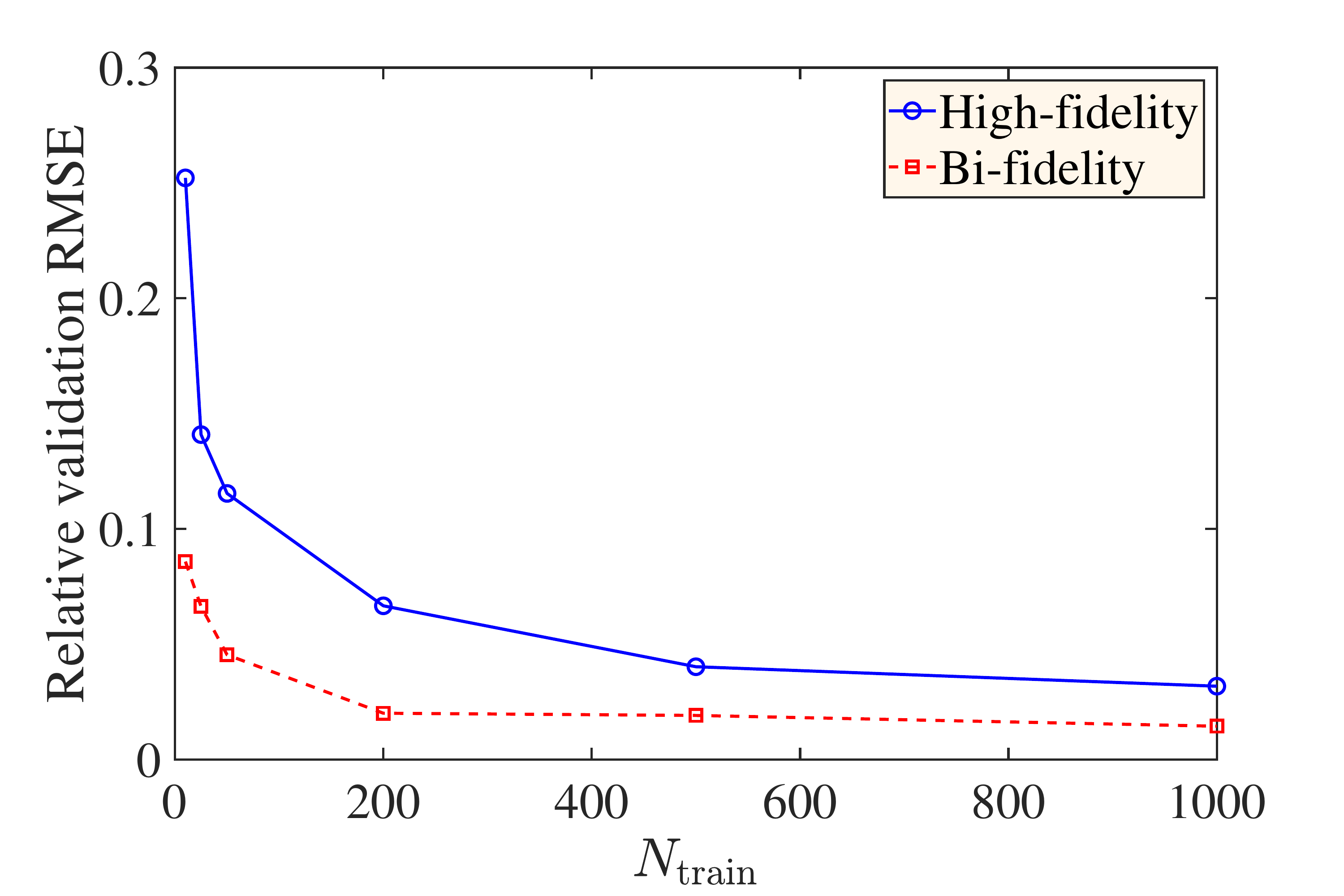}
    \caption{Relative validation error in predicting the auxiliary variable $z$ vs. number of training data points in Example 1. }
    \label{fig:Error_vs_Ndata}
\end{figure}

Table \ref{tab:ExI_results} presents the mean values of the relative validation errors, illustrating how the proposed approach reduces the error for each of these three QoIs by a factor of two to three, with the largest improvement coming when $z$ is used for training and validation. Further, Figure \ref{fig:Error_vs_Ndata} highlights the advantage of the bi-fidelity method compared to the standard ``high-fidelity only'' approach for modeling $z$.  The curves in this figure demonstrate that the advantage of the bi-fidelity approach is most pronounced when the number of training data points is small, \textit{i.e.}, when access to training data is limited. As the number of data points and access to training data increases this advantage gradually lessens, which is to be expected from any bi-fidelity approach. 

\begin{table}
\centering
		\caption{Mean of the relative validation RMSE, $\varepsilon_\mathrm{val}$, for degraded hysteretic system with $\Ntr=200$ in Example I. } \label{tab:ExI_results} 
		\begin{tabular}{c c c c c c c} 
			\hline 
			\Tstrut
			 \multirow{3}{*}{QoI} & \multicolumn{2}{c}{Mean relative} \\ 
			  & \multicolumn{2}{c}{validation RMSE, $\varepsilon_\mathrm{val}$} \Bstrut \\ \cline{2-3} \Tstrut
			  & Standard & Bi-fidelity \\[0.5ex] 
			\hline 
			\Tstrut
			$z$ & $6.6642\times10^{-2}$ & $2.2397\times10^{-2}$  \\ 
			$\ub$ & $4.7907\times10^{-3}$ & $2.0235\times10^{-3}$  \\
			$u_3$ & $4.0933\times10^{-3}$ & $2.2730\times10^{-3}$\Bstrut\\ [1ex] 
			\hline 
		\end{tabular}

\end{table}

\textbf{Case II:} In this case, we use the standard Bouc-Wen formulation in \eqref{eq:bouc} as low-fidelity model, but the high-fidelity model now features the combined effects of degradation and pinching in \eqref{eq:pinching} with parameters specified in Section \ref{sec:hyst}. For this application, the configuration of the DeepONet as well as the size of the training and validation datasets are kept the same as in Case I. During training, the Adam algorithm is used with a learning rate of $4\times10^{-3}$ for 10,000 epochs based on preliminary studies; this value is reduced by half every 2,500 epochs. Case II only considers a single QoI, auxiliary variable $z$, for training and validation.

\begin{figure}
    \centering 
     \begin{subfigure}[b]{0.45\textwidth}
         \centering
         \includegraphics[scale=0.275]{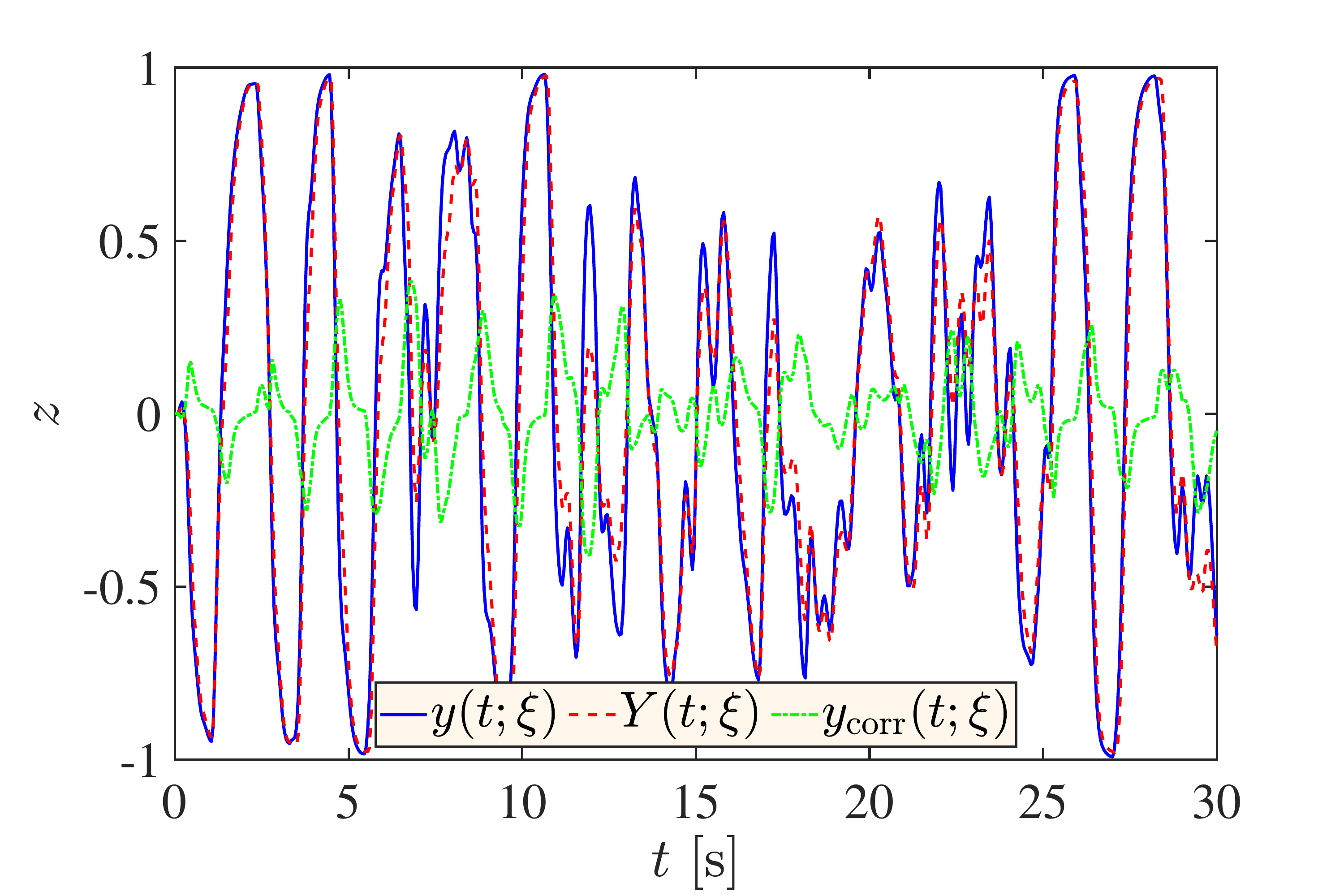}
         \caption{High-fidelity: Pinching and degradation \eqref{eq:pinching} with $\zeta_s=0.25$}
         \label{fig:ExI_z_Pinching1}
     \end{subfigure}
     \hfill
     \begin{subfigure}[b]{0.45\textwidth}
         \centering
         \includegraphics[scale=0.275]{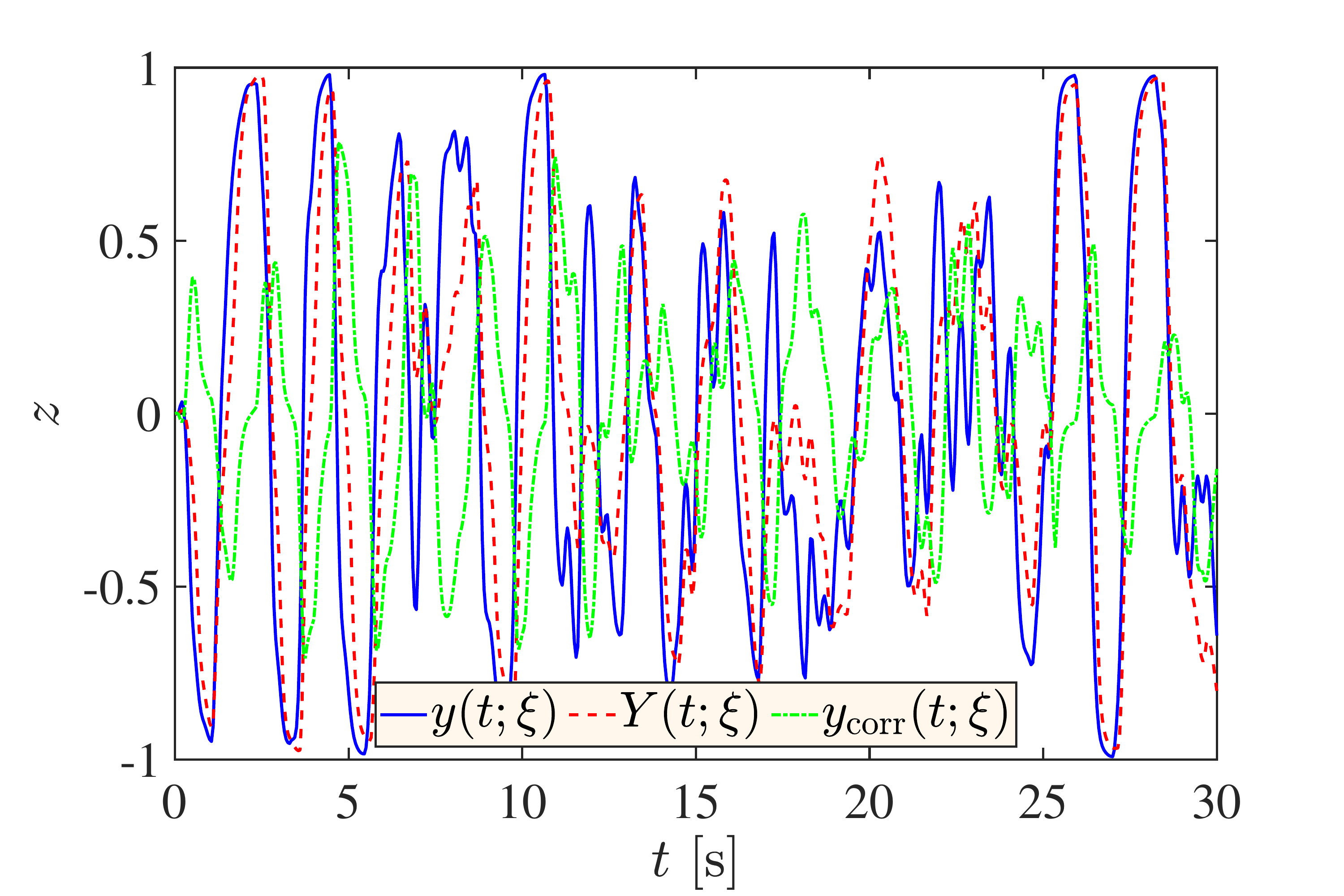}
         \caption{High-fidelity: Pinching and degradation \eqref{eq:pinching} with $\zeta_s=0.4$}
         \label{fig:ExI_z_Pinching2}
     \end{subfigure}
     \\ 
     \begin{subfigure}[b]{0.45\textwidth}
         \centering
         \includegraphics[scale=0.275]{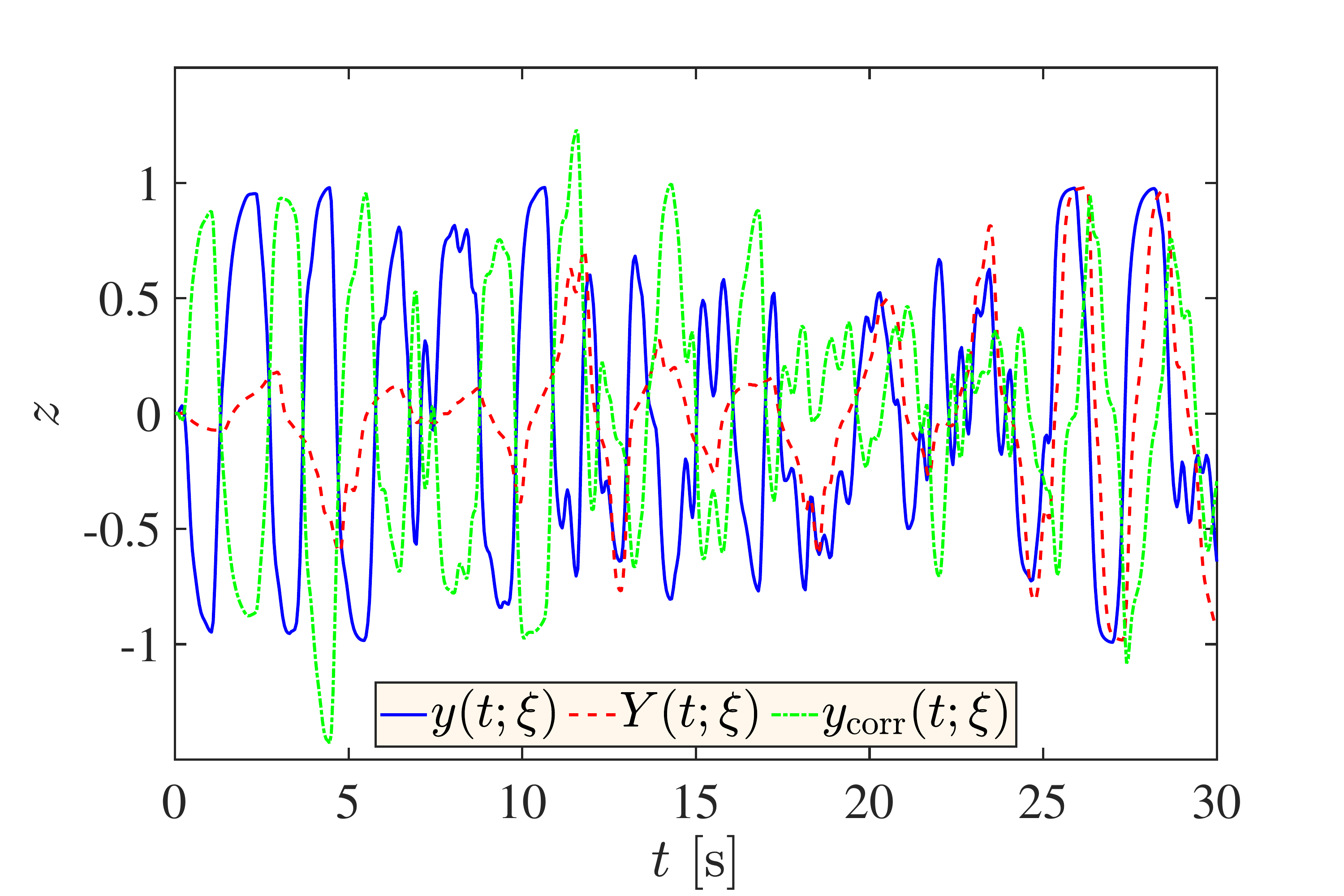}
         \caption{High-fidelity: Pinching and degradation \eqref{eq:pinching} with $\zeta_s=0.5$}
         \label{fig:ExI_z_Pinching3}
     \end{subfigure}
     \hfill
     \begin{subfigure}[b]{0.45\textwidth}
         \centering
         \includegraphics[scale=0.275]{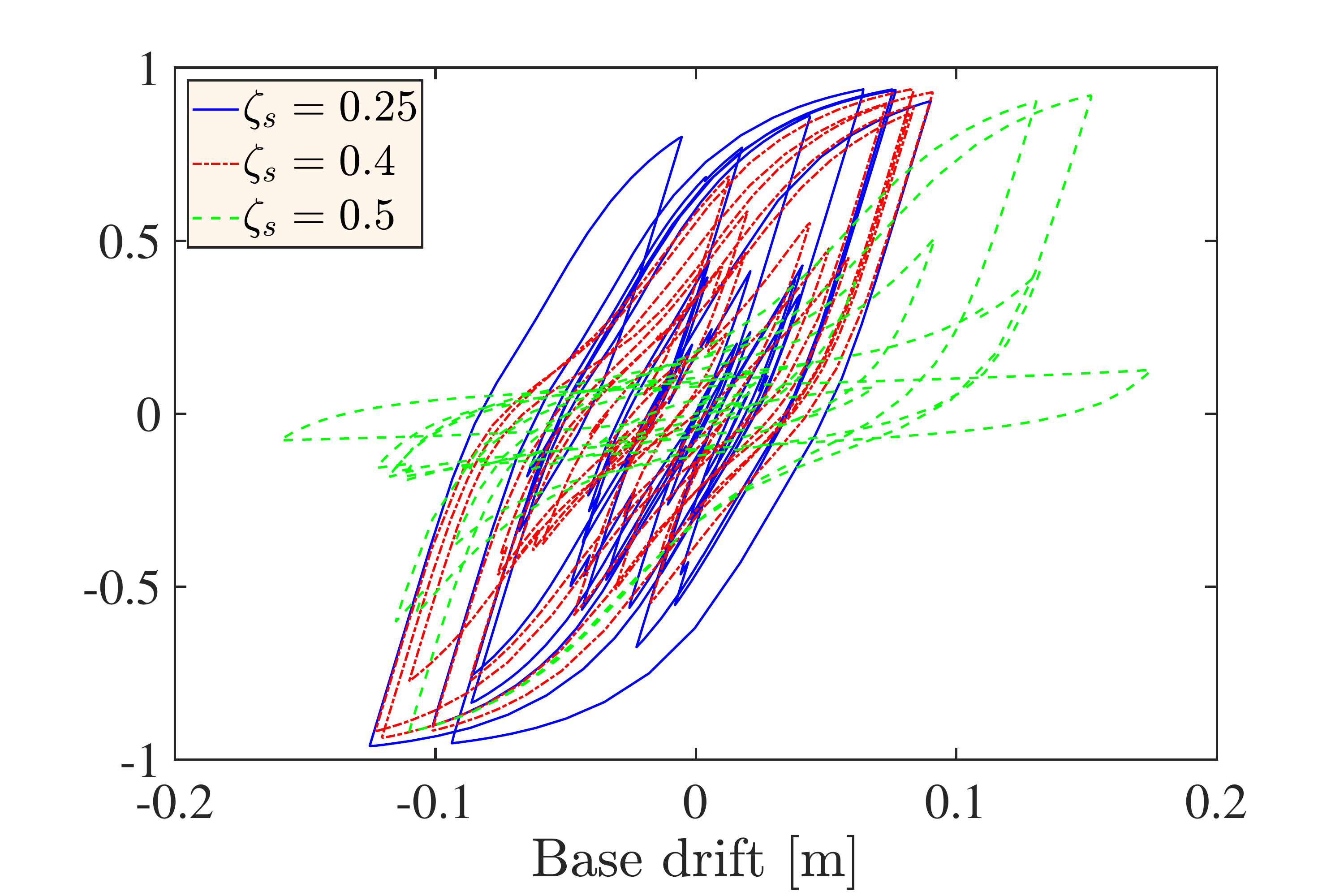}
         \caption{Hysteresis loops for three values of $\zeta_s$}
         \label{fig:ExI_z_Pinchingloop}
     \end{subfigure} 
    \caption{Comparison of auxiliary variable $z$ from the low- and high-fidelity model in Example 1. 
    Responses from the Bouc-Wen model (blue solid line) is used as the low-fidelity model, $y(t;\xii)$. The discrepancy $y_\mathrm{corr}(t;\xii)$ (green dashed line) with the degraded model $Y(t;\xii)$ (red dash-dotted line) is then modeled using a DeepONet. }
    \label{fig:z_Comp_pinching}
\end{figure}

Figure \ref{fig:z_Comp_pinching} uses individual realizations of the system to compare the auxiliary variable $z$ from \eqref{eq:pinching} against that from \eqref{eq:bouc} and the discrepancy between them for three different values of the total slip parameter $\zeta_s$. As the value of $\zeta_s$ increases, the discrepancy becomes more prominent and the advantage of a bi-fidelity approach diminishes.  Figure~\ref{fig:ExI_z_Pinchingloop} demonstrates how dramatically the hysteresis behavior changes when $\zeta_s$ undergoes a small change from 0.4 to 0.5.  Comparing these hysteresis plots for the true (high-fidelity) system to those for the low-fidelity system that uses the standard Bouc-Wen model shown in Figure~\ref{fig:bouc}, it is readily apparent that the standard Bouc-Wen model would be a poor approximator the high-fidelity system when $\zeta_s=0.5$.  Thus, the large discrepancy for this case is not surprising. 

\begin{figure}
    \centering 
     \begin{subfigure}[b]{0.45\textwidth}
         \centering
         \includegraphics[scale=0.275]{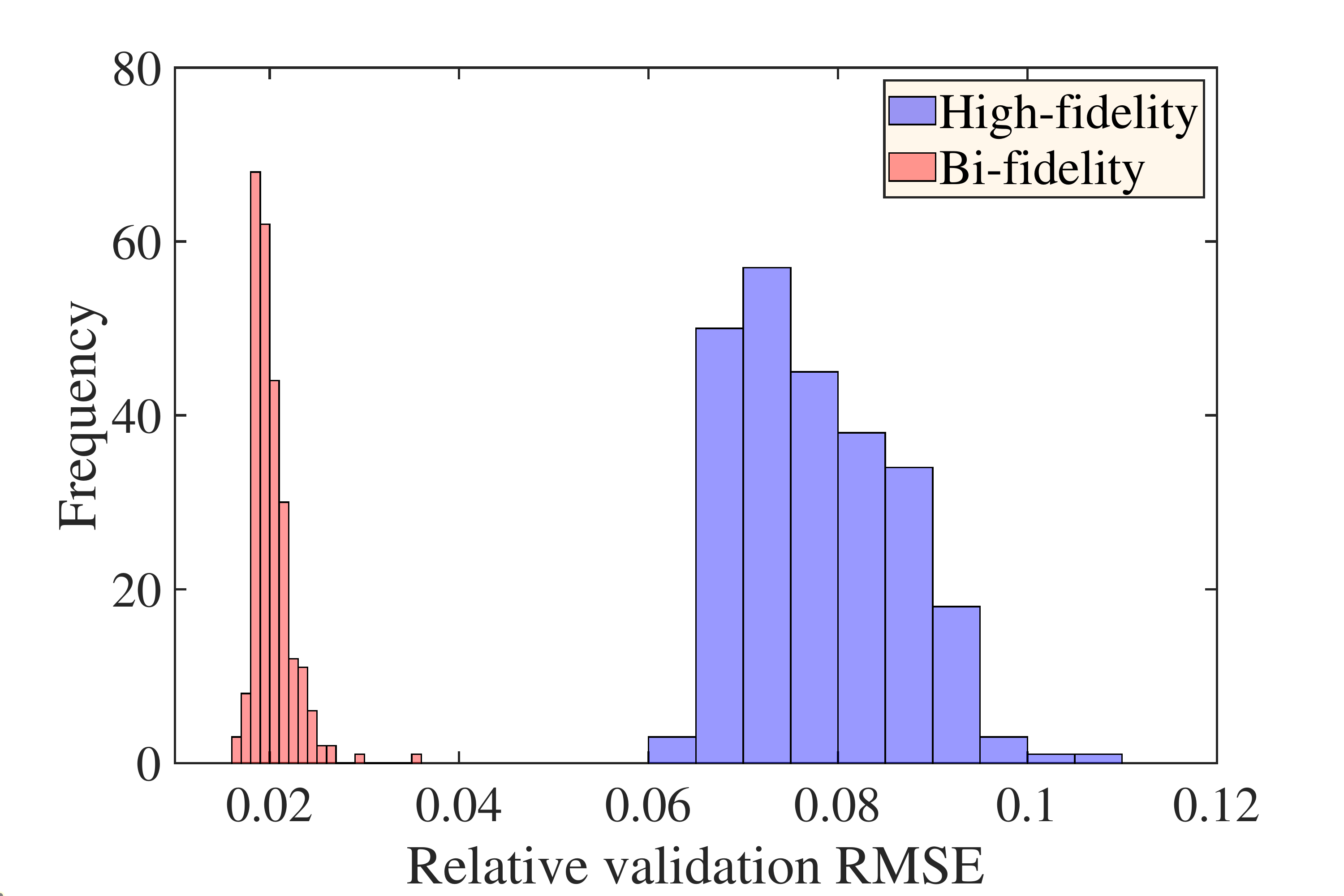}
         \caption{High-fidelity: Pinching and degradation \eqref{eq:pinching} with $\zeta_s=0.25$}
         \label{fig:ExI_Hist_Pinching1}
     \end{subfigure}
     \hfill
     \begin{subfigure}[b]{0.45\textwidth}
         \centering
         \includegraphics[scale=0.275]{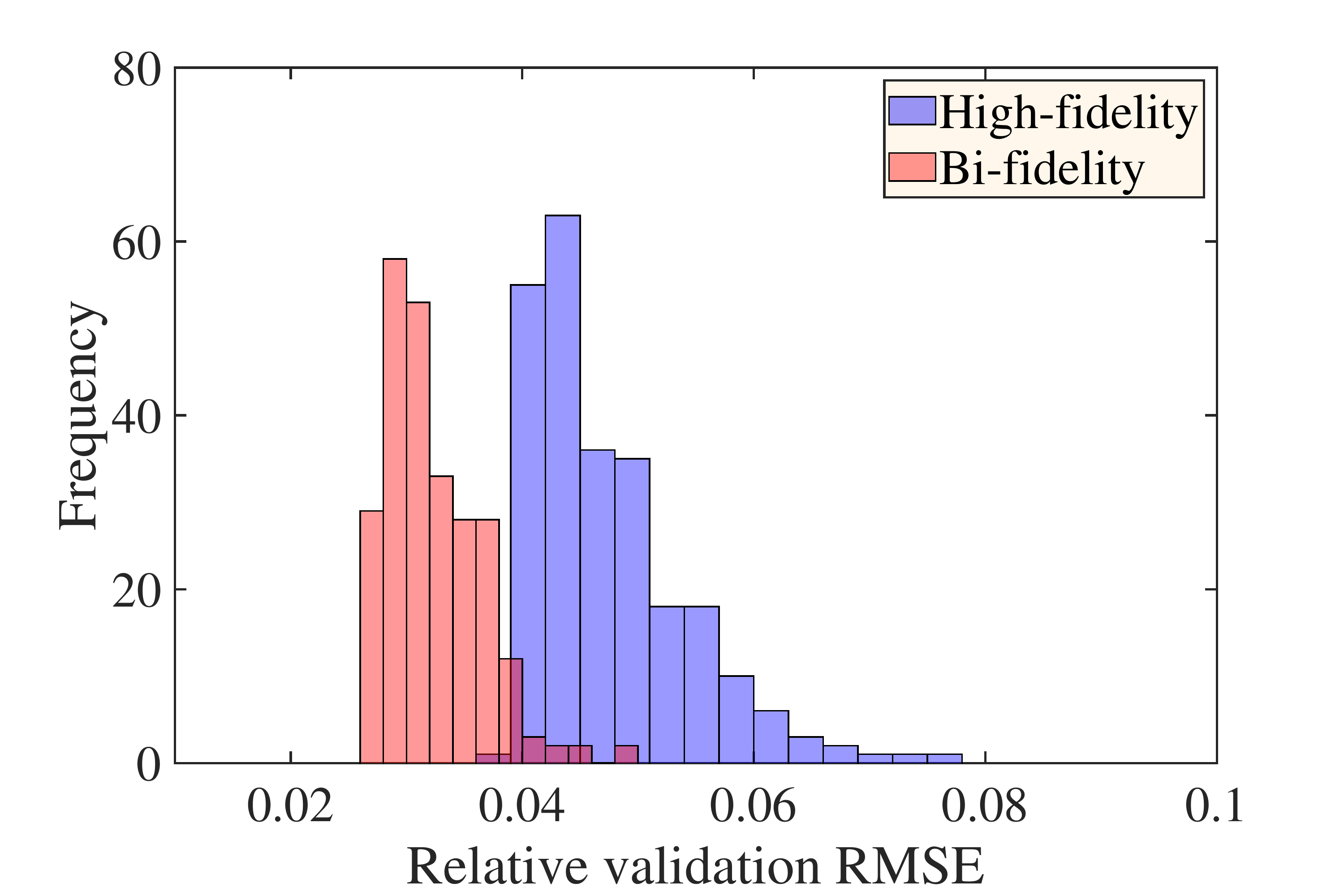}
         \caption{High-fidelity: Pinching and degradation \eqref{eq:pinching} with $\zeta_s=0.4$}
         \label{fig:ExI_Hist_Pinching2}
     \end{subfigure}
     \\
     \begin{subfigure}[b]{0.45\textwidth}
         \centering
         \includegraphics[scale=0.275]{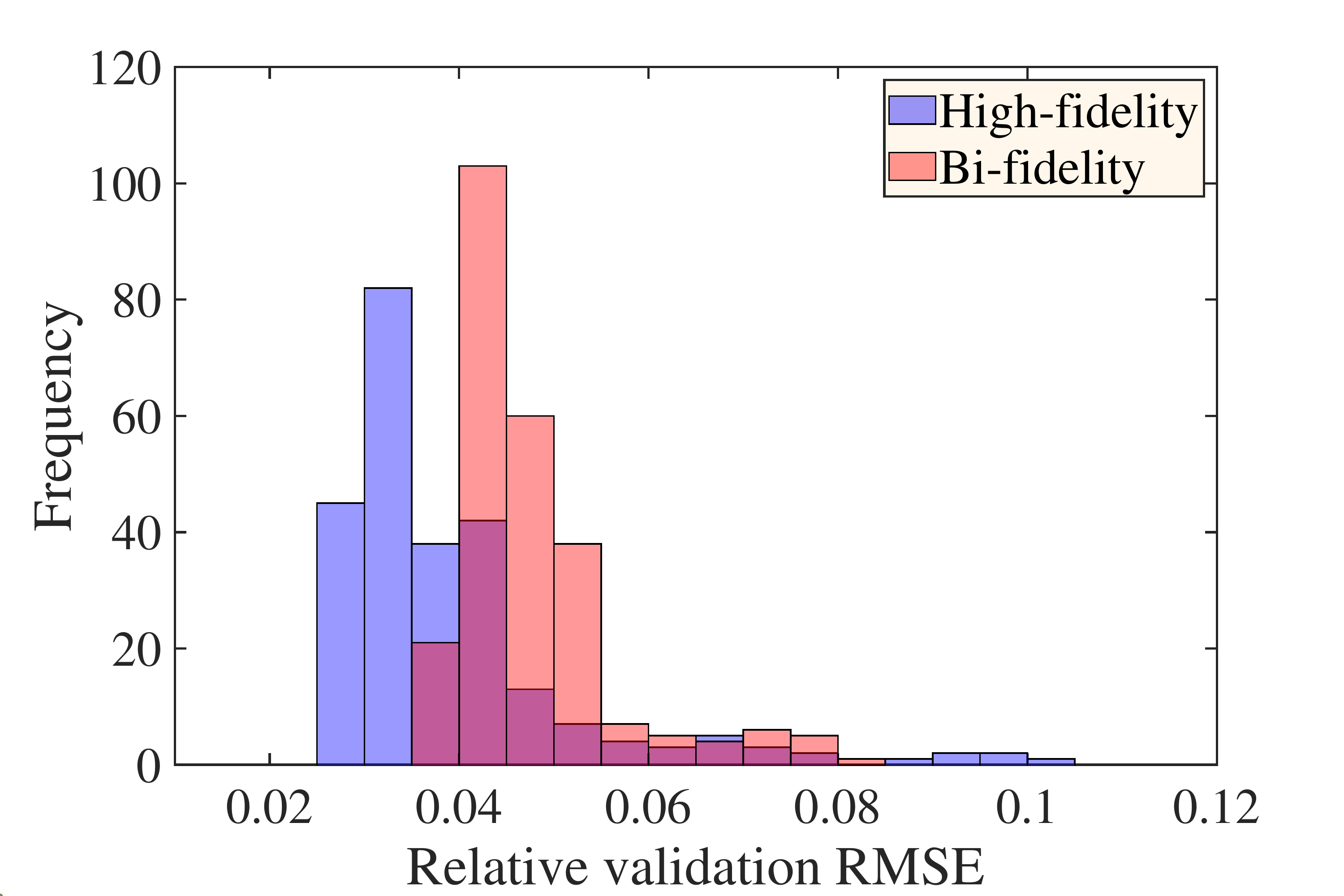}
         \caption{High-fidelity: Pinching and degradation \eqref{eq:pinching} with $\zeta_s=0.5$}
         \label{fig:ExI_Hist_Pinching3}
     \end{subfigure}
    \caption{Comparison of histograms of the relative validation error of the trained DeepONet using the high-fidelity pinching model and the bi-fidelity approach with Bouc-Wen model as the low-fidelity model in Case II of Example 1. }
    \label{fig:ExI_Histogram_Comp_Pinching}
\end{figure}

Figure \ref{fig:ExI_Histogram_Comp_Pinching} presents the relative validation RMS error histograms from the high-fidelity-only and bi-fidelity DeepONets for different values of $\zeta_s$.  While the distribution of validation errors for the bi-fidelity is rather significantly separated from the high-fidelity distribution for smaller values of $\zeta_s$, the histograms begin to converge as the value of $\zeta_s$ increases, providing further evidence of the how the advantage for the bi-fidelity approach diminishes.  Taken together, these figures emphasize how the choice of a low-fidelity model with similar behavior is an important assumption of the proposed approach, and that the advantages of this approach no longer remain if the low-fidelity model behavior is significantly different from the true system behavior. 


\subsection{Example 2: Nonlinear suspension of a vehicle} 
\begin{figure}[!htb]
		\centering
		\includegraphics[scale=0.5]{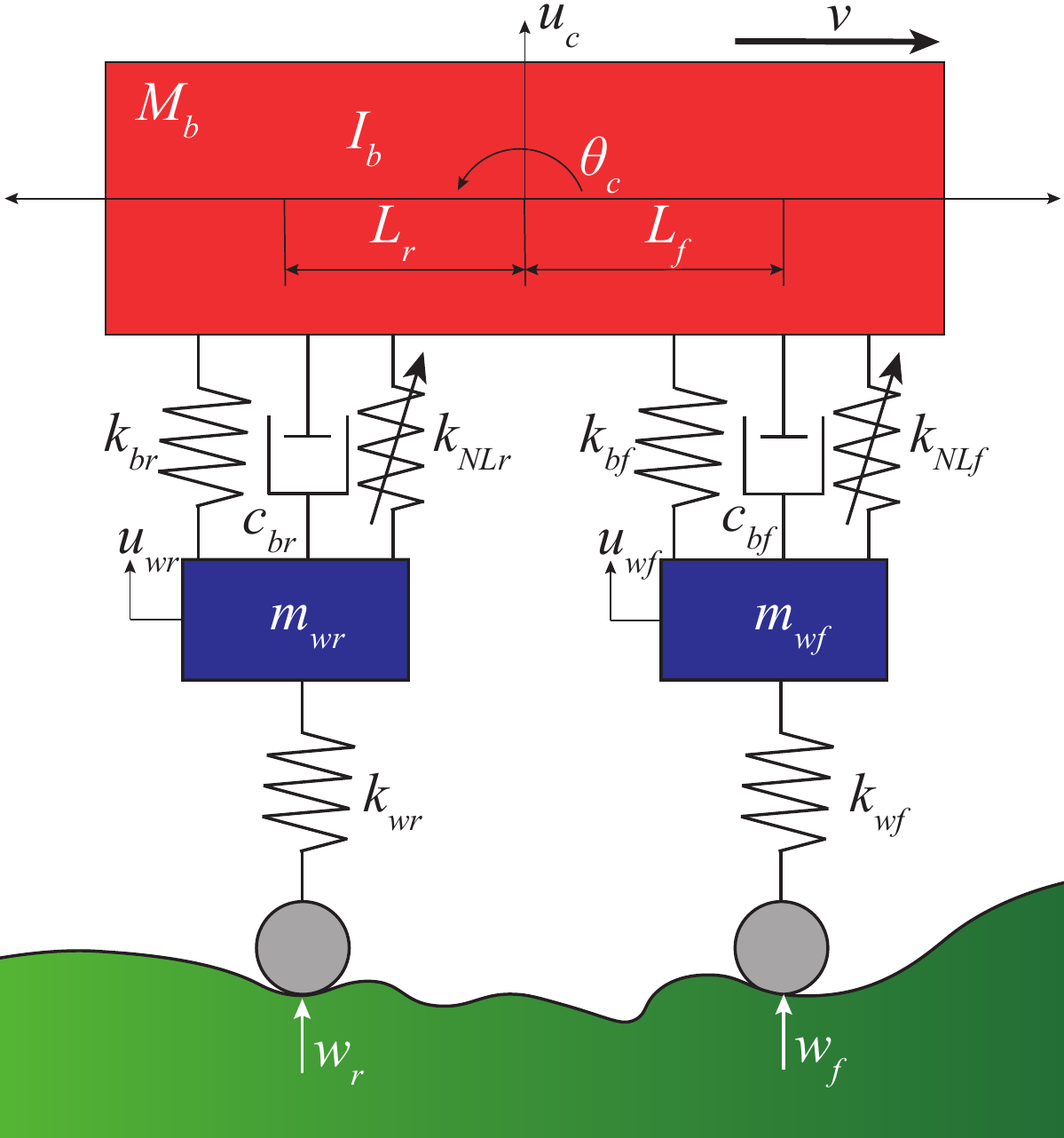}
	\caption{Schematic of half-car model with sprung mass used in Example 2. }
	\label{fig:car}
\end{figure}

The second 
example considers car models with nonlinear suspension. 
The car models span two levels of fidelity, a half-car model with unsprung mass (4 DOFs) \cite{YOSHIMURA199941} and a quarter-car model with unsprung mass (2DOFs) \cite{Agostinacchio:2014aa}.  
A schematic illustration of the half-car model with unsprung mass is shown in Figure~\ref{fig:car}.  Within the car model, $M_b$ is the mass for the vehicle body, $I_b$ is the mass moment of inertia for the vehicle body, and $m_{wf}$ and $m_{wr}$ are the masses of the front/rear wheels, respectively. The degrees of freedom are described by $u_c$ and $\theta_c$, where $u_c$ is the vertical displacement of the vehicle body at the centre of gravity and $\theta_c$ is the rotary angle of the vehicle body at the centre of gravity.  The vertical displacements of the front/rear wheels are described by $u_{wf}$, $u_{wr}$, respectively, and $L_f$ and $L_r$ define the distances to the front/rear suspension locations, with reference to the centre of gravity of the vehicle body, such that $L_f + L_r = L $. 

The suspension system connecting the car body to the wheels consists of three components, a linear spring, a viscous damper, and a nonlinear restoring force element, where $k_bf$ and $k_br$ are the linear springs of the front/rear suspensions, respectively, $c_{bf}$ and $c_br$ are the damping constants of the front/rear suspensions, respectively, and $k_NLf$ and $k_NLr$ are the nonlinear restoring force constants of the front/rear suspensions, respectively.  This model also considers stiffness constants for the wheel (unsprung masses), where $k_{wf}$ and $k_{wr}$ are the spring constants of the front/rear tires, respectively, and $w_f$ and $w_r$ are the irregular excitations from the road surface on the front and rear tires, respectively.  The velocity $v$ indicates the direction of travel. 

\begin{table}[!htb]
\begin{center}
		\caption{Parameter values for car model in Example 2. } \label{tab:car_param} 
		\begin{tabular}{c c l} 
			\hline 
			\Tstrut
			 Parameter & Value & Unit \\[0.5ex] 
			\hline 
			\Tstrut
			$M_b$ & ~~1794.40 & kg \\
            $I_b$ & 34430.50 & kg$\cdot$m$^2$\\
            $m_{wf}$ & ~~~~~~87.15 & kg \\
            $m_{wr}$ & ~~~~140.14 & kg \Bstrut\\ \hline
            \Tstrut$c_{bf}$ & 1190 & N$\cdot$s/m \\
            $c_{br}$ & 1000 & N$\cdot$s/m \Bstrut\\ \hline
            \Tstrut$L_{f}$ & 1.27 & m \\
            $L_{r}$ & 1.72 & m \Bstrut\\ \hline
			\Tstrut $v$ & $20$ & m/s\Bstrut\\ [1ex] 
			\hline 
		\end{tabular}
\end{center}
\end{table} 

\begin{table}[!htb]
\begin{center}
		\caption{Uncertain parameters for car model in Example 2.} \label{tab:car_param_unc} 
		\begin{tabular}{c l l c c c} 
			\hline 
			\Tstrut
			 Parameter & \multirow{2}{*}{Distribution} & \multirow{2}{*}{Mean} & \multirow{2}{*}{Std. Dev.}  \\ 
            (Unit) & & &  \\[0.5ex] 
			\hline 
            \Tstrut$k_{bf}$ [N/m] & Lognormal & ~~66,824 & 6,682.4 \\
            $k_{br}$ [N/m] & Lognormal & ~~18,615 & 1,861.5 \\
            $k_{wf}$ [N/m] & Lognormal & 101,115 & 10,111.5 \\
            $k_{wr}$ [N/m] & Lognormal & 10,111.5 & 10,111.5
            \\ 
			$r_k$ & Uniform & 0.1875 & 0.0361\Bstrut \\ [1ex] 
			\hline 
		\end{tabular}
\end{center}
\end{table}


The nonlinear restoring forces, $k_{NLr}$ and $k_{NLf}$, are described by the degrading Baber-Wen model shown in Equation~\eqref{eq:baber_wen} such that $k_{NLj} = \alpha_j \dot{z}_j$, where subscript $j$ is either $f$ or $r$ to denote front or rear, respectively.  Implementation of the degrading Baber-Wen model to compute $\dot{z}_j$ follows the methodology presented earlier, wherein $A_j = 2\beta_j = 2\gamma_j = {\kpre}_j/{\Qy}_j$, which keeps auxiliary variable $z$ in the range $[-1,1]$.  The yield force for the front, ${\Qy}_f$, is equivalent to 1.5\% of the total weight of the car body and front tire, while ${\Qy}_r$ considers 1.5\% of the combined weight of the car body and rear tire.  For both the front and rear suspension systems $\rk$ is assumed uncertain, but the pre-yield stiffness is related to the corresponding linear springs, \textit{i.e.} ${\kpre}_j=\rk\,k_{bj}$.  The peak of the nonlinear restoring force $\alpha_j$ is defined as $\alpha_j={\Qy}_j(1-\rk)$. The hysteresis shape parameter $\npow$ is set to 2 and the degradation parameters are chosen as $\delta_A=0.02$, $\delta_\nu=0.02$, $\delta_\eta=0.02$. 

The deterministic and uncertain parameter values are listed in Table~\ref{tab:car_param} and Table~\ref{tab:car_param_unc}.  Note that the suspension system of the rear wheel has an average value that is 1/10th that of the front wheel, but their variance is kept the same, as shown by the standard deviation values in the final column of Table~\ref{tab:car_param_unc}.  This naturally leads to the rear wheel suspension system having a much larger coefficient of variation; however, the results demonstrate that this does not significantly impact the development of the DeepONet.  A smaller stiffness value is chosen for the rear wheel suspension system in order to create better congruity between the half-car and quarter-car models, as the quarter-car model only uses the front tire suspension system.

The quarter-car model is created by simply taking the front half of the half-car model, \textit{i.e.}, using only the front tire and suspension along with the vehicle body; however, note that in the case of the quarter-car model, the mass of the car body is kept at $M_b$ in order to maintain comparable natural frequencies across model.  When considering the velocity and nonlinear suspension systems, the half-car model consists of 25 parameters, while the quarter-car model is fully described by only 12 parameters.  

The equations of motion for both car models follow the general form shown in Equation~\eqref{eq:mdof}. For the half-car model, the mass matrix is a diagonal matrix $\mathbf{M}=\mathrm{diag}(M_b,I_b,m_{wf},m_{wr})$, which corresponds to displacement vector $\uu=[u_c,\theta_c, u_{wf}, u_{wr}]^{\mathrm{T}}$. The damping $\mathbf{C}$ and stiffness $\mathbf{K}$ matrices for the half-car model are given by
\begin{subequations}
\begin{equation}
    \mathbf{C} = \left[\begin{array}{cccc}
        c_{bf}+c_{br} & c_{bf}L_f - c_{br}L_{r} & -c_{bf} & -c_{br} \\
        c_{bf}L_f - c_{br}L_{r} & c_{bf}L_f^2 + c_{br}L_{r}^2 & -c_{bf}L_f  & c_{br}L_r  \\
        -c_{bf} & -c_{bf}L_f  & c_{bf} & 0 \\
        -c_{br} & c_{br}L_r & 0 & c_{br} 
    \end{array}\right]
\end{equation}
\begin{equation}
    \mathbf{K} = \left[\begin{array}{cccc}
        k_{bf}+k_{br} & k_{bf}L_f - k_{br}L_{r} & -k_{bf} & -k_{br} \\
        k_{bf}L_f - k_{br}L_{r} & k_{bf}L_f^2 + k_{br}L_{r}^2 & -k_{bf}L_f  & k_{br}L_r \\
        -k_{bf} & -k_{bf}L_f  & k_{bf}+k_{wf} & 0 \\
        -k_{br} & k_{br}L_r  & 0 & k_{br}+k_{wr} 
    \end{array}\right]
\end{equation}
\end{subequations}
Since the hysteretic elements in the half-car model exist between the vertical DOFs, the influence matrix for the nonlinear restoring forces is given by
\begin{equation}
    \widetilde{\Lm} = \left[\begin{array}{cc}\
    \alpha_f & \alpha_r \\
    L_f\alpha_f & -L_r\alpha_r \\
    -\alpha_f & 0 \\
    0 & -\alpha_r
    \end{array}\right]
\end{equation}
with nonlinear term $\mathbf{g} = [\dot{z}_f,\, \dot{z}_r]^{\mathrm{T}}$.

For the quarter-car model, the displacement vector is modified to only include the vertical displacement of the car body and the front wheel, such that $\uu=[u_c, u_{wf}]^{\mathrm{T}}$.  The mass matrix is correspondingly modified to $\mathbf{M}=\mathrm{diag}(M_b,m_{wf})$, with damping and stiffness matrices given by
\begin{subequations}
\begin{equation}
    \mathbf{C} = \left[\begin{array}{cc}
        c_{bf} & -c_{bf}  \\
        -c_{bf} &  c_{bf}
    \end{array}\right]
\end{equation}
\begin{equation}
    \mathbf{K} = \left[\begin{array}{cccc}
        k_{bf} & -k_{bf} \\
        -k_{bf} & k_{bf}+k_{wf}
    \end{array}\right]
\end{equation}
\end{subequations}
The influence matrix simplifies to $\widetilde{\Lm} = [\alpha_f, -\alpha_f]^{\mathrm{T}}$ with nonlinear term $g = \dot{z}_f$.



Excitation is provided to the car models by a randomly generated road surface according to the ISO 8608 standard \cite{ISO8608}, which asserts that road surfaces are combinations of large number of periodic bumps with different amplitudes.  ISO 8608 classifies road roughness according to the power spectral density (PSD) of vertical displacements as a function of spatial frequency $\Omega$. Artificial road profiles can then be generated using a road class PSD according to the following equation 
\begin{equation}
    h(l) = \sum_{i=0}^N S_i \sin(2\pi \Omega_i l + \phi_i)
\end{equation}
where $h(l)$ is the road profile elevation along distance coordinate $l$, $\Omega_i$ is the $i$th spatial frequency and $\phi_i$ is a random phase angle following a uniform distribution in the range $[0,2\pi]$.  $S_n$ is the amplitude of the $n$th harmonic in the PSD of vertical displacement, which can be further expressed as
\begin{equation}
    S_i = 2^b \sqrt{2 \left(\frac{\Omega_0}{\Omega_i}\right)^a \Delta\Omega} \,\,\, \times 10^{-3}
\end{equation}
where $b$ denotes the road class and has a value that varies from 1 (very good) to 7 (extremely poor), $\Omega_0$ is the reference spatial frequency, $\Delta \Omega$ is the width of each spatial frequency band, and $a$ is a road surface index, which is generally taken as a constant value of 2 \cite{GANDHI20172}.  

When applied to the car model, the profile for the front tire $h_f(l)$ is simply taken as $h(l)$.  However, the road profile for the rear tire $h_r(l)$ is associated with a time delay according to $t_{\mathrm{delay}}=(L_f+L_r)/v$.
The road surface profiles must also be converted into forcing terms to form an excitation vector $\mathbf{w}$.  This is accomplished by multiplying by each profile by its corresponding wheel stiffness such that $\mathbf{w} =[{k_{wf}}h_f, {k_{wr}}h_r]\transpose= [w_f, w_r]\transpose$.

\subsubsection{Results} 
\begin{figure}[!htb] 
         \centering 
         \begin{subfigure}[b]{0.45\textwidth}
         \centering
         \includegraphics[scale=0.275]{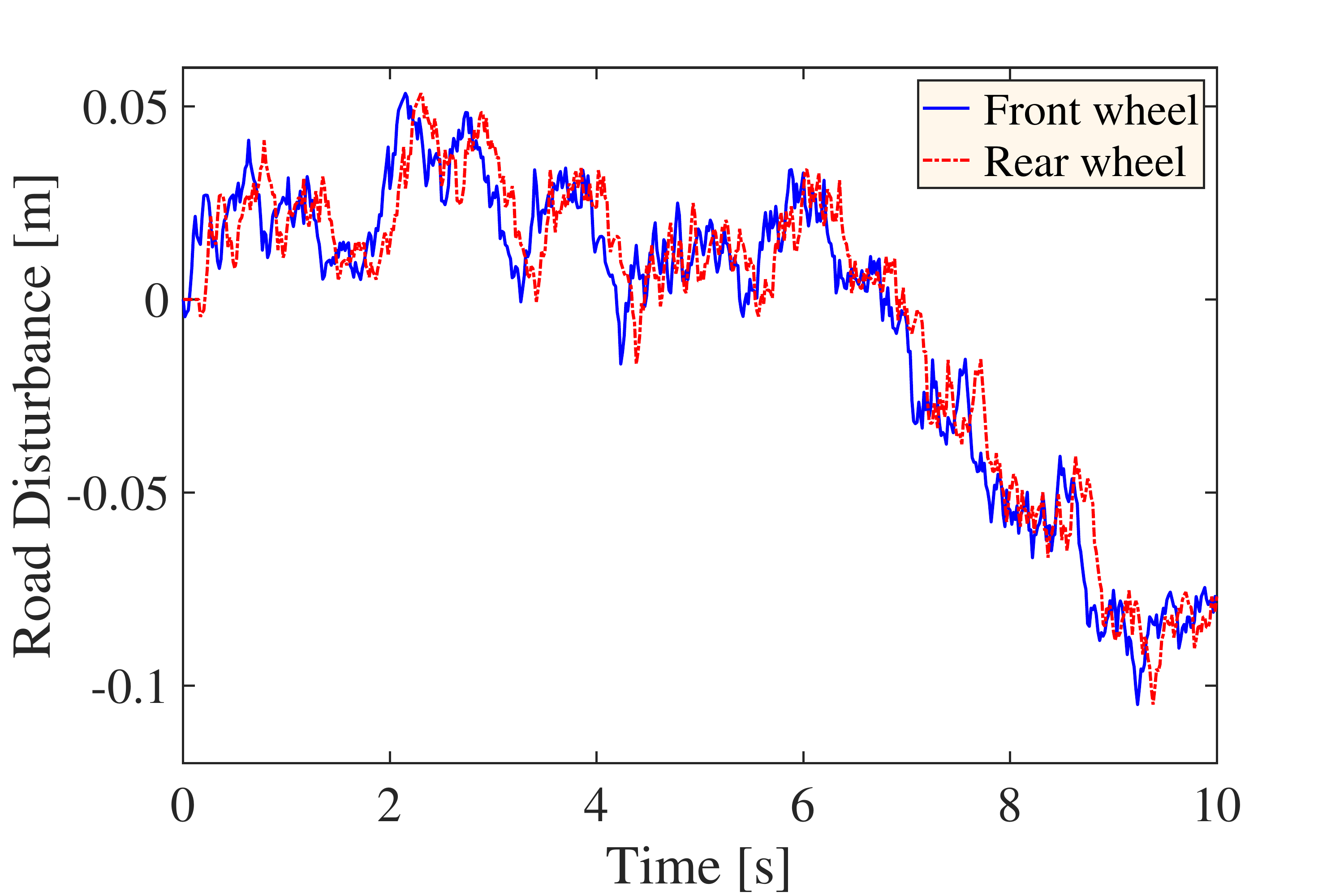}
         \caption{Sample road disturbance}
         \label{fig:rough}
     \end{subfigure}
     \hfill
     \begin{subfigure}[b]{0.45\textwidth}
         \centering
         \includegraphics[scale=0.275]{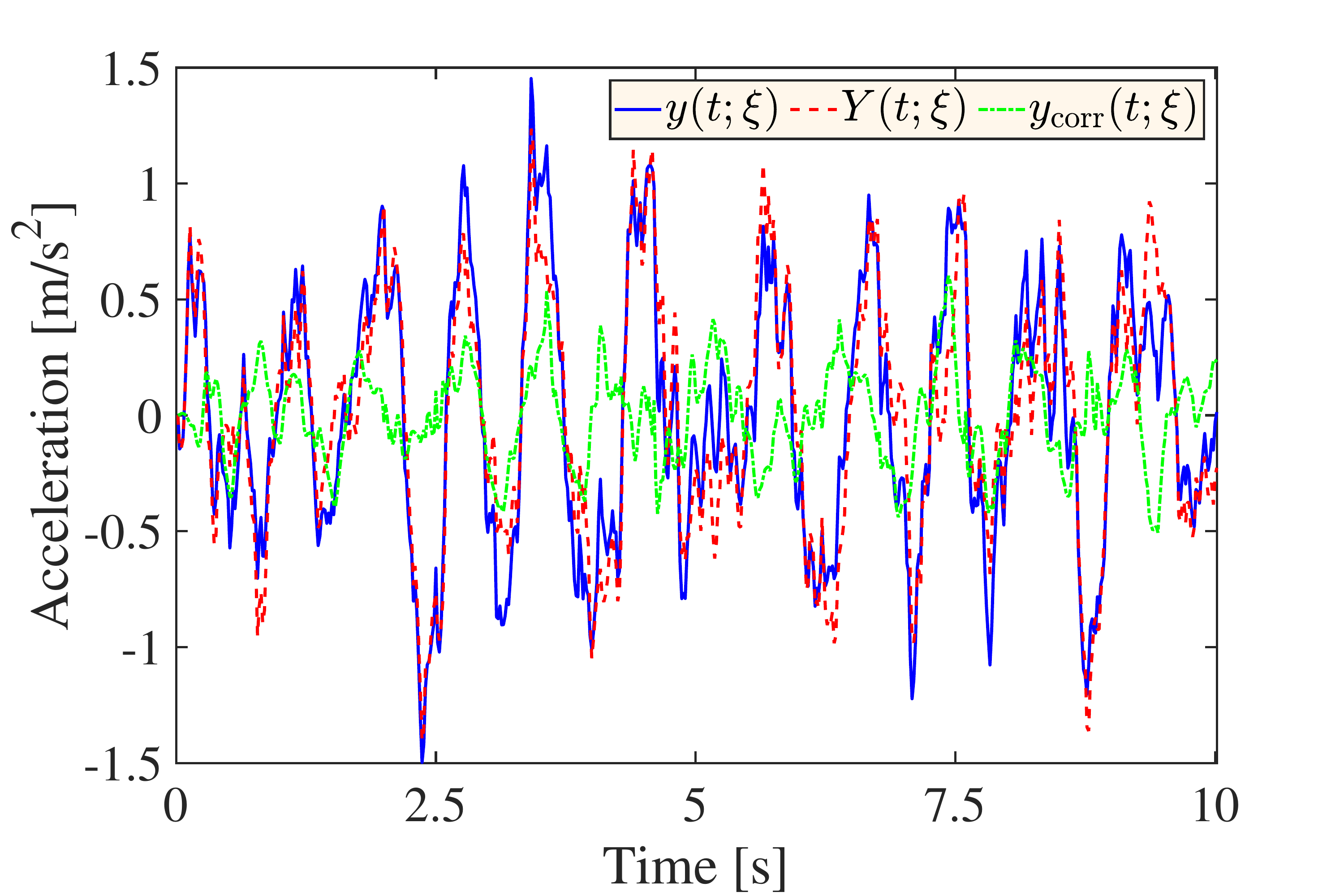}
         \caption{Comparison of responses}
         \label{fig:Car_Response}
     \end{subfigure}
         \caption{Sample road disturbance created by a road roughness profile with $b=1$ and typical responses of the car models used in Example 2.} 
         \label{fig:road_response}
\end{figure}

When generating the road roughness profile used, the road class parameter is initially taken as $b=1$.  Further, the reference spatial frequency is chosen as $\Omega_0 = 1$~cycle/m, while spatial frequency $\Omega_i$ varies from 0.01 to 100~cycles/m with a spatial bandwidth of $\Delta \Omega=1/d$, where $d$ is the distance traveled by the car.  This distance is simply the product of the car's velocity and the the duration of the simulation $t_{\mathrm{final}}$ such that $d=v\cdot t_{\mathrm{final}}$.  Figure~\ref{fig:rough} provides a realization of the road roughness profiles that serve as the road disturbances for the car models. 

When conducting the simulations, the acceleration of the center of mass is chosen as the observed measurements.  This selection is made for two primary reasons.  First, it is acknowledged that real-world scenarios would more readily permit the observation of acceleration as compared to displacement.  Second, using the acceleration at the center of mass, as opposed to another location, facilitates more direct comparisons between the half-car and quarter-car model.  Figure~\ref{fig:Car_Response} presents typical realizations of the acceleration responses from the ``low-fidelity'' quarter-car and ``high-fidelity'' half-car along with the discrepancy between them; note that these particular responses are generated from the disturbance(s) shown in Figure~\ref{fig:rough}.


A DeepONet configured similar to those in the previous example is utilized for the car model with the notable exception that $p$ (from \eqref{eq:deep}) has been increased to 10. This configuration is chosen according to the procedure previously detailed for the example in Section \ref{sec:ex1_results}. In this example, the prediction of the car's acceleration over a 10~s time span is used as the basis of comparing the performance of the bi-fidelity approach against a standard DeepONet.  In order to facilitate a fair comparison in terms of computational burden, the size of the training data sets are adjusted based on the costs associated with generating the training data.  For instance, it was found that the computational cost of the high-fidelity half-car model is 1.84 times the cost for low-fidelity quarter car model. Since the bi-fidelity network is trained using dataset $\Dtr$ with $\Ntr=250$, which includes 250 realizations of both the low- and high-fidelity models, the standard DeepONet, which only uses the high-fidelity model, is trained with $\Ntr=386$ so that the cost training data generation remains the same.  Both approaches are validated using dataset $\Dval$ with $\Nval=250$.

\begin{figure}[!htb] 
         \centering
         \includegraphics[scale=0.275]{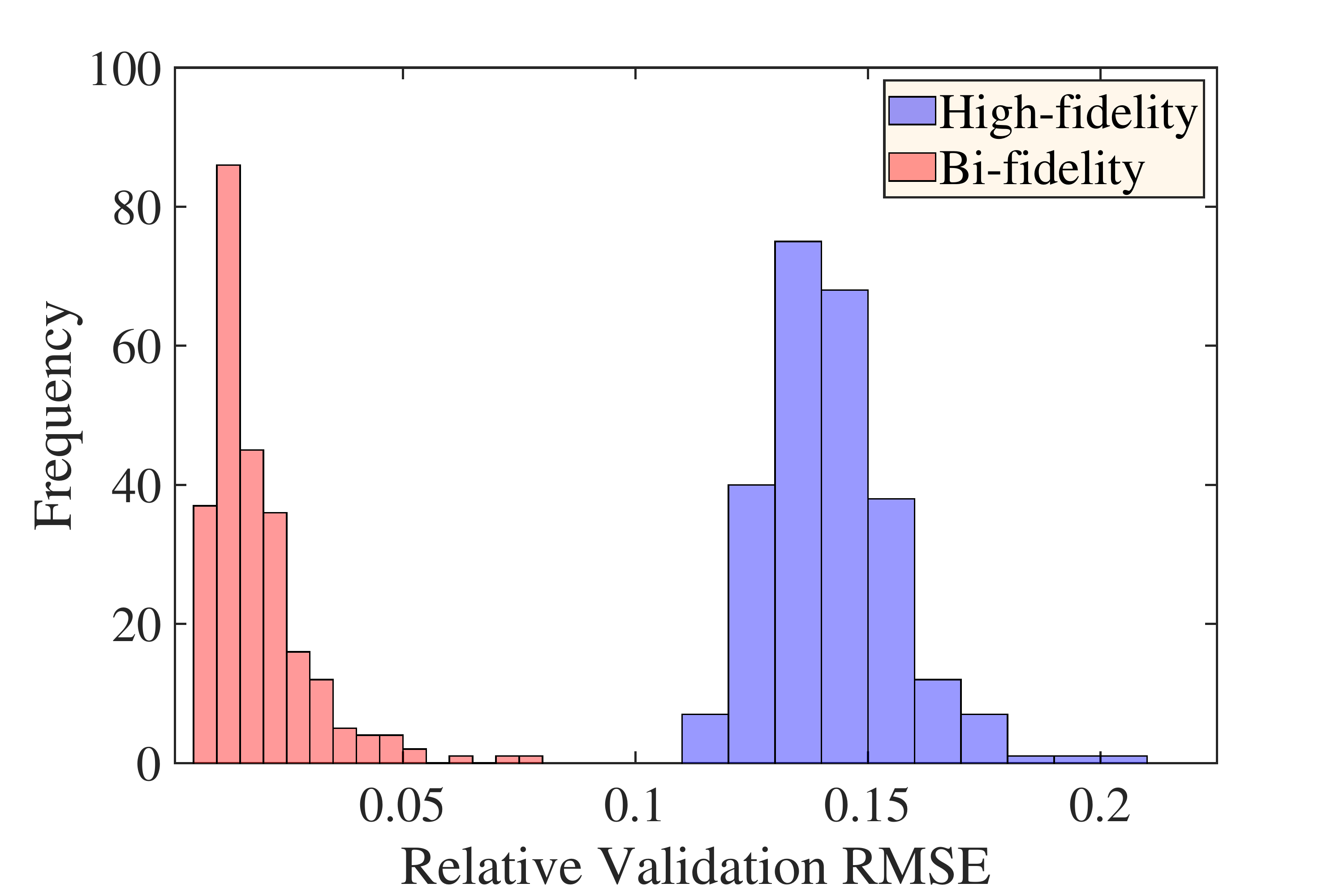}
         \caption{Comparison of histograms of the relative validation RMS error from the DeepONets trained using the high-fidelity model and the bi-fidelity approaches for Example 2.}
         \label{fig:hist_car} 
\end{figure}

For training, we utilize the Adam algorithm \cite{kingma2014adam} for 20,000 epochs with a learning rate of $10^{-3}$, which is halved every 2,500 epochs.  Note that the initial value of the learning rate is selected based on results from some preliminary runs. Once trained, the different DeepONets are then used to predict the data in the validation dataset.  Histograms of the relative validation RMS errors are compared in Figure \ref{fig:hist_car}. The mean relative validation errors from the bi-fidelity and standard DeepONets are $1.8737\times10^{-2}$ and $1.4191\times10^{-1}$, respectively.  Thus, the bi-fidelity DeepONet provides almost one order-of-magnitude improvement over the standard approach for this case.  The results of this example demonstrate that the assumed low-fidelity model does not need to have the same DOFs or model structure in order for the bi-fidelity DeepONet to offer an advantage. 

\subsection{Example 3: 100 Degree-of-freedom Base-isolated Building} 
\subsubsection{Description}

An 11-story 2-bay structure with a hysteretic base-isolation layer, adapted from Kamalzare et al. \cite{kamalzare2015efficient},  
is used for the third and final numerical example. 
Consistent mass matrices are used for the beams while ignoring the weights of the columns. The base layer moves horizontally with rigid in-plane behavior. With these assumptions the governing differential equations of the base-isolated structure are given by
\begin{equation}\label{eq:11story}
\begin{split}
&\Mm \ddot{\uu}_\mathrm{s} + \Cm \dot{\uu}_\mathrm{s} + \Km \uu_\mathrm{s} = - \Mm \rr \ddot{u}_\mathrm{g} + \Cm \rr \dot{u}_\mathrm{b} + \Km \rr u_\mathrm{b};\\
& m_\mathrm{b} \ddot{u}_\mathrm{b} + \left( c_\mathrm{b} + \rr^T \Cm \rr \right) \dot{u}_\mathrm{b} + \left( k_\mathrm{b} + \rr^T \Km \rr \right) {u}_\mathrm{b} + f_\mathrm{b} = -m_\mathrm{b}\ddot{u}_\mathrm{b} + \rr^T \Cm \rr \dot{\uu}_\mathrm{s} + \rr^T \Km \rr \uu_\mathrm{s},\\
\end{split}
\end{equation}
where $\Mm$ is the mass matrix, $\Cm$ is the damping matrix, and $\Km$ is stiffness matrix of the superstructure; $\uu_\mathrm{s}$ is displacement of the superstructure relative to the ground; $\ddot{u}_\mathrm{g}$ is the ground acceleration; and the influence vector for the ground acceleration is $\rr = [1,0,0,\dots,1,0,0]^T$, where the ones correspond to the horizontal displacement DOF. 
For the superstructure Rayleigh damping with 3\% damping ratios for the 1$^\mathrm{st}$ and 10$^\mathrm{th}$ is assumed. 
The structure has 100 DOF in total with three DOF per node for the 33 nodes of the superstructure and one horizontal DOF for the base layer. 
A record of the $1940$ El Centro earthquake measured at the N-S Imperial Valley Irrigation District substation with peak ground acceleration $0.348g$ is used as the ground acceleration $\ddot{u}_\mathrm{g}$. 

\begin{figure}[!htb]
		\centering
		\includegraphics[scale=0.64]{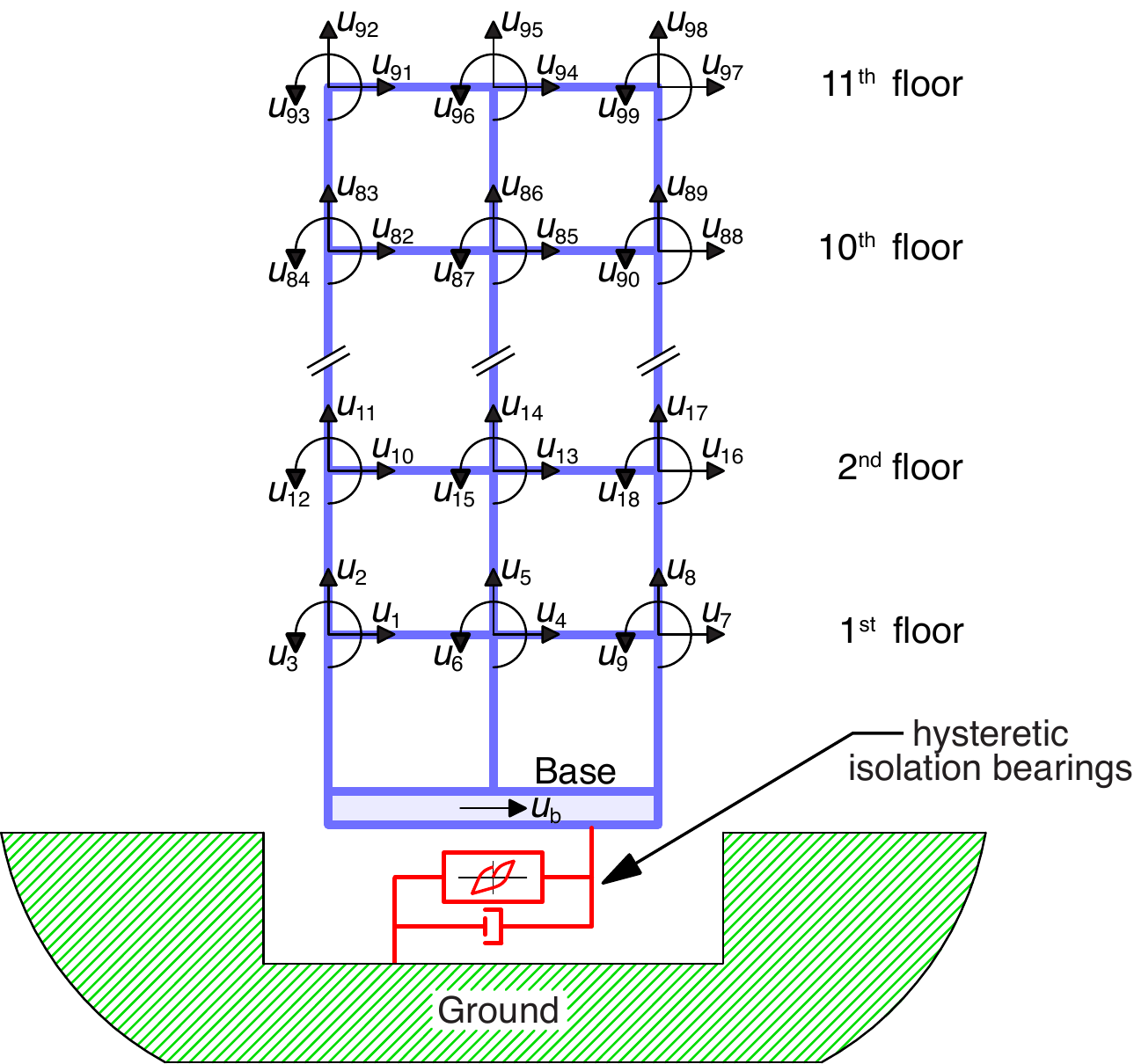}
	\caption{A 100 degree-of-freedom structure with a hysteretic base isolation layer. (adapted from Kamalzare et al. \cite{kamalzare2015efficient} and De et al. \cite{de2016computationally}) is used in Example 3. }
	\label{fig:100dof}
\end{figure}

As with Example 1, the low-fidelity model assumes a pristine condition with no degradation.  Thus, 
the restoring force in the passive base layer is governed by the standard Bouc-Wen hysteresis model as described in \eqref{eq:bouc} \cite{wen1976method} when generating the low-fidelity dataset. 
On the other hand, the ``true'' response, \textit{i.e.}, the response of the high-fidelity model, is obtained from the same model with two adjustments: 1) the passive isolation layer is governed by the Baber-Wen hysteresis model (see \eqref{eq:baber_wen}) that includes the effects of degradation and 2) the superstructure is assumed to be degraded by 50 years of urban corrosion \cite{kayser1989reliability} on its steel sections.  The effect of corrosion on the steel is modeled according to
\begin{equation}
    d_{\mathrm{loss}} = B_1t^{B_{2}}
\end{equation}
where $d_{\mathrm{loss}}$ is the corrosion penetration depth, $t$ is time in years, and $B_{1}$ and $B_{2}$ are empirical parameters that control the corrosion rate with values of 80.2 and 0.59, respectively \cite{kayser1989reliability}.  The penetration depth is translated to loss of steel by treating the corrosion penetration as a uniform loss of thickness throughout the steel beams, which then manifests as a reduction in the cross-sectional area and moment of inertia.

The governing equations of this structure can be converted to the state-space form in \eqref{eq:state-space} using  $\X = [\uu_\mathrm{s}^{\transpose} \quad u_\mathrm{b} \quad \dot{\uu}_\mathrm{s}^T \quad \dot{u}_\mathrm{b}]^{\transpose}$. Four parameters, namely $c_\mathrm{b}$, $\kpost$, $r_k$, and $\Qy$ of the base isolation restoring force, are assumed uncertain with their probability distributions given in Table \ref{tab:ExI_param} following De et al. \cite{de2016computationally}. These distributions are chosen such that significant variations are observed in the response. Further, a lognormal distribution is assumed for $\kpost$ and a Gaussian distribution truncated below at zero is assumed for $c_\mathrm{b}$ to ensure that both of these parameters remain positive. The yield force of the isolation layer $\Qy$ is given as a percentage of the total weight of the structure. A typical realization of the roof displacements in the horizontal direction are shown, without any added noise, for the pristine low-fidelity model and the corroding and degrading high-fidelity model in Figure~\ref{fig:ExIII_Response} along with the discrepancy between them. 



\begin{table}
\begin{center}
	\begin{threeparttable}[!htp]
		\caption{Probability distributions of the uncertain parameters and the lower and upper bounds of the design parameter in Example II. } \label{tab:ExII_param} 
		\begin{tabular}{c c c c c c c} 
			\hline 
			\Tstrut
			  Parameter & \multirow{2}{*}{Distribution} & \multirow{2}{*}{Mean} & \multirow{2}{*}{Std. Dev.} & Lower  & Upper \\ 
            (Unit) & & & & bound & bound \\[0.5ex] 
			\hline 
			\Tstrut
			 $\kpost$ [kN/m] & Lognormal & 750 & 20 & 0 & $\infty$\\ 
			 $c_\mathrm{b}$ [kN$\cdot$s/m] & Truncated Gaussian$^*$ & 35 & 2.5 & 0 & $\infty$\\
			 $r_k$ & Uniform & 0.1875 & 0.0361 & 0.125 & 0.250 \Bstrut\\ 
			 $\Qy$ (\%)$^\dagger$ & Uniform & 5.0 & 0.5773 & 4.0 & 6.0 \\ [1ex] 
			\hline 
		\end{tabular}
		\begin{tablenotes}
			\item[$^*$] Truncated below at zero; \item[$^\dagger$] in \% of the total weight of the structure.
		\end{tablenotes}
	\end{threeparttable}
\end{center}
\end{table} 

\begin{table}
\begin{center}
		\caption{Mean of the relative validation RMSE, $\varepsilon_\mathrm{val}$, in predicting the horizontal roof displacement for degraded hysteretic system with $\Ntr=250$ in Example 3. } \label{tab:ExIII_result} 
		\begin{tabular}{c c c c c c c} 
			\hline 
			\Tstrut
			 \multirow{3}{*}{Noise} & \multicolumn{2}{c}{Mean relative} \\ 
			  & \multicolumn{2}{c}{validation RMSE, $\varepsilon_\mathrm{val}$} \Bstrut \\ \cline{2-3} \Tstrut
			  & Standard & Bi-fidelity \\[0.5ex] 
			\hline 
			\Tstrut
			$0\%$ & $2.1717\times10^{-2}$ & $4.6541\times10^{-3}$  \\ 
                $5\%$ & $2.4413\times10^{-2}$ & $6.4530\times10^{-3}$ \\
			$10\%$ & $2.5730\times10^{-2}$ & $8.8876\times10^{-3}$  \\ 
   [1ex] 
			\hline 
		\end{tabular}
\end{center}
\end{table}

\subsubsection{Results} 

We use a DeepONet with a branch network consisting of three hidden layers with 100 neurons in each layer and a trunk network consisting of two hidden layers with 100 neurons each. For the activation function, we use the ELU (exponential linear unit) \cite{clevert2015fast}. This DeepONet utilizes a setting of $p=20$. The number of hidden layers in the trunk and branch networks, as well as the number of neurons per hidden layer, are selected using an iterative procedure, where the number of neurons per hidden layer is gradually increased to an upper limit (100 in this example).  Hidden layers are then added to the network until no reduction of $\epsv$ is observed. For the number of terms $p$ in \eqref{eq:deep}, we increase $p$ by 10 to its maximum value, which is the same as the number of neurons in the hidden layers, to observe any reduction of $\epsv$. 

In this example, the prediction of the horizontal roof displacement for a time span of 30~s is used to compare the performance of the bi-fidelity approach against a standard DeepONet. 
The bi-fidelity network, which requires realizations of both the ``pristine'' low-fidelity model and the degraded high-fidelity model, is trained using dataset $\Dtr$ with $\Ntr=250$ realizations of the roof displacement. To maintain an equivalent computational load for training data generation, the standard DeepOnet, which only uses the degraded model, is trained with $\Ntr=500$ because the computational cost of simulating the corroded and degraded building is nearly the same as the cost associated with pristine model. 
For training, we utilize the Adam algorithm \cite{kingma2014adam} for 10,000 epochs with a learning rate of $2\times10^{-3}$, which is halved every 2,500 epochs.  Note that the initial value of the learning rate is selected based on results from some preliminary runs. 
The trained DeepONets are used to predict the data in the validation dataset $\Dval$ with $\Nval=250$. 

\begin{figure}
    \centering 
     \begin{subfigure}[b]{0.45\textwidth}
         \centering
         \includegraphics[scale=0.275]{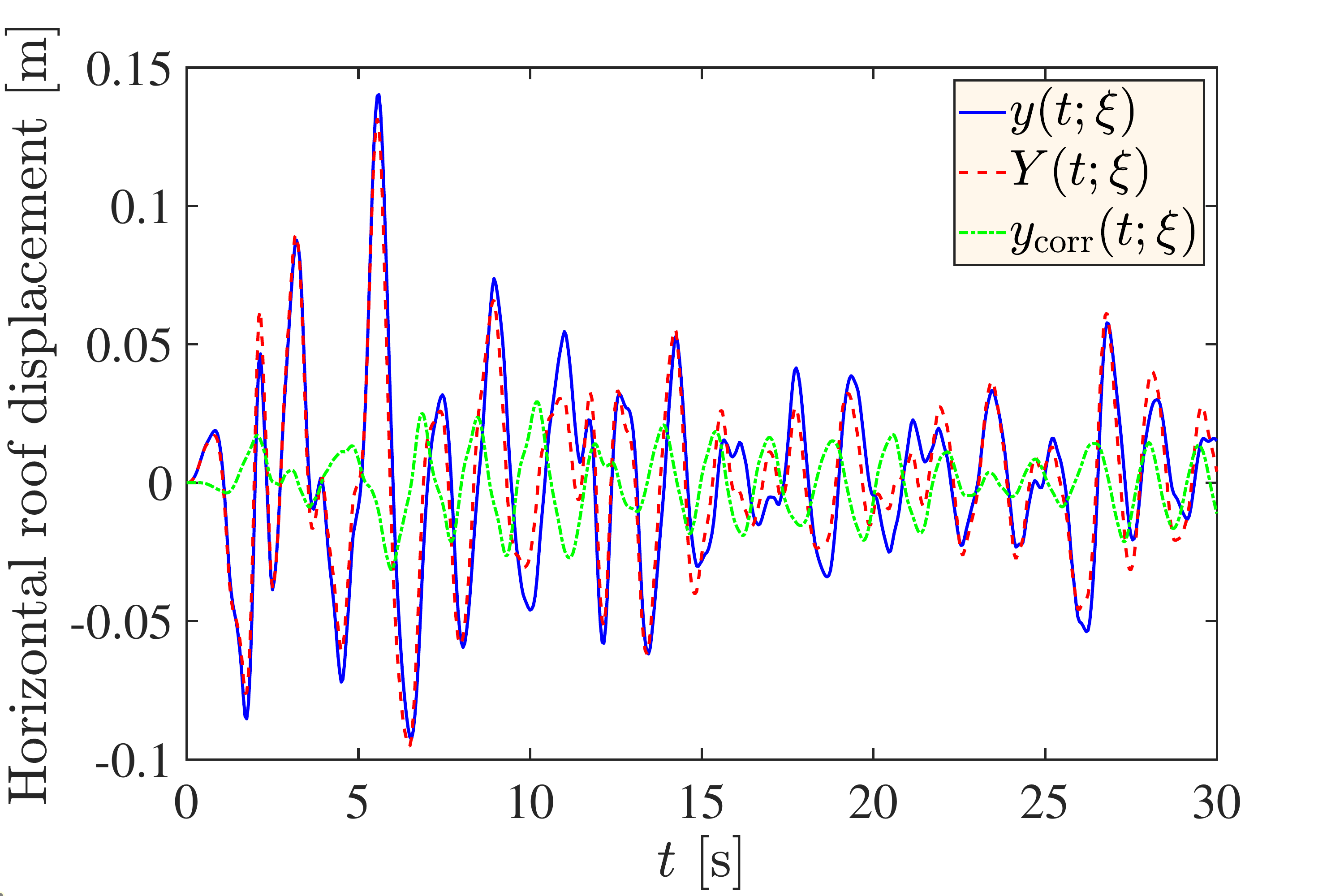}
         \caption{Comparison of responses with no added noise}
         \label{fig:ExIII_Response}
     \end{subfigure}
     \hfill
     \begin{subfigure}[b]{0.45\textwidth}
         \centering
         \includegraphics[scale=0.275]{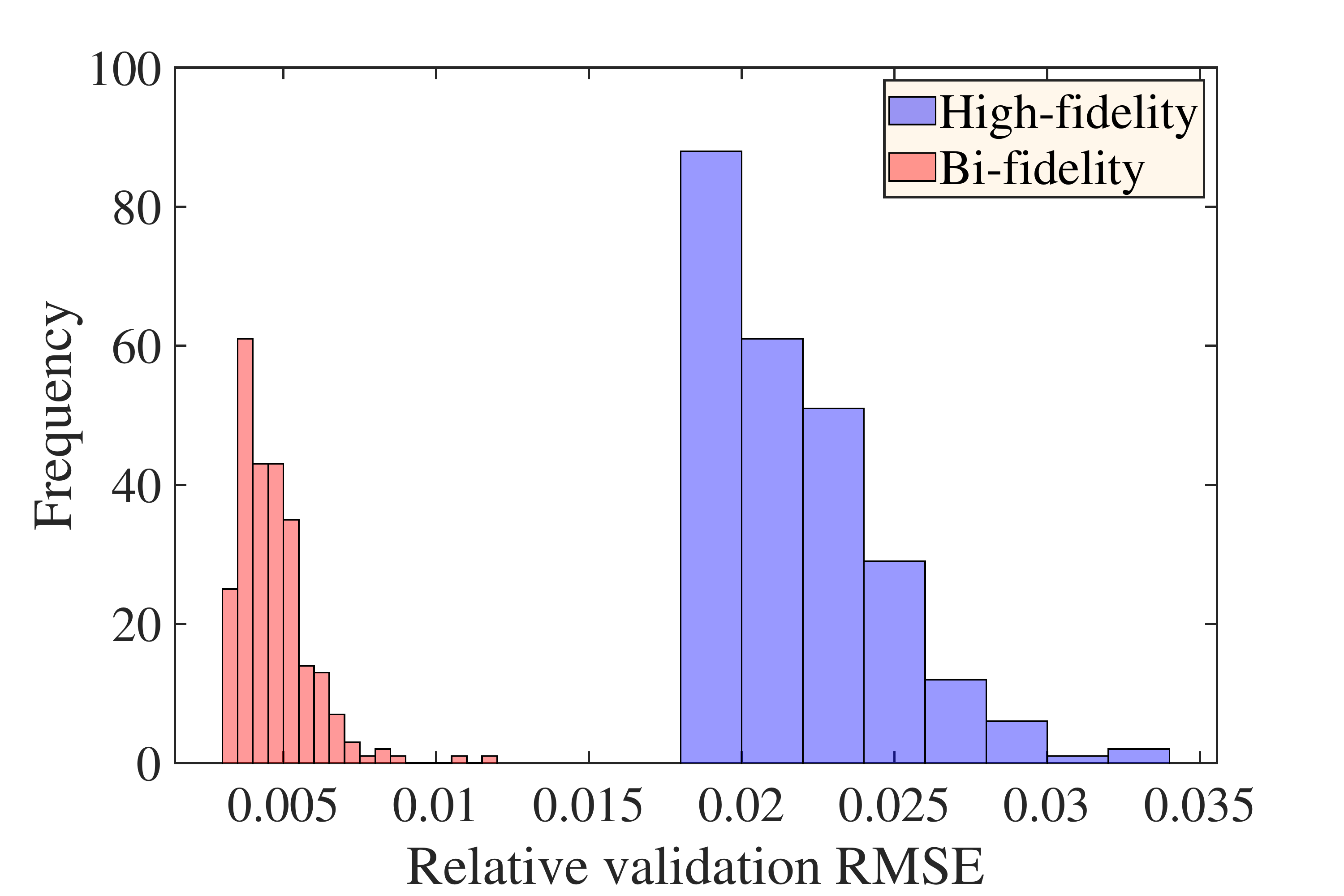}
         \caption{Comparison of relative validation error with no noise  in the training dataset}
         \label{fig:ExIII_hist1}
     \end{subfigure} \\ 
     \begin{subfigure}[b]{0.45\textwidth}
         \centering
         \includegraphics[scale=0.275]{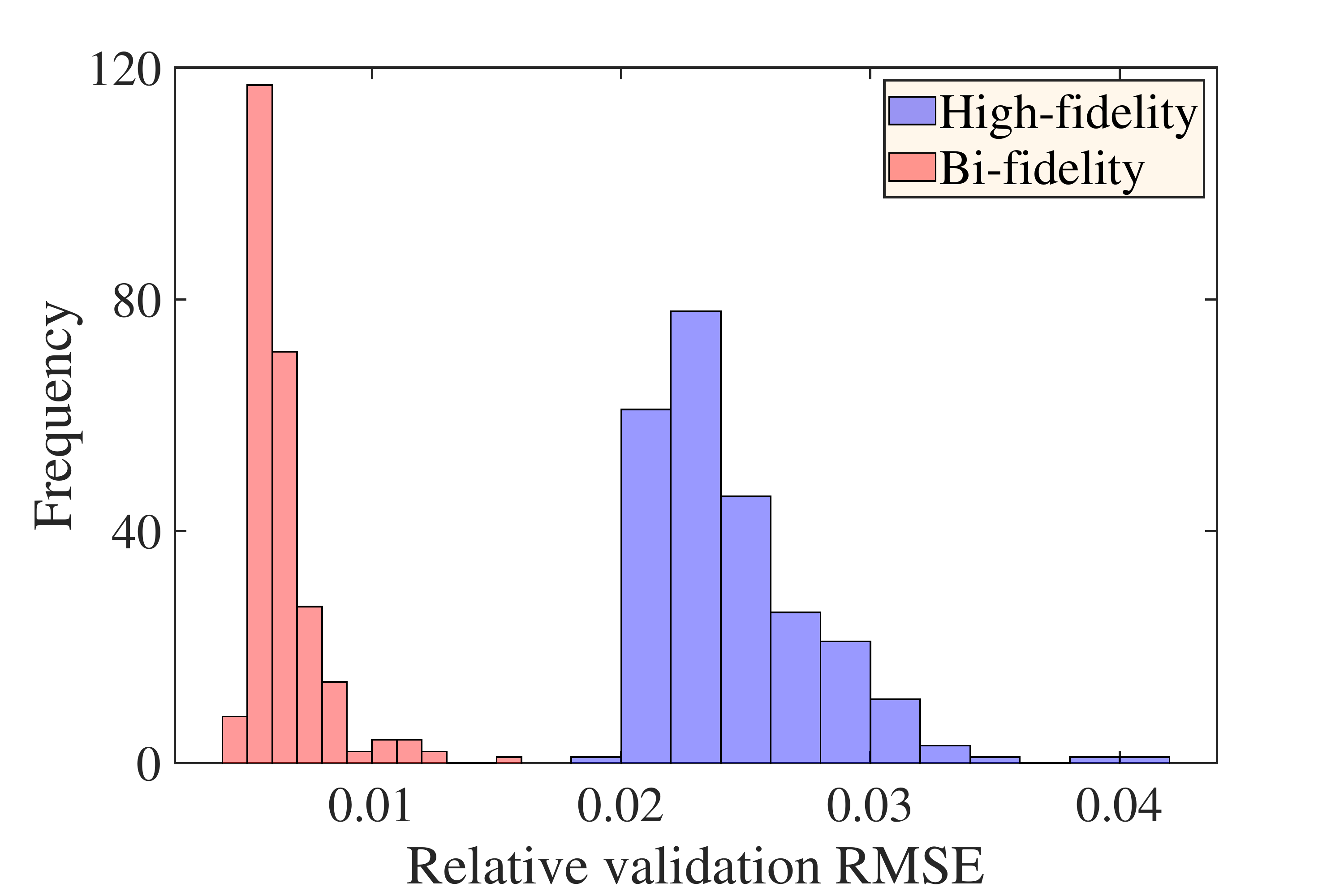}
         \caption{Comparison of relative validation error with 5\% noise  in the training dataset} 
         \label{fig:ExIII_hist2}
     \end{subfigure} \hfill 
     \begin{subfigure}[b]{0.45\textwidth}
         \centering
         \includegraphics[scale=0.275]{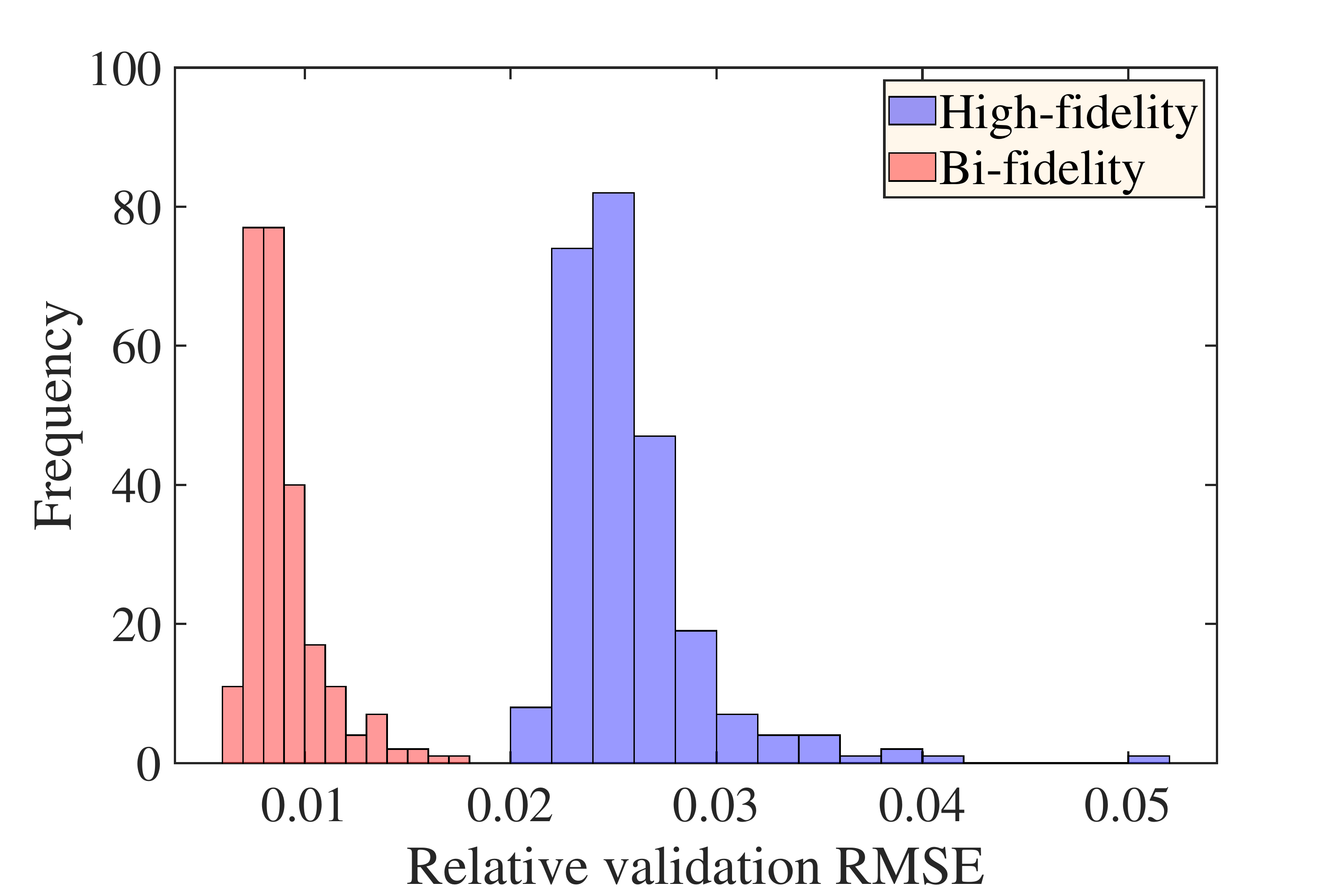}
         \caption{Comparison of relative validation error with 10\% noise in the training dataset}
         \label{fig:ExIII_hist3} 
     \end{subfigure} 
    \caption{Comparison of typical responses and histograms of relative validation error for the 100 DOF building used in Example 3. } 
    \label{fig:ExIII_Comp}
\end{figure}

To better facilitate comparisons between the two DeepONet approaches, three cases are conducted: a nominal case with no noise added to the training data and then two additional cases that add zero-mean Guassian white noise with standard deviation of 5\% and 10\% of the standard deviation of the original response 
to the training data. Figure \ref{fig:ExIII_Comp} compares histograms of the relative validation RMS error for the three cases.  From these histograms, it is clear that optimal performance for the bi-fidelity DeepONet is achieved when no noise is added to the training data and that the advantage of the bi-fidelity method dissipates as the intensity of the noise increases.  However, even with 10\% added noise, Figure~\ref{fig:ExIII_hist2} still shows notable separation between the performance of the bi-fidelity DeepONet and the standard implementation.  The mean relative validation errors from the two  DeepONet approaches are listed in Table \ref{tab:ExIII_result}, which further demonstrates the significant improvements of the bi-fidelity DeepONet even in the presence of noise.

\section{Conclusions} 

This paper proposes a bi-fidelity approach for modeling partially unknown and degrading responses from hysteretic systems using DeepONets.  The bi-fidelity DeepONet leverages low fidelity representations of a ``true'' degrading hysteretic system, \textit{e.g.}, representations that describe the hysteretic behavior but omit any degradation parameters, to offer computationally less expensive means of training while boasting superior accuracy for subsequent response predictions.  Importantly, the bi-fidelity DeepONet only attempts to learn the discrepancy between the low-fidelity representation and the true degrading system, as opposed to learning to predict the full response.  The effectiveness of the proposed bi-fidelity DeepONet is demonstrated through a series of numerical examples.

It was shown though the first example that the bi-fidelity DeepONet offers superior performance to a standard DeepONet implementation for a variety of response quantities of interest on a 4DOF structure, as the bi-fidelity implementation regularly delivered reductions in the mean validation error by a factor of two or three. It was similarly shown the performance advantages are most pronounced when the number of training data are limited, as the performance advantage steadily declines as the amount of available training data increases.  Further, it was shown that the degree of ``mismatch'' between the true system and assumed low-fidelity model can eventually compromise performance, as a bi-fidelity DeepONet with a standard Bouc-Wen assumption for its low-fidelity implementation is a poor approximator for a true system with pinching and degrading hysteresis.

It was also shown that the proposed bi-fidelity DeepONet offers improved performance over a standard implementation even when the low-fidelity model is missing DOFs from the true system, which was demonstrated with half- and quarter-car models.  Similarly, even when controlling for computational costs associated with generating training data for large structural models, \textit{e.g.} those with 100 DOFs, the bi-fidelity DeepONet routinely provides lower validation errors.  Further, this advantage was maintained in the presence of additive noise up to 10\% RMS.  Taken together, these structural examples demonstrate that the bi-fidelity DeepONet approach is a powerful tool when modeling and predicting the behavior of degrading hysteretic systems. Future studies should explore extensions into multi-fidelity implementations, \textit{i.e.}, utilizing more than 2 levels of fidelity, systems with greater nonlinear complexity, and applications of reduced order modeling for developing low-fidelity representations.

\bibliographystyle{plain} 
\bibliography{references.bib} 

\begin{thebibliography}{10}

\bibitem{Agostinacchio:2014aa}
M.~Agostinacchio, D.~Ciampa, and S.~Olita.
\newblock The vibrations induced by surface irregularities in road pavements
  --a matlab{\textregistered}approach.
\newblock {\em European Transport Research Review}, 6(3):267--275, 2014.

\bibitem{ahmed2023multifidelity}
Shady~E Ahmed and Panos Stinis.
\newblock A multifidelity deep operator network approach to closure for
  multiscale systems.
\newblock {\em arXiv preprint arXiv:2303.08893}, 2023.

\bibitem{noori1985random}
Thomas~T. Baber and Mohammad~N. Noori.
\newblock Random vibration of degrading, pinching systems.
\newblock {\em Journal of Engineering Mechanics}, 111(8):1010--1026, 1985.

\bibitem{baber1981random}
Thomas~T Baber and Yi-Kwei Wen.
\newblock Random vibration hysteretic, degrading systems.
\newblock {\em Journal of the Engineering Mechanics Division},
  107(6):1069--1087, 1981.

\bibitem{blakseth2021deep}
Sindre~Stenen Blakseth, Adil Rasheed, Trond Kvamsdal, and Omer San.
\newblock Deep neural network enabled corrective source term approach to hybrid
  analysis and modeling.
\newblock {\em arXiv preprint arXiv:2105.11521}, 2021.

\bibitem{bottou2010large}
L{\'e}on Bottou.
\newblock Large-scale machine learning with stochastic gradient descent.
\newblock In {\em Proceedings of COMPSTAT'2010}, pages 177--186. Springer,
  2010.

\bibitem{bottou2012stochastic}
L{\'e}on Bottou.
\newblock Stochastic gradient descent tricks.
\newblock In {\em Neural networks: Tricks of the trade}, pages 421--436.
  Springer, 2012.

\bibitem{cai2021physics}
Shengze Cai, Zhiping Mao, Zhicheng Wang, Minglang Yin, and George~Em
  Karniadakis.
\newblock Physics-informed neural networks ({PINNs}) for fluid mechanics: A
  review.
\newblock {\em arXiv preprint arXiv:2105.09506}, 2021.

\bibitem{cai2021deepm}
Shengze Cai, Zhicheng Wang, Lu~Lu, Tamer~A Zaki, and George~Em Karniadakis.
\newblock {DeepM\&Mnet}: Inferring the electroconvection multiphysics fields
  based on operator approximation by neural networks.
\newblock {\em Journal of Computational Physics}, 436:110296, 2021.

\bibitem{chen1995universal}
Tianping Chen and Hong Chen.
\newblock Universal approximation to nonlinear operators by neural networks
  with arbitrary activation functions and its application to dynamical systems.
\newblock {\em IEEE Transactions on Neural Networks}, 6(4):911--917, 1995.

\bibitem{chi2012response}
Guoyi Chi, Shuangquan Hu, Yanhui Yang, and Tao Chen.
\newblock Response surface methodology with prediction uncertainty: A
  multi-objective optimisation approach.
\newblock {\em Chemical engineering research and design}, 90(9):1235--1244,
  2012.

\bibitem{clevert2015fast}
Djork-Arn{\'e} Clevert, Thomas Unterthiner, and Sepp Hochreiter.
\newblock Fast and accurate deep network learning by exponential linear units
  ({ELUs}).
\newblock {\em arXiv preprint arXiv:1511.07289}, 2015.

\bibitem{de2021uncertainty}
Subhayan De.
\newblock Uncertainty quantification of locally nonlinear dynamical systems
  using neural networks.
\newblock {\em Journal of Computing in Civil Engineering}, 35(4):04021009,
  2021.

\bibitem{de2019hybrid}
Subhayan De, Patrick~T Brewick, Erik~A Johnson, and Steven~F Wojtkiewicz.
\newblock A hybrid probabilistic framework for model validation with
  application to structural dynamics modeling.
\newblock {\em Mechanical Systems and Signal Processing}, 121:961--980, 2019.

\bibitem{de2020transfer}
Subhayan De, Jolene Britton, Matthew Reynolds, Ryan Skinner, Kenneth Jansen,
  and Alireza Doostan.
\newblock On transfer learning of neural networks using bi-fidelity data for
  uncertainty propagation.
\newblock {\em International Journal for Uncertainty Quantification}, 10(6),
  2020.

\bibitem{de2020topology}
Subhayan De, Jerrad Hampton, Kurt Maute, and Alireza Doostan.
\newblock Topology optimization under uncertainty using a stochastic
  gradient-based approach.
\newblock {\em Structural and Multidisciplinary Optimization},
  62(5):2255--2278, 2020.

\bibitem{de2022bi}
Subhayan De, Malik Hassanaly, Matthew Reynolds, Ryan~N King, and Alireza
  Doostan.
\newblock Bi-fidelity modeling of uncertain and partially unknown systems using
  {DeepONets}.
\newblock {\em arXiv preprint arXiv:2204.00997}, 2022.

\bibitem{de2016computationally}
Subhayan De, Erik~A Johnson, Steven~F Wojtkiewicz, and Patrick~T Brewick.
\newblock Computationally-efficient {B}ayesian model selection for locally
  nonlinear structural dynamical systems.
\newblock {\em Journal of Engineering Mechanics}, 144(5):04018022, 2018.

\bibitem{deng2021convergence}
Beichuan Deng, Yeonjong Shin, Lu~Lu, Zhongqiang Zhang, and George~Em
  Karniadakis.
\newblock Convergence rate of {DeepONets} for learning operators arising from
  advection-diffusion equations.
\newblock {\em arXiv preprint arXiv:2102.10621}, 2021.

\bibitem{doostan2011non}
Alireza Doostan and Houman Owhadi.
\newblock A non-adapted sparse approximation of {PDEs} with stochastic inputs.
\newblock {\em Journal of Computational Physics}, 230(8):3015--3034, 2011.

\bibitem{forrester2007multi}
Alexander~IJ Forrester, Andr{\'a}s S{\'o}bester, and Andy~J Keane.
\newblock Multi-fidelity optimization via surrogate modelling.
\newblock {\em Proceedings of the royal society a: mathematical, physical and
  engineering sciences}, 463(2088):3251--3269, 2007.

\bibitem{GANDHI20172}
Puneet Gandhi, S~Adarsh, and KI~Ramachandran.
\newblock Performance analysis of half car suspension model with 4 {DOF} using
  {PID}, {LQR}, {FUZZY} and {ANFIS} controllers.
\newblock {\em Procedia Computer Science}, 115:2--13, 2017.
\newblock 7th International Conference on Advances in Computing \&
  Communications, ICACC-2017, 22-24 August 2017, Cochin, India.

\bibitem{geneva2020modeling}
Nicholas Geneva and Nicholas Zabaras.
\newblock Modeling the dynamics of {PDE} systems with physics-constrained deep
  auto-regressive networks.
\newblock {\em Journal of Computational Physics}, 403:109056, 2020.

\bibitem{ghanem2003stochastic}
Roger~G Ghanem and Pol~D Spanos.
\newblock {\em Stochastic finite elements: a spectral approach}.
\newblock Courier Corporation, 2003.

\bibitem{giunta2006promise}
AA~Giunta, JM~McFarland, LP~Swiler, and MS~Eldred.
\newblock The promise and peril of uncertainty quantification using response
  surface approximations.
\newblock {\em Structures and Infrastructure Engineering}, 2(3-4):175--189,
  2006.

\bibitem{goswami2021physics}
Somdatta Goswami, Minglang Yin, Yue Yu, and George Karniadakis.
\newblock A physics-informed variational {DeepONet} for predicting the crack
  path in brittle materials.
\newblock {\em arXiv preprint arXiv:2108.06905}, 2021.

\bibitem{howard2022multifidelity}
Amanda~A Howard, Mauro Perego, George~E Karniadakis, and Panos Stinis.
\newblock Multifidelity deep operator networks.
\newblock {\em arXiv preprint arXiv:2204.09157}, 2022.

\bibitem{ISO8608}
ISO.
\newblock Mechanical vibration --- road surface profiles --- reporting of
  measured data.
\newblock ISO Standard 8608, International Organization for Standardization,
  2016.

\bibitem{isukapalli1998stochastic}
SS~Isukapalli, A~Roy, and PG~Georgopoulos.
\newblock Stochastic response surface methods (srsms) for uncertainty
  propagation: application to environmental and biological systems.
\newblock {\em Risk analysis}, 18(3):351--363, 1998.

\bibitem{kamalzare2015efficient}
Mahmoud Kamalzare, Erik~A Johnson, and Steven~F Wojtkiewicz.
\newblock Efficient optimal design of passive structural control applied to
  isolator design.
\newblock {\em Smart structures and systems}, 15(3):847--862, 2015.

\bibitem{karniadakis2021physics}
George~Em Karniadakis, Ioannis~G Kevrekidis, Lu~Lu, Paris Perdikaris, Sifan
  Wang, and Liu Yang.
\newblock Physics-informed machine learning.
\newblock {\em Nature Reviews Physics}, 3(6):422--440, 2021.

\bibitem{kayser1989reliability}
Jack~R Kayser and Andrzej~S Nowak.
\newblock Reliability of corroded steel girder bridges.
\newblock {\em Structural Safety}, 6(1):53--63, 1989.

\bibitem{kennedy2001bayesian}
Marc~C Kennedy and Anthony O'Hagan.
\newblock Bayesian calibration of computer models.
\newblock {\em Journal of the Royal Statistical Society: Series B (Statistical
  Methodology)}, 63(3):425--464, 2001.

\bibitem{kingma2014adam}
Diederik~P Kingma and Jimmy Ba.
\newblock Adam: A method for stochastic optimization.
\newblock {\em arXiv preprint arXiv:1412.6980}, 2014.

\bibitem{kovachki2021universal}
Nikola Kovachki, Samuel Lanthaler, and Siddhartha Mishra.
\newblock On universal approximation and error bounds for fourier neural
  operators.
\newblock {\em arXiv preprint arXiv:2107.07562}, 2021.

\bibitem{kovachki2021neural}
Nikola Kovachki, Zongyi Li, Burigede Liu, Kamyar Azizzadenesheli, Kaushik
  Bhattacharya, Andrew Stuart, and Anima Anandkumar.
\newblock Neural operator: Learning maps between function spaces.
\newblock {\em arXiv preprint arXiv:2108.08481}, 2021.

\bibitem{lanthaler2021error}
Samuel Lanthaler, Siddhartha Mishra, and George~Em Karniadakis.
\newblock Error estimates for {DeepONets}: A deep learning framework in
  infinite dimensions.
\newblock {\em arXiv preprint arXiv:2102.09618}, 2021.

\bibitem{li2020fourier}
Zongyi Li, Nikola Kovachki, Kamyar Azizzadenesheli, Burigede Liu, Kaushik
  Bhattacharya, Andrew Stuart, and Anima Anandkumar.
\newblock Fourier neural operator for parametric partial differential
  equations.
\newblock {\em arXiv preprint arXiv:2010.08895}, 2020.

\bibitem{li2020multipole}
Zongyi Li, Nikola Kovachki, Kamyar Azizzadenesheli, Burigede Liu, Kaushik
  Bhattacharya, Andrew Stuart, and Anima Anandkumar.
\newblock Multipole graph neural operator for parametric partial differential
  equations.
\newblock {\em arXiv preprint arXiv:2006.09535}, 2020.

\bibitem{li2020neural}
Zongyi Li, Nikola Kovachki, Kamyar Azizzadenesheli, Burigede Liu, Kaushik
  Bhattacharya, Andrew Stuart, and Anima Anandkumar.
\newblock Neural operator: Graph kernel network for partial differential
  equations.
\newblock {\em arXiv preprint arXiv:2003.03485}, 2020.

\bibitem{li2021markov}
Zongyi Li, Nikola Kovachki, Kamyar Azizzadenesheli, Burigede Liu, Kaushik
  Bhattacharya, Andrew Stuart, and Anima Anandkumar.
\newblock Markov neural operators for learning chaotic systems.
\newblock {\em arXiv preprint arXiv:2106.06898}, 2021.

\bibitem{li2021physics}
Zongyi Li, Hongkai Zheng, Nikola Kovachki, David Jin, Haoxuan Chen, Burigede
  Liu, Kamyar Azizzadenesheli, and Anima Anandkumar.
\newblock Physics-informed neural operator for learning partial differential
  equations.
\newblock {\em arXiv preprint arXiv:2111.03794}, 2021.

\bibitem{lu2021one}
Lu~Lu, Haiyang He, Priya Kasimbeg, Rishikesh Ranade, and Jay Pathak.
\newblock One-shot learning for solution operators of partial differential
  equations.
\newblock {\em arXiv preprint arXiv:2104.05512}, 2021.

\bibitem{lu2019deeponet}
Lu~Lu, Pengzhan Jin, and George~Em Karniadakis.
\newblock {DeepONet}: Learning nonlinear operators for identifying differential
  equations based on the universal approximation theorem of operators.
\newblock {\em arXiv preprint arXiv:1910.03193}, 2019.

\bibitem{lu2021learning}
Lu~Lu, Pengzhan Jin, Guofei Pang, Zhongqiang Zhang, and George~Em Karniadakis.
\newblock Learning nonlinear operators via {DeepONet} based on the universal
  approximation theorem of operators.
\newblock {\em Nature Machine Intelligence}, 3(3):218--229, 2021.

\bibitem{lu2021comprehensive}
Lu~Lu, Xuhui Meng, Shengze Cai, Zhiping Mao, Somdatta Goswami, Zhongqiang
  Zhang, and George~Em Karniadakis.
\newblock A comprehensive and fair comparison of two neural operators (with
  practical extensions) based on {FAIR} data.
\newblock {\em arXiv preprint arXiv:2111.05512}, 2021.

\bibitem{lu2022multifidelity}
Lu~Lu, Rapha{\"e}l Pestourie, Steven~G Johnson, and Giuseppe Romano.
\newblock Multifidelity deep neural operators for efficient learning of partial
  differential equations with application to fast inverse design of nanoscale
  heat transport.
\newblock {\em Physical Review Research}, 4(2):023210, 2022.

\bibitem{ma2004parameter}
F.~Ma, A.~Bockstedte, G.~C. Foliente, P.~Paevere, and H.~Zhang.
\newblock Parameter analysis of the differential model of hysteresis.
\newblock {\em Journal of Applied Mechanics}, 71(3):342--349, 2004.

\bibitem{ma2006system}
F~Ma, CH~Ng, and N~Ajavakom.
\newblock On system identification and response prediction of degrading
  structures.
\newblock {\em Structural Control and Health Monitoring: The Official Journal
  of the International Association for Structural Control and Monitoring and of
  the European Association for the Control of Structures}, 13(1):347--364,
  2006.

\bibitem{marcati2021exponential}
Carlo Marcati and Christoph Schwab.
\newblock Exponential convergence of deep operator networks for elliptic
  partial differential equations.
\newblock {\em arXiv preprint arXiv:2112.08125}, 2021.

\bibitem{Nagarajaiah2000}
S.~Nagarajaiah and X.~Sun.
\newblock Response of base-isolated {USC} hospital building in {Northridge}
  earthquake.
\newblock {\em Journal of Structural Engineering}, 126(10):1177--1186, 2000.

\bibitem{raissi2019physics}
Maziar Raissi, Paris Perdikaris, and George~E Karniadakis.
\newblock Physics-informed neural networks: A deep learning framework for
  solving forward and inverse problems involving nonlinear partial differential
  equations.
\newblock {\em Journal of Computational Physics}, 378:686--707, 2019.

\bibitem{ramallo2002smart}
J~C Ramallo, E~A Johnson, and B~F Spencer, Jr.
\newblock ``{Smart}'' base isolation systems.
\newblock {\em Journal of Engineering Mechanics}, 128(10):1088--1099, 2002.

\bibitem{spencer1998benchmark}
B~F Spencer, S~J Dyke, and H~S Deoskar.
\newblock Benchmark problems in structural control: part {I} -- active mass
  driver system.
\newblock {\em Earthquake Engineering \& Structural Dynamics},
  27(11):1127--1140, 1998.

\bibitem{spencer1}
B.~F. Spencer, Jr. and S.~Nagarajaiah.
\newblock State of the art of structural control.
\newblock {\em Journal of Structural Engineering}, 129(7):845--856, 2003.

\bibitem{thakur2022multi}
Akshay Thakur, Tapas Tripura, and Souvik Chakraborty.
\newblock Multi-fidelity wavelet neural operator with application to
  uncertainty quantification.
\newblock {\em arXiv preprint arXiv:2208.05606}, 2022.

\bibitem{tripathy2018deep}
Rohit~K Tripathy and Ilias Bilionis.
\newblock Deep {UQ}: Learning deep neural network surrogate models for high
  dimensional uncertainty quantification.
\newblock {\em Journal of computational physics}, 375:565--588, 2018.

\bibitem{tripura2022wavelet}
Tapas Tripura and Souvik Chakraborty.
\newblock Wavelet neural operator: a neural operator for parametric partial
  differential equations.
\newblock {\em arXiv preprint arXiv:2205.02191}, 2022.

\bibitem{viana2021survey}
Felipe~AC Viana and Arun~K Subramaniyan.
\newblock A survey of {B}ayesian calibration and physics-informed neural
  networks in scientific modeling.
\newblock {\em Archives of Computational Methods in Engineering}, pages 1--30,
  2021.

\bibitem{wang2021long}
Sifan Wang and Paris Perdikaris.
\newblock Long-time integration of parametric evolution equations with
  physics-informed {DeepONets}.
\newblock {\em arXiv preprint arXiv:2106.05384}, 2021.

\bibitem{wang2021learning}
Sifan Wang, Hanwen Wang, and Paris Perdikaris.
\newblock Learning the solution operator of parametric partial differential
  equations with physics-informed {DeepOnets}.
\newblock {\em arXiv preprint arXiv:2103.10974}, 2021.

\bibitem{wen1976method}
Yi-Kwei Wen.
\newblock Method for random vibration of hysteretic systems.
\newblock {\em Journal of the Engineering Mechanics Division}, 102(2):249--263,
  1976.

\bibitem{williams2006gaussian}
Christopher~K Williams and Carl~Edward Rasmussen.
\newblock {\em Gaussian processes for machine learning}.
\newblock MIT press Cambridge, MA, 2006.

\bibitem{winovich2019convpde}
Nick Winovich, Karthik Ramani, and Guang Lin.
\newblock Convpde-uq: Convolutional neural networks with quantified uncertainty
  for heterogeneous elliptic partial differential equations on varied domains.
\newblock {\em Journal of Computational Physics}, 394:263--279, 2019.

\bibitem{wu2008real}
Meiliang Wu and Andrew Smyth.
\newblock Real-time parameter estimation for degrading and pinching hysteretic
  models.
\newblock {\em International Journal of Non-Linear Mechanics}, 43(9):822--833,
  2008.

\bibitem{xiu2002wiener}
Dongbin Xiu and George~Em Karniadakis.
\newblock The {W}iener--{A}skey polynomial chaos for stochastic differential
  equations.
\newblock {\em SIAM journal on scientific computing}, 24(2):619--644, 2002.

\bibitem{yang2021b}
Liu Yang, Xuhui Meng, and George~Em Karniadakis.
\newblock B-pinns: Bayesian physics-informed neural networks for forward and
  inverse pde problems with noisy data.
\newblock {\em Journal of Computational Physics}, 425:109913, 2021.

\bibitem{yang2020physics}
Liu Yang, Dongkun Zhang, and George~Em Karniadakis.
\newblock Physics-informed generative adversarial networks for stochastic
  differential equations.
\newblock {\em SIAM Journal on Scientific Computing}, 42(1):A292--A317, 2020.

\bibitem{yang2019adversarial}
Yibo Yang and Paris Perdikaris.
\newblock Adversarial uncertainty quantification in physics-informed neural
  networks.
\newblock {\em Journal of Computational Physics}, 394:136--152, 2019.

\bibitem{YOSHIMURA199941}
T~Yoshimura, K~Nakaminami, M~Kurimoto, and J~Hino.
\newblock Active suspension of passenger cars using linear and fuzzy-logic
  controls.
\newblock {\em Control Engineering Practice}, 7(1):41--47, 1999.

\bibitem{zhang2019quantifying}
Dongkun Zhang, Lu~Lu, Ling Guo, and George~Em Karniadakis.
\newblock Quantifying total uncertainty in physics-informed neural networks for
  solving forward and inverse stochastic problems.
\newblock {\em Journal of Computational Physics}, 397:108850, 2019.

\bibitem{zhu2019physics}
Yinhao Zhu, Nicholas Zabaras, Phaedon-Stelios Koutsourelakis, and Paris
  Perdikaris.
\newblock Physics-constrained deep learning for high-dimensional surrogate
  modeling and uncertainty quantification without labeled data.
\newblock {\em Journal of Computational Physics}, 394:56--81, 2019.

\end{thebibliography}

\end{document}